\documentclass{article}

\usepackage{microtype}
\usepackage{graphicx}
\usepackage{caption}
\usepackage{subcaption}
\usepackage{hyperref}
\usepackage{enumitem}
\usepackage{pdflscape}

\usepackage{booktabs}
\usepackage{multicol}
\usepackage{multirow}       
\usepackage{makecell}       
\usepackage{adjustbox}

\newcommand{\I}{\mathbb{I}}
\newcommand{\score}{s} 

\usepackage[preprint]{icml2026}

\usepackage{amsmath}
\usepackage{amssymb}
\usepackage{mathtools}
\usepackage{placeins}
\usepackage[capitalize,noabbrev]{cleveref}

\usepackage[most]{tcolorbox}
\usepackage{xcolor}

\newtcblisting{promptbox}[1][]{
  breakable,
  colback=gray!5,
  colframe=gray!50!black,
  fonttitle=\bfseries,
  boxrule=0.5pt,
  arc=2pt,
  listing only,
  listing options={
    basicstyle=\ttfamily\footnotesize,
    breaklines=true,
    breakatwhitespace=true,
    columns=fullflexible,
    keepspaces=true,
    showstringspaces=false,
  },
  #1
}

\icmltitlerunning{Medmarks}

\begin{document}

\twocolumn[
  \icmltitle{Medmarks: A Comprehensive Open-Source LLM Benchmark Suite for Medical Tasks}

  \begin{icmlauthorlist}
    \icmlauthor{Benjamin Warner}{sophont,medarc}
    \icmlauthor{Ratna Sagari Grandhi}{medarc}
    \icmlauthor{Max Kieffer}{medarc}
    \icmlauthor{Aymane Ouraq}{medarc}
    \icmlauthor{Saurav Panigrahi}{medarc}
    \icmlauthor{Geetu Ambwani}{medarc}
    \icmlauthor{Kunal Bagga}{medarc}
    \icmlauthor{Nikhil Khandekar}{medarc}
    \icmlauthor{Arya Hariharan}{medarc}
    \icmlauthor{Nishant Mishra}{medarc}
    \icmlauthor{Manish Ram}{medarc}
    \icmlauthor{Shamus Sim Zi Yang}{medarc}
    \icmlauthor{Ahmed Essouaied}{medarc}
    \icmlauthor{Adepoju Jeremiah Moyondafoluwa}{medarc}
    \icmlauthor{Robert Scholz}{medarc}
    \icmlauthor{Bofeng Huang}{medarc}
    \icmlauthor{Molly Beavers}{medarc}
    \icmlauthor{Srishti Gureja}{medarc}
    \icmlauthor{Anish Mahishi}{medarc}
    \icmlauthor{Sameed Khan}{medarc}
    \icmlauthor{Maxime Griot}{medarc}
    \icmlauthor{Hunar Batra}{medarc}
    \icmlauthor{Jean-Benoit Delbrouck}{medarc}
    \icmlauthor{Siddhant Bharadwaj}{medarc}
    \icmlauthor{Ronald Clark}{medarc}
    \icmlauthor{Ashish Vashist}{medarc}
    \icmlauthor{Anas Zafar}{medarc}
    \icmlauthor{Leema Krishna Murali}{medarc}
    \icmlauthor{Harsh Deshpande}{medarc}
    \icmlauthor{Ameen Patel}{prime}
    \icmlauthor{William Brown}{prime}
    \icmlauthor{Johannes Hagemann}{prime}
    \icmlauthor{Connor Lane}{sophont,medarc}
    \icmlauthor{Paul Steven Scotti}{sophont,medarc}
    \icmlauthor{Tanishq Mathew Abraham}{sophont,medarc}
  \end{icmlauthorlist}

  \icmlaffiliation{sophont}{Sophont}
  \icmlaffiliation{medarc}{MedARC}
  \icmlaffiliation{prime}{Prime Intellect}

  \icmlcorrespondingauthor{Benjamin Warner}{contact@sophontai.com}

  \icmlkeywords{Medical LLMs, Benchmarking, Evaluation, LLM-as-a-Judge}

  \vskip 0.3in
]

\printAffiliationsAndNotice{}  

\begin{abstract}
Evaluating large language models (LLMs) for medical applications remains challenging due to benchmark saturation, limited data accessibility, and insufficient coverage of relevant tasks. Existing suites have either saturated, heavily depend on restricted datasets, or lack comprehensive model coverage. We introduce \textsc{Medmarks}, a fully open-source evaluation suite with 30 benchmarks spanning question answering, information extraction, medical calculations, and open-ended clinical reasoning. We perform a systematic evaluation of 61 models across 71 configurations using verifiable metrics and LLM-as-a-Judge. Our results show that frontier reasoning models (Gemini 3 Pro Preview, GPT-5.1, \& GPT-5.2) achieve the highest performance across both benchmarks, most frontier proprietary models are significantly more token efficient than open-weight alternatives, medically fine-tuned models outperform their generalist counterparts, and that models are susceptible to answer-order bias (particularly smaller models and Grok 4). A subset of our evals (\textsc{Medmarks-T}) can be directly used  as reinforcement learning environments to post-train LLMs for medical reasoning. Code is available at \url{https://github.com/MedARC-AI/Medmarks}.
\end{abstract}

\section{Introduction}
Large language models (LLMs) have been explored for a variety of medical use-cases, with tasks spanning hospital administrative workflows, clinical decision support, patient-facing chatbots, and more \cite{brodeur2026state}. Additionally, clinicians and other healthcare professionals have begun integrating LLMs into routine workflows, both through public-facing interfaces such as ChatGPT and through LLM-enabled tools embedded within electronic health record systems~\cite{openai2026healthcare, griot_implementation_2025}.

\begin{figure}[ht]
  \begin{center}
    \centerline{\includegraphics[width=\columnwidth]{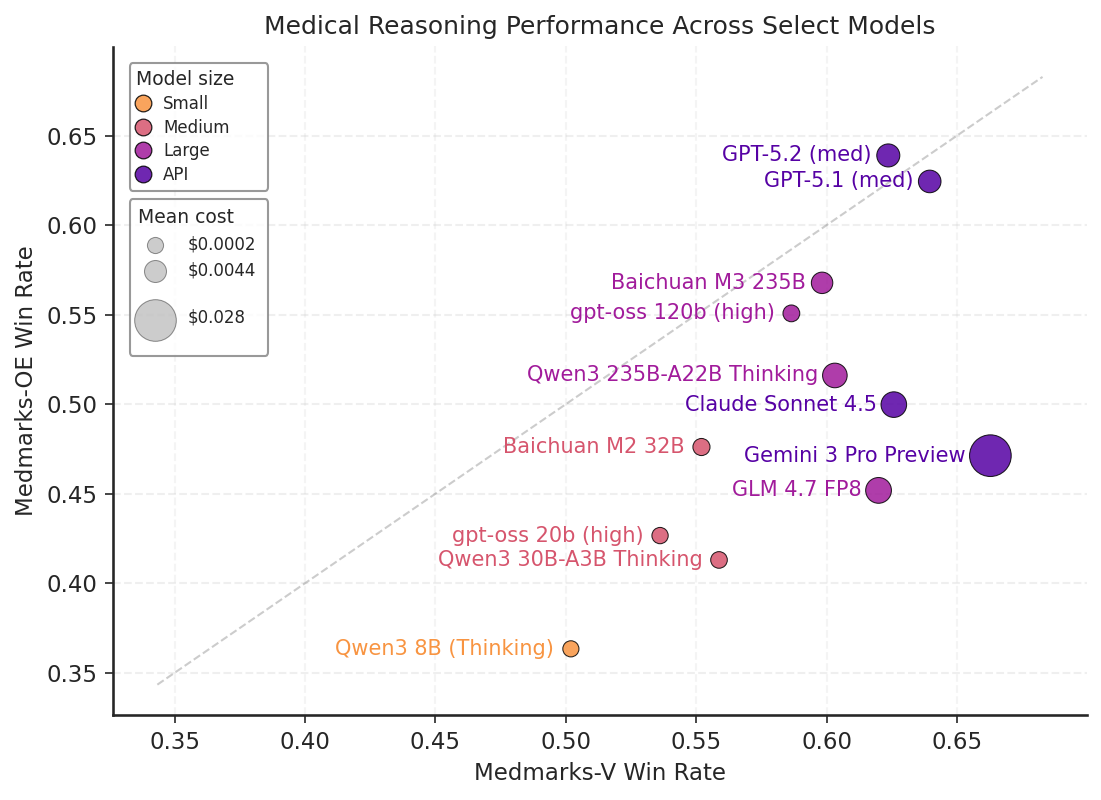}}
    \caption{Results on \textsc{Medmarks-V} and \textsc{Medmarks-OE} for subset of models evaluated on both benchmarks.}
    \label{fig:overall}
  \end{center}
\end{figure}

Accurately tracking the medical capabilities of frontier LLMs and understanding their limitations is crucial to ensuring the safe deployment of current LLMs and to improving future generations of LLMs for medical applications. Medical LLM benchmarks have seen wide adoption but they have all saturated, heavily depend on restricted datasets, or lack comprehensive task and model coverage. For example, the MultiMedQA benchmark suite \cite{singhal2023large} has mostly saturated because it mainly comprises basic question answering tasks on medical knowledge recall that frontier models have largely mastered. Beyond performance saturation, these benchmarks typically do not accurately reflect real-world use cases such as generating a treatment plan from medical reports or open-ended conversations between patients and physicians. The few exceptions are limited in other ways, such as HealthBench \cite{arora2025healthbench} mostly focusing on patient-facing medical conversations or MedHELM \cite{bedi_holistic_2026} being largely restricted to proprietary datasets that prevent community replication. A comparison of \textsc{Medmarks} with these and other medical LLM benchmark suites is in Appendix~\ref{app:suite_comparison}.

There is a need for a regularly updated, fully open-source and easy-to-run medical LLM evaluation suite capable of benchmarking a wide swath of models and datasets across clinically relevant tasks. To this end, we introduce \textsc{Medmarks}, an evaluation suite to assess the medical capabilities of LLMs. To our knowledge, this suite is the largest completely open-source automated evaluation suite for medical capabilities, with a total of 30 benchmarks. These benchmarks are divided into subsets: \textsc{Medmarks-V} (multiple-choice question answering and other verifiable tasks) and \textsc{Medmarks-OE} (open-ended, non-verifiable tasks evaluated using LLM-as-a-Judge). \textsc{Medmarks} benchmarks span question answering, information extraction, consumer health questions, logical reasoning questions, EHR interactions, medical calculations, and more. 

We evaluated a total of 61 models with 71 configurations, including both proprietary model APIs from frontier labs and open-source models running locally, ranking models based on a weighted mean win rate. To our knowledge, \textsc{Medmarks} constitutes the largest publicly documented evaluation of LLMs on medical capabilities to date, covering a broader set of  model families and sizes than prior medical benchmarking efforts. We observe that frontier and open-source reasoning models, such as GPT-5.1 and GPT-5.2, Qwen3 235B-A22B Thinking, and Baichuan M3 235B were consistently among the best performing models on our benchmarks. Also, we observed that smaller medically tuned LLMs like Baichuan-M2 outperformed much larger generalist models like MiniMax M2 and GLM-4.5 Air, highlighting the potential for medical-specific post-training.

To facilitate medical-specific post-training, we separately highlight all datasets from \textsc{Medmarks-V} and \textsc{Medmarks-OE} that come with training/test splits (collectively termed \textsc{Medmarks-T} for "trainable"). Since all datasets in \textsc{Medmarks} are implemented as verifiers environments \cite{brown_verifiers_2025} with corresponding reward functions, researchers can easily use \textsc{Medmarks-T} for further post-training of LLMs on medical reasoning tasks.

To summarize, our primary contributions are as follows:
\begin{itemize}[leftmargin=*]
    \item We collate 30 open-source public benchmarks spanning a variety of tasks from multiple-choice questions to those more closely representative of real-world use-cases. We organize our benchmarks into verifiable and non-verifiable subsets: \textsc{Medmarks-V} and \textsc{Medmarks-OE}.
    \item We evaluated 61 models with 71 different configurations, including proprietary frontier models and open-source models. Overall, frontier reasoning models achieved the highest performance. Open-source models can approach the performance of frontier models but are often very token-inefficient.
    \item All our benchmarks are accessible as verifiers environments \cite{brown_verifiers_2025}, meaning that they are portable to post-training frameworks such as Prime-RL, Tinker, and SkyRL. Environments suitable for training with explicit reward functions (i.e., those with defined train/test splits) are labeled as \textsc{Medmarks-T}. We demonstrate example usage of our framework for reinforcement learning with verifiable rewards (RLVR) on medical tasks.
\end{itemize}

To facilitate reproducibility, we make our benchmarking code open-source at \url{https://github.com/MedARC-AI/medmarks}. The \textsc{Medmarks} leaderboards are accessible at \url{https://medmarks.ai}. We intend for \textsc{Medmarks} to be a living leaderboard that incorporates new models as they are released, tracking the frontier of LLM medical capabilities.

\section{Methods}
\label{sec:benchmark-construction}

\subsection{Benchmark Selection and Summary}

We implement 30 open-source benchmarks, eight with training and evaluation splits (\textsc{Medmarks-T}). Full list of benchmarks is available in Table \ref{tab:datasets}, with details in Appendix \ref{app:dataset_details}.

Within the subset of our benchmarks that are question answering, many of them go beyond basic medical knowledge recall. For example, MedCalc-Bench~\cite{khandekar2024medcalc} evaluates an LLM's ability to perform relevant medical calculations, MetaMedQA~\cite{griot_large_2025} evaluates a model's metacognitive abilities, and SCTpublic~\cite{mccoy2025assessment} evaluates clinical reasoning under uncertainty. We aim to cover a broad set of task types, such as analyzing and correcting clinical notes (MEDEC, \citet{abacha2025medecbenchmarkmedicalerror}), medical coding (MedConceptsQA, \citet{shoham2024medconceptsqa}), and note generation (ACI-Bench, \citet{yim2023aci}).

Beyond verifiable tasks, we include several open-ended benchmarks, such as patient-facing question answering (HealthBench \cite{arora2025healthbench}, MedicationQA \cite{benabacha2019medicationqa}) and diagnostic reasoning (MedCaseReasoning \cite{wu2025medcasereasoning}). 

We also include AgentClinic \cite{schmidgall2024agentclinic} and MedR-Bench \cite{qiu2025quantifying} benchmarks, which are multi-turn agentic environments. These evaluations are closer to realistic use-cases, requiring reasoning about patient interactions, incomplete information, and tool usage. 

\subsection{Grading of Multiple-Choice Questions}

At first glance, grading the results from a multiple-choice benchmark appears easy: either the model returns the correct answer or not. However, in preliminary experiments we noticed models would be given an incorrect grade despite choosing the correct answer due to improper formatting. Smaller models and medically finetuned models had a greater tendency to ignore formatting instructions and give unexpected but correct answers, but even GPT-5.1 would sometimes write a paragraph despite being prompted for a concise answer format.

To resolve this, we constructed a multiple-choice grading function that accepts the multiple-choice letter (or number), the exact answer text, or the letter with the exact answer text. This function also strips dangling thinking traces, normalizes capitalization, punctuation, and whitespace, accepts optional answer prefixes to anchor the answer, and attempts to account for negation. See Appendix~\ref{app:mcq_grade_func} for more details.

\subsection{Evaluating Open-Ended Tasks}

Before modern LLMs, open-ended datasets were typically evaluated in one of three ways: (1) using human graders, which was expensive and not scalable; (2) using lexical overlap metrics such as ROUGE-L \citep{lin2004automatic} or n-gram overlap which tends to fail on semantically identical but different text (e.g., abbreviations or synonyms); or (3) using a semantic similarity score such as BERTscore \citep{zhang2019bertscore}, which lacks the language model's world knowledge. 

Inspired by the MedHELM benchmark \citep{bedi_holistic_2026}, we elected to upgrade these older approaches using LLM-as-a-Judge \citep{gu2025surveyllmasajudge} with modern LLMs. Using LLM-as-a-Judge allows us to account for semantic similarity between reference answers and the evaluated model's answers and benefit from the latest model's broad world knowledge. \footnote{An unanticipated benefit and issue is that some new models have good enough world knowledge to correct poor reference answers.} Like MedHELM, we utilize multiple judge LLMs to evaluate model responses to avoid biases of a single judge LLM.

We reused any LLM judge prompt whenever the original dataset papers defined their LLM judge prompt. If the dataset did not contain a specified prompt, we then used preexisting LLM judge prompts, such as those found in Stanford's HELM benchmark, with light editing when needed. If there was no pre-existing LLM-as-a-Judge prompt, we created our own judge prompts informed by No Free Labels \citep{krumdick2025no} and industry best practices. An example of our prompts can be found in Appendix~\ref{app:sample_judge_prompt}.

Creating a new benchmark suite meant we could also upgrade LLM-as-a-Judge models to the latest high performing small and medium models. We considered modern small and medium LLMs, e.g., GPT-5 mini \& nano and comparable options, and selected GPT-5 mini as our base judge. We paired GPT-5 mini with Grok 4.1 Fast for numerical ratings and Gemini 3 Flash Preview for all other LLM-as-a-Judge ratings for a total of two judges across almost all datasets\footnote{Our human graders preferred GPT-5 nano over all other models for MedCaseReasoning and we used GPT-5 mini instead of GPT 4.1 for HealthBench}. See \cref{app:judge_selection} for details.

\subsection{Model Prompting Details}

Choosing the right prompts is key for effective model evaluation, with entire libraries dedicated to prompt optimization (e.g., DSPy \citep{khattab2023dspy}). For each benchmark in our suite, we selected an appropriate prompt using a tiered approach: First, we chose original benchmark prompts with minimal modification where they existed. If no original prompt was available, we used community standard prompts from HELM \citep{liang2022holistic} or the EleutherAI LM Eval Harness \citep{gao2021framework}. Finally, when neither of these existed, we created our own prompts based on preexisting community prompts. Where existing prompts did not specify output formatting, we paired user prompts with minimal system prompts that instructed the model to think step by step before answering and provided answer-formatting instructions.

For a few datasets we introduced modifications to the task prompt to save time and cost of evaluation. Details are provided in Appendix \ref{app:eval_changes}.

\subsection{Model Evaluation Details}

We evaluated a total of 61 models on 71 configurations, classified by model size (see Figure \ref{benchmark_scores_by_size_violin}). We not only evaluate generalist models but also six medical-specific LLMs: AntAngelMed 100B \cite{antangelmed_2025}, Baichuan-M2 \cite{dou2025baichuan}, Baichuan-M3 \cite{baichuan-m3}, MedGemma-4B, MedGemma-4B 1.5, and MedGemma-27B \cite{sellergren2025medgemma}. Since benchmarks may have a varying set of metrics, we utilize a weighted mean win rate to perform an overall comparison of models across benchmarks (more details in Appendix \ref{app:win_rate}).

Benchmarks were implemented and executed via the verifiers library \cite{brown_verifiers_2025}. For open-source models, we ran inference using a vLLM server \cite{kwon2023efficient} on up to eight H100s on a single node. For API models, we utilized the Prime Intellect Inference API \footnote{\href{https://docs.primeintellect.ai/inference/overview}{Prime Intellect Inference API docs}}, except for Claude Sonnet 4.5 and Gemini 3 Pro Preview, where we used a mixture of Prime Inference, Anthropic, and Gemini APIs.

All language models were evaluated using their officially recommended sampling parameters, e.g. temperature, top\_k, min\_p, etc. Where model creators did not specify sampling parameters, we elected to use community settings or common defaults\footnote{The three models are Llama 3 family, Claude Sonnet 4.5, and Grok 4. The documentation for the latter two models suggests sampling parameter ranges for different tasks, while Llama 3 offered no sampling parameter guidance.}. LLM-as-a-Judge models use their default sampling arguments, falling back to OpenAI's HealthBench \citep{arora2025healthbench} settings for models without specified sampling parameters.

\section{Results}

We organize our results around a series of practical questions relevant to developing and deploying medical LLMs. Our key findings are as follows: (1) frontier reasoning models achieve the highest performance with a notable exception of Gemini 3 Pro Preview on open-ended tasks (\labelcref{sec:medmarks-v-results,sec:medmarks-oe-results}); (2) benchmarks span a wide difficulty gradient, with expert-level clinical reasoning tasks like MedXpertQA \citep{zuo2025medxpertqa} remaining largely unsolved (\labelcref{sec:benchmark-difficulty}); (3) larger models usually outperform smaller ones, though Qwen3 reasoning models perform above their parameter weight class (\labelcref{sec:model-size}); (4) medical fine-tuning yields near-Pareto improvements over base models (\labelcref{sec:medical-ft}); (5) proprietary models are up to 5$\times$ more token-efficient than their open-weight counterparts, revealing a significant optimization gap (\labelcref{sec:efficiency}); (6) reasoning post-training generally improves performance but fails for the Ministral family, and increased reasoning budgets yield near-monotonic gains (\labelcref{sec:reasoning-pt,sec:reasoning-budget}); (7) models tend to generate more tokens on questions they answer incorrectly (\labelcref{sec:overthinking}); (8) quantization is benign at 8-bit but introduces consistent penalties at 4-bit (\labelcref{sec:quantization}); (9) smaller models tend to be more susceptible to answer-order bias (\cref{sec:order-bias}); and (10) \textsc{Medmarks-T} environments can support RL-based medical post-training (\labelcref{sec:medmarks-t}). Together, these analyses provide a snapshot of current medical LLM capabilities, efficiency trade-offs, and evaluation robustness.

\subsection{\textsc{Medmarks-V} Results}
\label{sec:medmarks-v-results}

We evaluated 61 models on 71 configurations across 19 different verifiable benchmarks. This includes multiple choice question answering tasks like MedQA \citep{jin2021disease}, Medbullets \citep{chen2025benchmarking}, etc. but also other verifiable tasks like medical calculations (MedCalc-Bench \citep{khandekar2024medcalc}). The results for the top twelve LLMs on \textsc{Medmarks-V} are presented in Table \ref{tab:verifiable-leaderboard}. The full results for all models and tasks are presented in Figure \ref{fig:benchmark_heatmap}.

\begin{table}
\caption{Top 12 Models on \textsc{Medmarks-V}.}
\label{tab:verifiable-leaderboard}
\begin{tabular}{llr}
\toprule
    Model & Size & Win Rate \\
\midrule
    Gemini 3 Pro Preview & API & 0.6628 \\
    GPT-5.1 (med) & API & 0.6395 \\
    Grok 4 & API & 0.6343 \\
    Claude Sonnet 4.5 & API & 0.6258 \\
    GPT-5.2 (med) & API & 0.6236 \\
    GLM 4.7 FP8 & Large & 0.6199 \\
    Qwen3 235B-A22B Thinking & Large & 0.6032 \\
    Baichuan M3 235B & Large & 0.5983 \\
    Qwen3 Next 80B-A3B Thinking & Large & 0.5888 \\
    MiniMax M2.1 & Large & 0.5882 \\
    gpt-oss 120b (high) & Large & 0.5865 \\
    gpt-oss 120b (med) & Large & 0.5771 \\
\bottomrule
\end{tabular}
\end{table}

\subsection{\textsc{Medmarks-OE} Results}
\label{sec:medmarks-oe-results}

We take a subset of top-performing models across model sizes on \textsc{Medmarks-V} and report their LLM-as-a-Judge scores on \textsc{Medmarks-OE}. Results for all LLMs evaluated on \textsc{Medmarks-OE} are presented in Table \ref{tab:oe-leaderboard}. Full results for all models and tasks are presented in Figure \ref{fig:llm_judge_heatmap}.

\begin{table}[t]
  \caption{\textsc{Medmarks-OE}.}
  \label{tab:oe-leaderboard}
  \centering
  \begin{tabular}{llr}
    \toprule
    Model & Size & Win Rate \\
    \midrule
    GPT-5.2 (med) & API & 0.6389 \\
    GPT-5.1 (med) & API & 0.6244 \\
    Baichuan M3 235B & Large & 0.5678 \\
    gpt-oss 120b (high) & Large & 0.5507 \\
    Qwen3 235B-A22B Thinking & Large & 0.5161 \\
    Claude Sonnet 4.5 & API & 0.4998 \\
    Baichuan M2 32B & Medium & 0.4761 \\
    Gemini 3 Pro Preview & API & 0.4713 \\
    GLM 4.7 FP8 & Large & 0.4519 \\
    gpt-oss 20b (high) & Medium & 0.4266 \\
    Qwen3 30B-A3B Thinking & Medium & 0.4130 \\
    Qwen3 8B (Thinking) & Small & 0.3634 \\
    \bottomrule
  \end{tabular}
\end{table}

With the exception of MedCaseReasoning and HealthBench\footnote{For MedCaseReasoning, our graders overwhelmingly preferred GPT-5 nano over all other candidate models and for HealthBench we replaced the slow and expensive GPT-4.1 with GPT-5 mini.}, all datasets were evaluated using two judges: GPT-5 mini paired with either Gemini 3 Flash Preview or Grok 4.1 Fast following the selection process described in \cref{sec:benchmark-construction}.

\subsection{How Difficult Are the Benchmarks?}
\label{sec:benchmark-difficulty}

We observe that \textsc{Medmarks-V} benchmarks span a range of difficulty.

Figure \ref{benchmark_scores_violin_llm_models} reports the distribution of model performance for each benchmark. The simplest benchmarks cluster near the top with mean scores above 0.80, led by PubHealthBench \citep{harris2025pubhealthbench} (0.831), LongHealth \citep{adams2025longhealth} (0.829 \& 0.802), and CareQA \citep{ariasduart2025automaticevaluationhealthcarellms} (0.825), demonstrating that models handle simple long-context and general knowledge tasks effectively. A broad middle tier spanning mean scores of 0.368 to 0.784 includes datasets of moderate difficulty such as M-ARC \citep{kim2025limitations} (0.368), MedCalc-Bench \citep{khandekar2024medcalc} (0.439), MedConceptsQA \citep{shoham2024medconceptsqa} (0.526-0.781), MedMCQA \citep{pal2022medmcqa} (0.656), and MedQA \citep{jin2021disease} (0.784).\footnote{MedCalc-Bench tests the medical calculation capabilities of LLMs without any calculator tool. We also evaluated tool-calling capable models on MedCalc-Bench with a python and calculator tool, \cref{app:medcalcbench_tools}.} The most challenging benchmark in the Verifiable subset is MedXpertQA \citep{zuo2025medxpertqa} (0.236-0.237), making it the benchmark with the most headroom for future models to improve on.

Turning our attention to \textsc{Medmarks-OE}, we likewise see a range of difficulty, with MedCaseReasoning \citep{wu2025medcasereasoning} and HealthBench \citep{arora2025healthbench} among the hardest datasets. Interestingly, MedXpertQA is about as difficult for these models as the hard subset of HealthBench.

The performance gap from easiest to hardest datasets reveals a clear difficulty gradient, suggesting models excel at simpler long-context benchmarks but struggle substantially with the specialized medical reasoning and understanding required for expert-level clinical performance.

\subsection{How Does Model Performance Change With Model Size?}
\label{sec:model-size}

\begin{figure}[ht]
  \begin{center}
    \centerline{\includegraphics[width=\columnwidth]{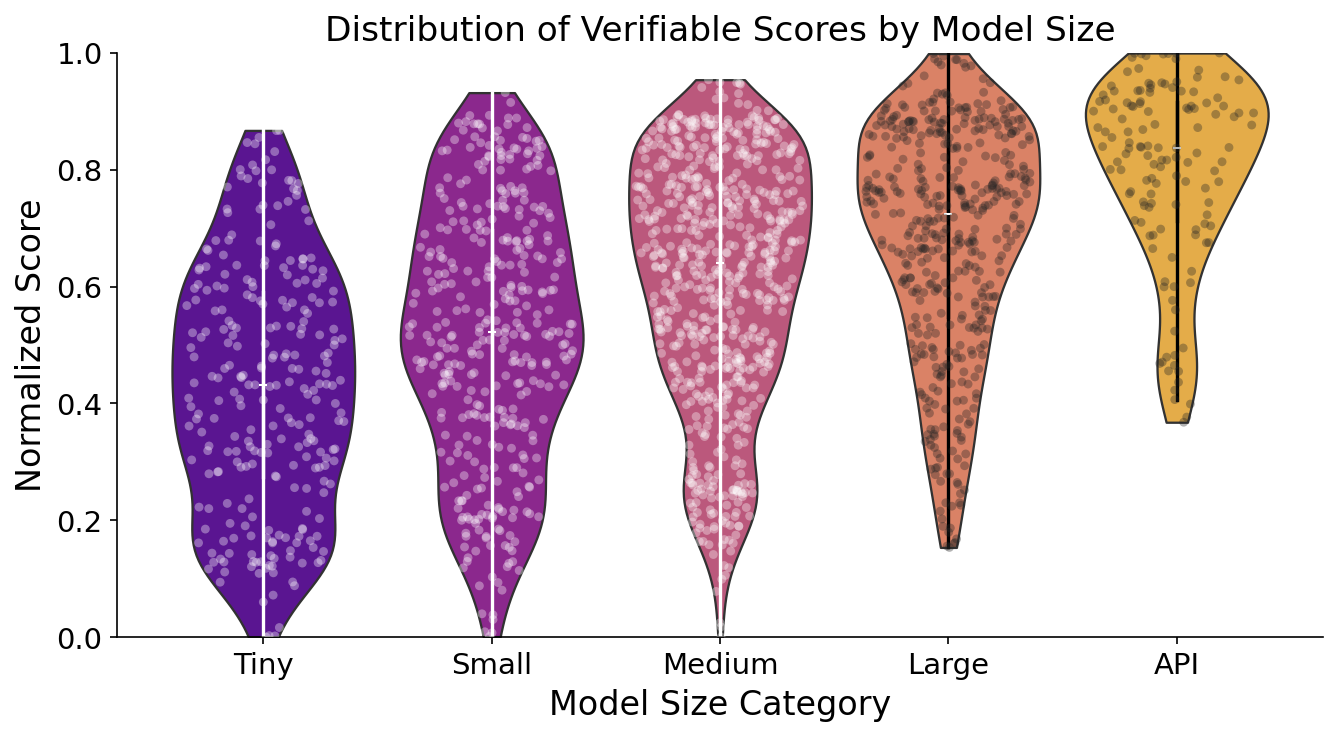}}
    \caption{Distribution of model scores based on model category for the \textsc{Medmarks-V} subset.}
    \label{benchmark_scores_by_size_violin}
  \end{center}
\end{figure}

Overall, larger models usually outperform smaller models across all benchmarks (\cref{benchmark_scores_by_size_violin}), with notable exceptions highlighted in \cref{tab:over-under}.

On the simplest tasks like LongHealth Task 1 and LongHealth Task 2 \citep{adams2025longhealth}, all model sizes achieve reasonable performance, though substantial gaps remain. Large models reach 0.87 on these tasks, while tiny models achieve only 0.57-0.70 demonstrating that even on easy benchmarks, model parameter count provides meaningful advantages.

Conversely, on the hardest datasets like MedXpertQA \citep{zuo2025medxpertqa}, all model sizes cluster near the bottom (large: $\sim 0.29$, tiny: $\sim 0.15$), suggesting these tasks exceed current medical capabilities across the entire size spectrum (\cref{performance_by_size_scatter}).

The performance advantage of larger models is most pronounced on moderate-to-difficult datasets. For example, on MedConceptsQA Easy, the gap reaches 0.531 (large=0.945 vs. tiny=0.414), while on the hardest MedXpertQA tasks, the gap narrows to just 0.14. This suggests model size matters significantly for tasks within the capability range of current models, but provides diminishing returns on extremely difficult specialized medical reasoning tasks.

As \cref{tab:over-under} shows, there are some exceptions to this trend. Qwen3 \cite{yang2025qwen3} thinking models consistently punch above their weight class, with 4B outperforming the average Small model, 14B outperforming the average Medium model, and 30B-A3B outperforming the average Large model.\footnote{Qwen3 30B-A3B FP8 and AWQ 8-bit also outperformed the next size, but were not shown for brevity.}

\begin{table}[t]
    \centering
    \caption{\textsc{Medmarks-V} over- and underperformance}
    \label{tab:over-under}
    \begin{subtable}{\linewidth}
        \centering
        \caption{Average win rate by size}
        \begin{tabular}{lllll}
            \toprule
            Tiny & Small & Medium & Large & API \\
            \midrule
            0.373 & 0.455 & 0.502 & 0.546 & 0.645 \\
            \bottomrule
        \end{tabular}
    \end{subtable}

    \vspace{1em} 

    \begin{subtable}{\linewidth}
        \centering
        \caption{Individual models with notable over- or underperformance}
        \begin{tabular}{llr}
            \toprule
            Model & Size & Win Rate \\
            \midrule
            Granite 4.0H Tiny         & Small  & 0.350 \\
            Olmo 3 7B Instruct        & Small  & 0.362 \\
            Jamba2 Mini 52B           & Large  & 0.453 \\
            Qwen3 4B Thinking         & Tiny   & 0.483 \\
            Hermes 4 70B              & Large  & 0.483 \\
            Ling Flash 2.0            & Large  & 0.487 \\
            AntAngelMed 100B          & Large  & 0.489 \\
            Qwen3 14B Thinking        & Small  & 0.544 \\
            Baichuan M2 32B           & Medium & 0.558 \\
            Qwen3 30B-A3B Thinking    & Medium & 0.564 \\
            \bottomrule
        \end{tabular}
    \end{subtable}
\end{table}

Additionally, there are models, like IBM's Granite 4.0, Olmo 3 \citep{olmo2025olmo3}, and Hermes 4 70B, which consistently underperform models in the weight class below them.

\subsection{Do Medical-Specific LLMs Perform Better Than Their General-Purpose Counterparts?}
\label{sec:medical-ft}

It is commonly debated whether general-purpose LLMs are sufficient \cite{nori2023generalistfoundationmodelsoutcompete} or if we need to specifically train domain-specific LLMs for the medical domain \cite{Lehman2023DoWS}. We evaluated six recent medical LLMs: Baichuan-M2 \citep{dou2025baichuan}, Baichuan-M3 \cite{baichuan-m3}, AntAngelMed 100B \cite{antangelmed_2025}, MedGemma-4B, MedGemma-4B 1.5, and MedGemma-27B \citep{sellergren2025medgemma}. We omit other medical LLMs due to being severely outdated and finetuned from outdated base models.

\begin{figure}[ht]
  \begin{center}
    \centerline{\includegraphics[width=\columnwidth]{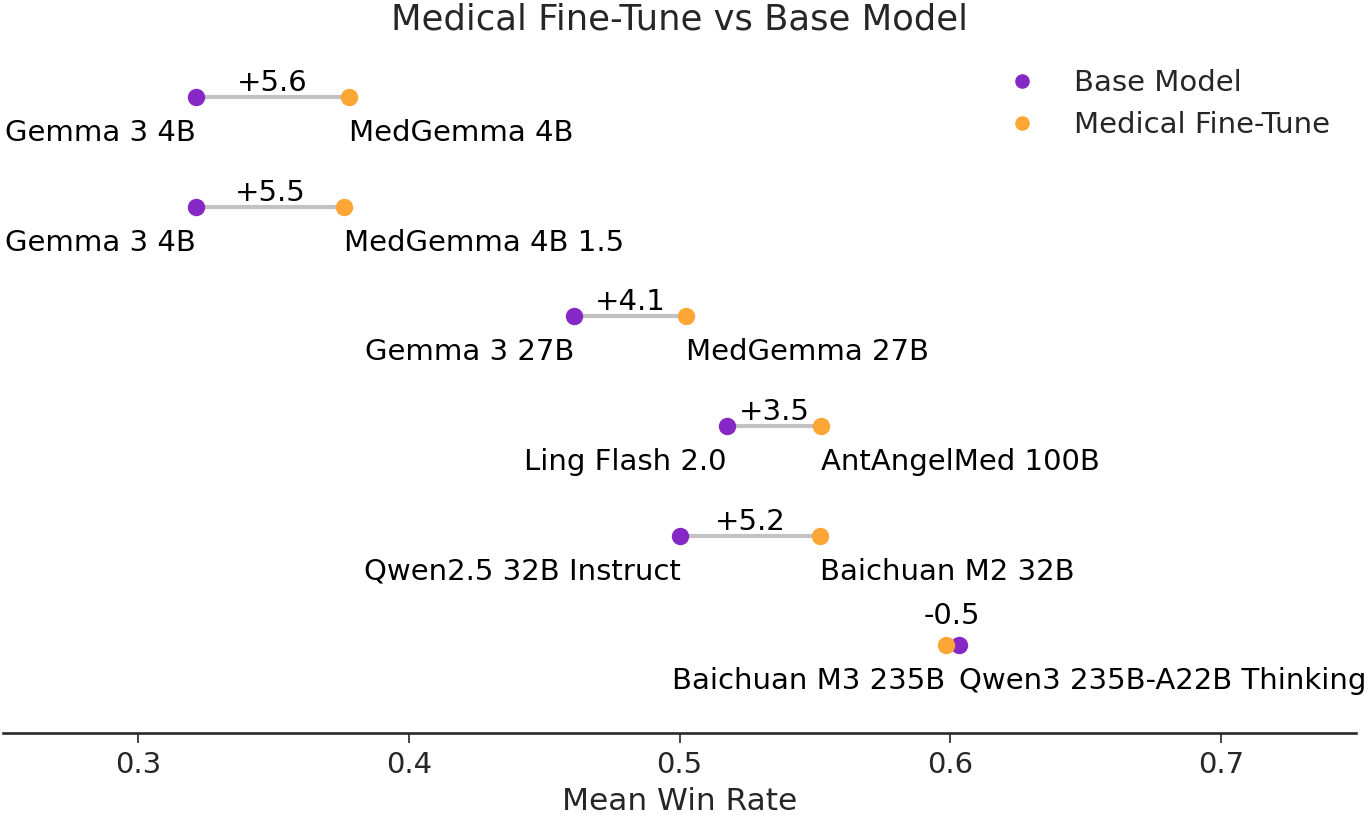}}
    \caption{Win Rate change between medical finetunes and their base models}
    \label{fig:med_comparison_compact}
  \end{center}
\end{figure}

\cref{fig:med_comparison_compact} plots the mean win rate of the medical LLMs and corresponding generalist models on \textsc{Medmarks-V}. We note a significant boost in mean win rate from both Gemma 3 4B to MedGemma 4B \& 4B 1.5 (0.321 to 0.378 \& 0.376, respectively), Gemma 3 27B to MedGemma 27B (0.461 to 0.502), and Ling Flash 2.0 to AntAngelMed (0.517 to 0.552). These gains hold across the majority of benchmarked health datasets, providing evidence that adapting models to the medical domain can be quite useful.

Baichuan-M2 outperforms its base model, Qwen 2.5 32B (0.552 to 0.500), but this is comparing a reasoning model to an instruct model. It is also narrowly outperformed by a newer generalist model, Qwen3 30B-A3B \citep{yang2025qwen3} on \textsc{Medmarks-V} (0.552 to 0.559) but still outperforms it on \textsc{Medmarks-OE} (0.476 to 0.413). Baichuan-M3 breaks this trend by underperforming Qwen3 235B-A22B Thinking on Verified tasks (0.598 to 0.603). However, Baichuan M3 has excellent performance on Open-Ended tasks, outperforming Qwen3 0.5678 to 0.5161, suggesting a better training recipe could increase performance on both benchmarks.

We hypothesize that further medical domain adaptation of the best-performing general-purpose language models will lead to even further gains in performance.

\subsection{Which Models Are More Cost-Efficient and Token-Efficient?}
\label{sec:efficiency}

\begin{figure}[ht]
  \begin{center}
    \centerline{\includegraphics[width=\columnwidth]{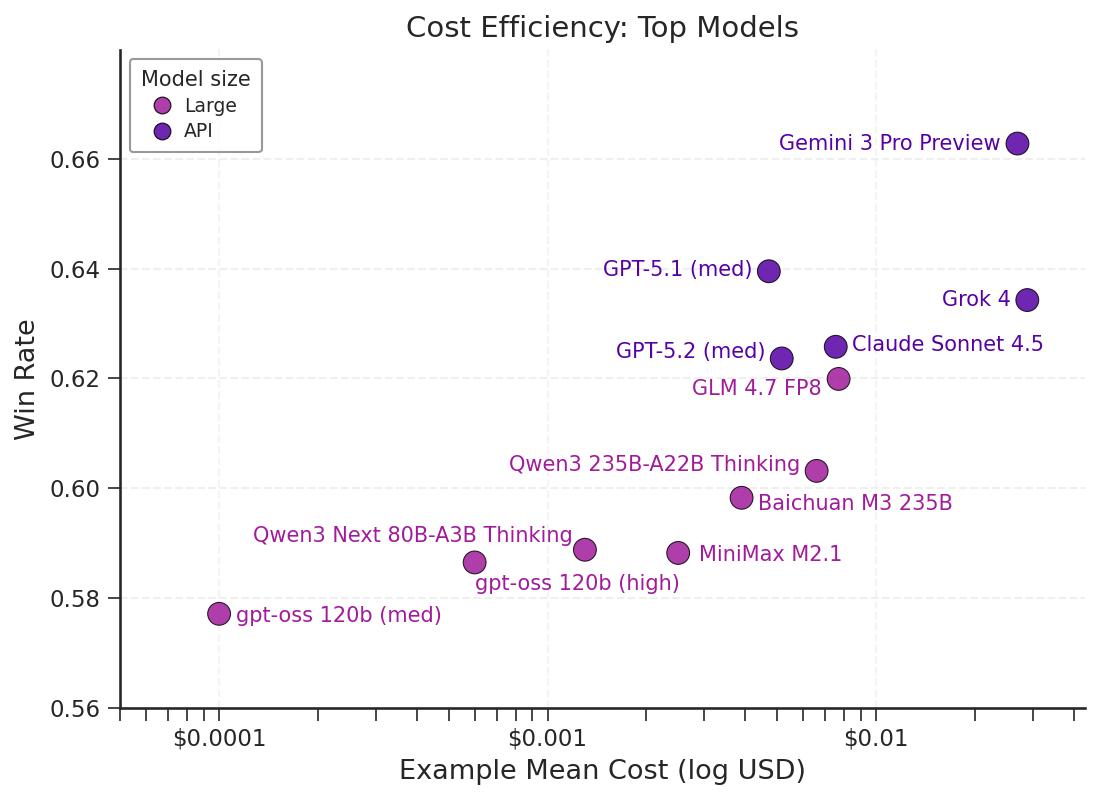}}
    \caption{A scatter plot of the mean win rate on \textsc{Medmarks-V} by cost for the model APIs evaluated.}
    \label{fig:accuracy_vs_cost_scatter}
  \end{center}
\end{figure}

We measured average inference cost per example and the total inference cost on \textsc{Medmarks-V} for the top 12 performing models, as shown in \cref{fig:accuracy_vs_cost_scatter}. We estimated the cost of running all local models on \textsc{Medmarks-V} using a price of \$2 per H100 hour. \cref{tab:estimated-inference} shows the full model set. 

Of the five API models evaluated, both Gemini 3 Pro Preview and Grok-4 stand out in their expense. Both cost an order of magnitude more per query on our verifiable medical benchmarks. On the other hand, GPT-5.1 is the cheapest while also outperforming Grok 4. Despite underperforming GPT-5.1 (med), Claude Sonnet 4.5, and GPT-5.2 (med), the best open model, GLM-4.7 FP8, costs more to run per query than the most cost-efficient frontier models using our H100-hour estimate. Baichuan M3 is noticeably cheaper to run than its base model, Qwen3 235B-A22B Thinking, due to the latter's prolific use of reasoning tokens as discussed in the next paragraph\footnote{One caveat with this analysis is while we attempted to use efficient vLLM baselines for all models, we did not have time to dial in optimized vLLM settings on all 61 models.}.

\begin{figure}[ht]
  \begin{center}
    \centerline{\includegraphics[width=\columnwidth]{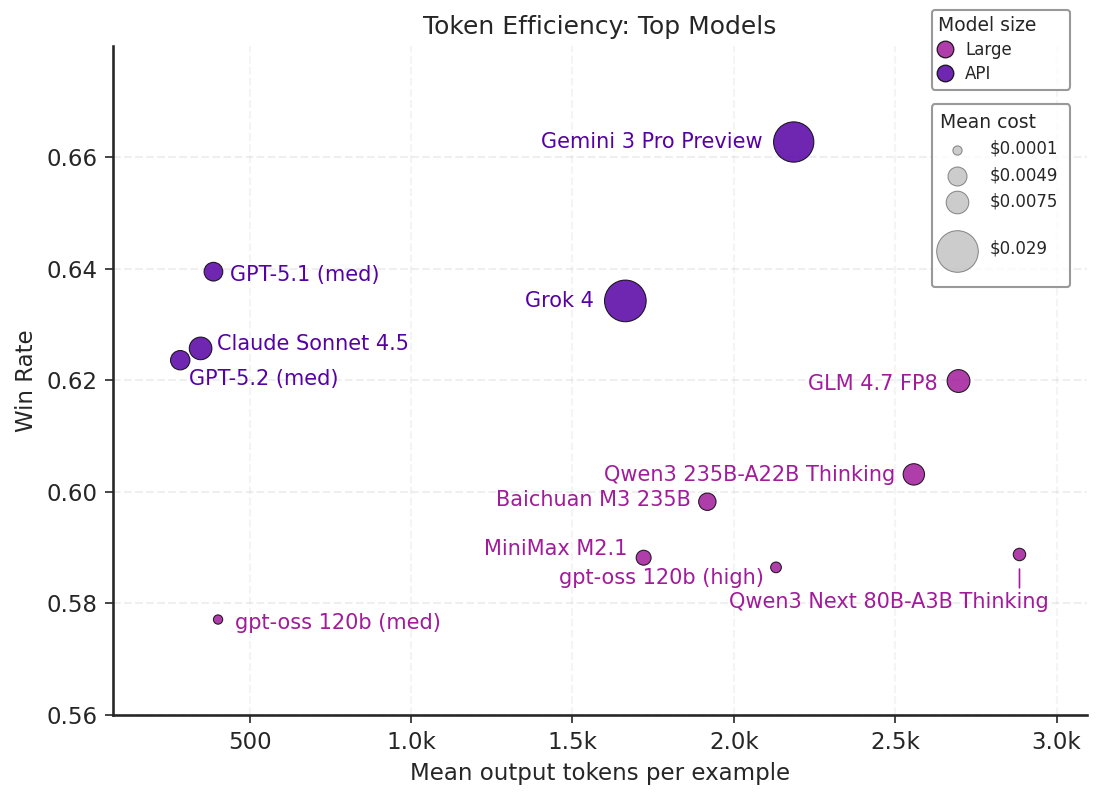}}
    \caption{A scatter plot of mean win rate on \textsc{Medmarks-V} by tokens for top 12 models evaluated.}
    \label{fig:top_token_efficency}
  \end{center}
\end{figure}

We also recorded the token use of all 61 models and 71 variations on our verifiable datasets, with the top 12 models shown in \cref{fig:top_token_efficency} and all 71 in \cref{token_efficiency}. The results reveal a massive optimization gap across thinking models. The Pareto frontier is dominated by frontier reasoning models like GPT 5.1/5.2 and non-thinking models like Claude Sonnet 4.5\footnote{We elected to evaluate frontier models in their default settings, medium reasoning for GPT 5.1/5.2: non-thinking for Sonnet 4.5, and high reasoning for Gemini 3 Pro Preview}, which achieve high mean win rate ($>0.62$) while keeping token use remarkably low ($<500$ tokens). Grok-4 and Gemini 3 Pro Preview once again stand out as an exception, revealing their significant cost is due to brute forcing answers with a large reasoning token spend. As we consider large open-weight alternatives (models that can fit on a single H100 node), this efficiency further collapses. GLM 4.7 FP8 approaches the win rate of GPT-5.2 (med) but demands over 5x the token volume ($\sim 2{,}700+$) to come up short. On the efficient side of the open model Pareto frontier, gpt-oss-120b (med) is nearly as token efficient as both GPT 5 models, but performs significantly worse. This bifurcation indicates that while open architectures are starting to close in on performance of frontier reasoning models on verifiable medical datasets, they have not yet solved the computational cost of the reasoning process.

\subsection{Does Reasoning Post-Training Improve Model Performance?}
\label{sec:reasoning-pt}

\begin{figure}[ht]
  \begin{center}
    \centerline{\includegraphics[width=\columnwidth]{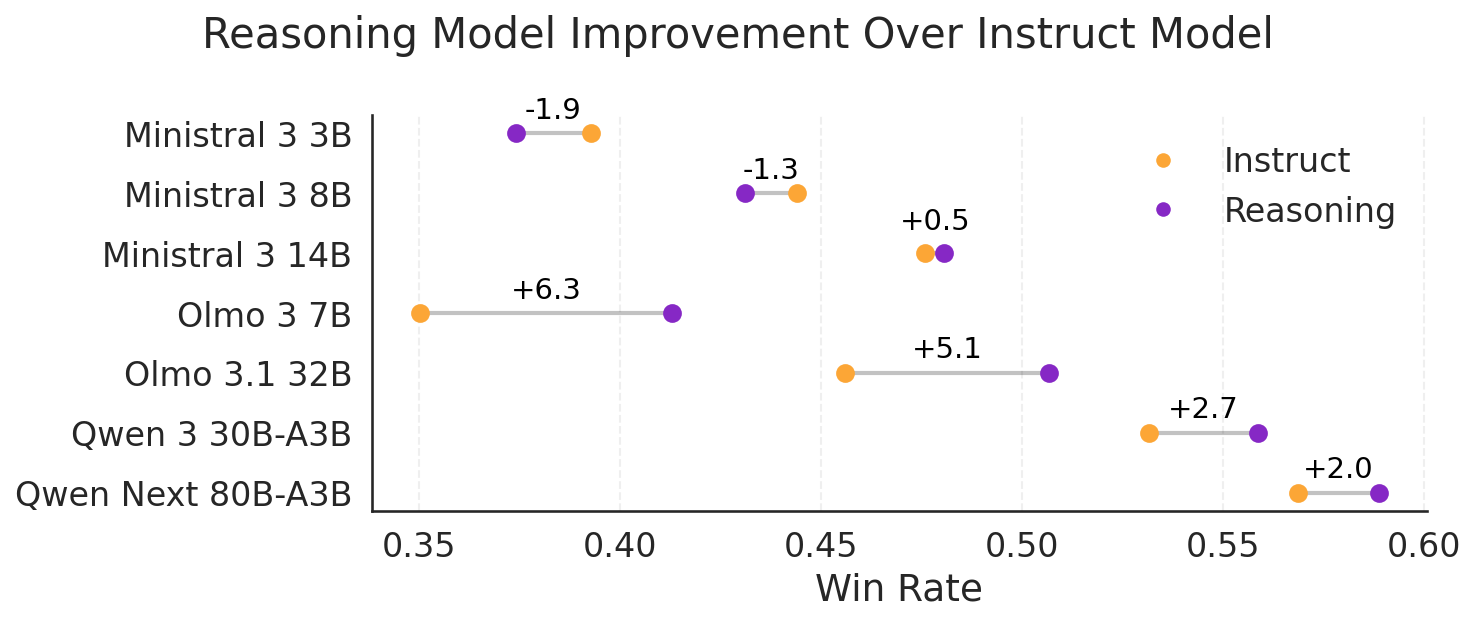}}
    \caption{Win Rate change between instruction and reasoning models}
    \label{fig:model_family_comparison_compact}
  \end{center}
\end{figure}

How do thinking models perform compared to their instruction-tuned counterparts? In general, post-training a base or instruct model into a reasoning model using the modern bag of tricks increases the score of the reasoning model relative to the comparable instruction model. We can see this in \cref{fig:model_family_comparison_compact}, where "Reasoning" variants of the model usually have a higher mean win rate than "Instruct" variants.

Adding reasoning does not always guarantee improvement, as seen with the Ministral family of models. Here, the reasoning models underperform, sometimes significantly, compared to their instruction counterparts on almost all datasets. From our medical benchmark alone, we cannot tell if this is a case of catastrophic forgetting or overfitting, overly divergent post-training data between the instruction and reasoning models, undertrained reasoning models, an issue with verifiable rewards, or something else.

We also compared VLMs vs LLMs by benchmarking Qwen3 VL 30B-A3B \citep{bai2025qwen3vltechnicalreport} against Qwen3 30B-A3B. With these two models we see a decrease in medical performance when adding multimodal support. This suggests a more careful training regimen may be needed when adding multimodal support to a pure language model.

\subsection{Do Models Overthink When They Fail?}
\label{sec:overthinking}

While reasoning models show improved performance, it raises the question of how reasoning models behave when they get questions wrong. We select the best, worst, and two midrange models per size category (excluding ``duplicate models'' like gpt-oss reasoning levels and Olmo 3 vs 3.1) and compare how many tokens they generated for correct and incorrect responses.

\cref{thinking_length_violin} highlights a subset of reasoning models to show the trend that more tokens are typically generated for questions that are answered incorrectly. This trend holds across model size and model providers. GLM-4.5 Air and Mirothinker 1.5 30B are particularly interesting outliers where a large proportion of the responses with incorrect outputs had very long generations.\footnote{Of note, Mirothinker 1.5 30B most consistently ran into the maximum generation limit, doing so ~13\% of the time, followed by DASD 30B-A3B (~8\%), Trinity Nano Preview (~3\%), GLM 4.5 Air (~2\%), MedGemma 4B 1.5 (~2\%), Trinity Mini (~1\%), and SmolLM3 3B (~1\%) rounding up the list of models hitting our ~32K token cap more than 1\% of evaluations.}

\subsection{Does Increased Reasoning Effort Improve Performance?}
\label{sec:reasoning-budget}

OpenAI's gpt-oss \cite{openai_gpt-oss-120b_2025} models allow us to directly compare thinking budgets across multiple model sizes. OpenAI's gpt-oss models 20B (with 4B active parameters) and 120B (with 5B active parameters) support setting reasoning effort between low, medium (the default), and high, with higher reasoning effort tending to produce more reasoning tokens before the final answer. This allows us to directly test in a controlled setting the effect of more reasoning tokens on downstream performance.

\begin{figure}[ht]
  \begin{center}
    \centerline{\includegraphics[width=\columnwidth]{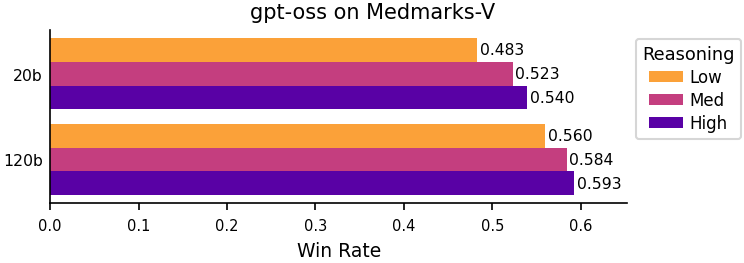}}
    \caption{Win Rate change between gpt-oss reasoning level}
    \label{fig:gpt_oss_comparison}
  \end{center}
\end{figure}

As shown in \cref{fig:gpt_oss_comparison}, we observe that while gpt-oss 20b (medium) and gpt-oss 120b (medium) are well-performing models in their own right, increasing the reasoning token budget to high produces stronger results\footnote{And vice-versa for decreasing the reasoning budget to low.}. gpt-oss 120b (high) is the eleventh strongest model on our verified benchmark suite, outperforming three similar sized or larger models with more active parameters: GLM-4.5 Air (106B-A12B) \citep{zeng2025glm}, INTELLECT-3 (106B-A12B) \cite{primeintellectteam2025intellect3technicalreport}, and MiniMax M2 (230B-A10B) \cite{minimax_m2.1_minimaxio_2025}. gpt-oss 120b (high) \cite{openai_gpt-oss-120b_2025} is the strongest Western open-source model tested, and gpt-oss 20b high and medium reasoning easily beat other modern Western models, including Gemma 3, MedGemma, and Olmo 3 series of models.

From the per-dataset results in \cref{gpt_oss_bar_comparison}, we can see that increasing the reasoning level results in an almost Pareto improvement on model performance across datasets except PubMedQA \citep{jin2019pubmedqa} and MedHALT-NOTA \citep{pal2023med}.

\cref{gpt_oss_violin} demonstrates that gpt-oss-20B and 120B also exhibit the ``overthinking'' problem where models spend more tokens to reason on questions they eventually get incorrect. However, given that increasing reasoning tokens increases performance overall, it is incorrect to conclude that increased thinking leads to more incorrect answers. Rather it would appear that harder questions, or questions the model doesn't know the answer to, lead to more reasoning in an attempt to figure out the correct answer.

\subsection{Does Quantization Affect Model Performance?}
\label{sec:quantization}

\begin{figure}[ht]
  \begin{center}
    \centerline{\includegraphics[width=\columnwidth]{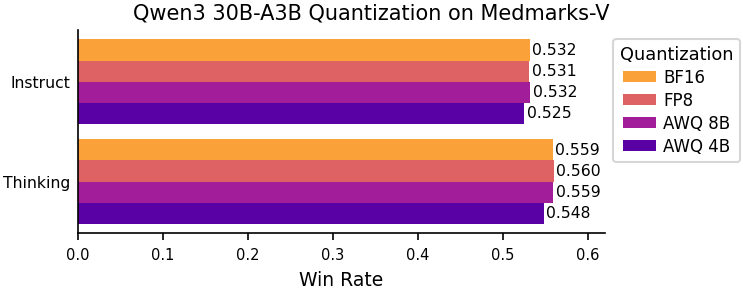}}
    \caption{Win Rate change between quantized models}
    \label{icml-quantization_comparison_compact}
  \end{center}
\end{figure}

Often, models are quantized to save memory and increase inference speed. However, depending on the quantization method, this can degrade model performance. Here we study this in the context of our medical benchmarks, focusing on Qwen3 30B-A3B Instruct and Thinking as examples. We ran both the official BF16 precision and FP8 quantized weights and community-quantized AWQ \cite{lin2024awqactivationawareweightquantization} 8-bit and 4-bit versions (\cref{icml-quantization_comparison_compact}).

Our results show minimal performance degradation from quantization on most datasets, with the more aggressively quantized AWQ 4-bit model suffering a small but consistent penalty on most datasets.

Our results align with \cite{zheng2025empirical}, who report that model degradation starts at 4-bit quantization for AWQ and GPTQ formats, and becomes more pronounced with more aggressive schemes (3-bit, 2-bit, and A4W8 SmoothQuant) and in smaller models\footnote{Presumably quantization aware trained models, such as gpt-oss which was released in mixed bf16-mxfp4 precision, result in less performance degradation than other quantization methods, but we are unable to test this.}.

\subsection{Is There Order Bias for Multiple Choice Tasks?}
\label{sec:order-bias}

Prompt format can significantly affect the performance of foundation models. \citet{gu2025illusion} found that modern vision-language models are susceptible to varying the order of multiple choice answers on multimodal medical benchmarks.

We chose to run three rollouts of almost all our multiple choice benchmarks. The first rollout used the original dataset order and the second two rollouts involved randomly shuffling answer order to test if modern language models are biased from answer choice position.

The difference between the maximum and minimum score for each of the multiple choice datasets (which we refer to here as spread) is visualized in \cref{benchmark_variance_heatmap}. While the effects are not evenly distributed across datasets, we can draw two conclusions from this chart: 1) modern language models can still be tripped up by multiple choice question order, even frontier models, 2) smaller models tend to suffer more from choice order effects than larger models.

As a specific example, we observed Grok 4 exhibits high scoring variance when shuffling the answer order in M-ARC. Because M-ARC is a small dataset, we spot-checked variance by running additional rollouts with the same answer order for all three frontier models. In this setting, Grok 4's variance decreased from 0.11 to 0.02. In contrast, GPT-5.1 (medium)'s spread decreased from 0.05 to 0.03 and Sonnet-4.5 from 0.05 to 0.02. This suggests that in Grok 4's case, multiple choice answer reordering has a larger effect than model sampling randomness.

MedBullets \citep{chen2025benchmarking} gives us another means to examine this question. MedBullets has two flavors: four options and five options, with the latter adding one extra incorrect answer. If models are confident in their answers, then adding an additional distraction answer should result in little to no change in the model's score.

\begin{figure}[ht]
  \begin{center}
    \centerline{\includegraphics[width=\columnwidth]{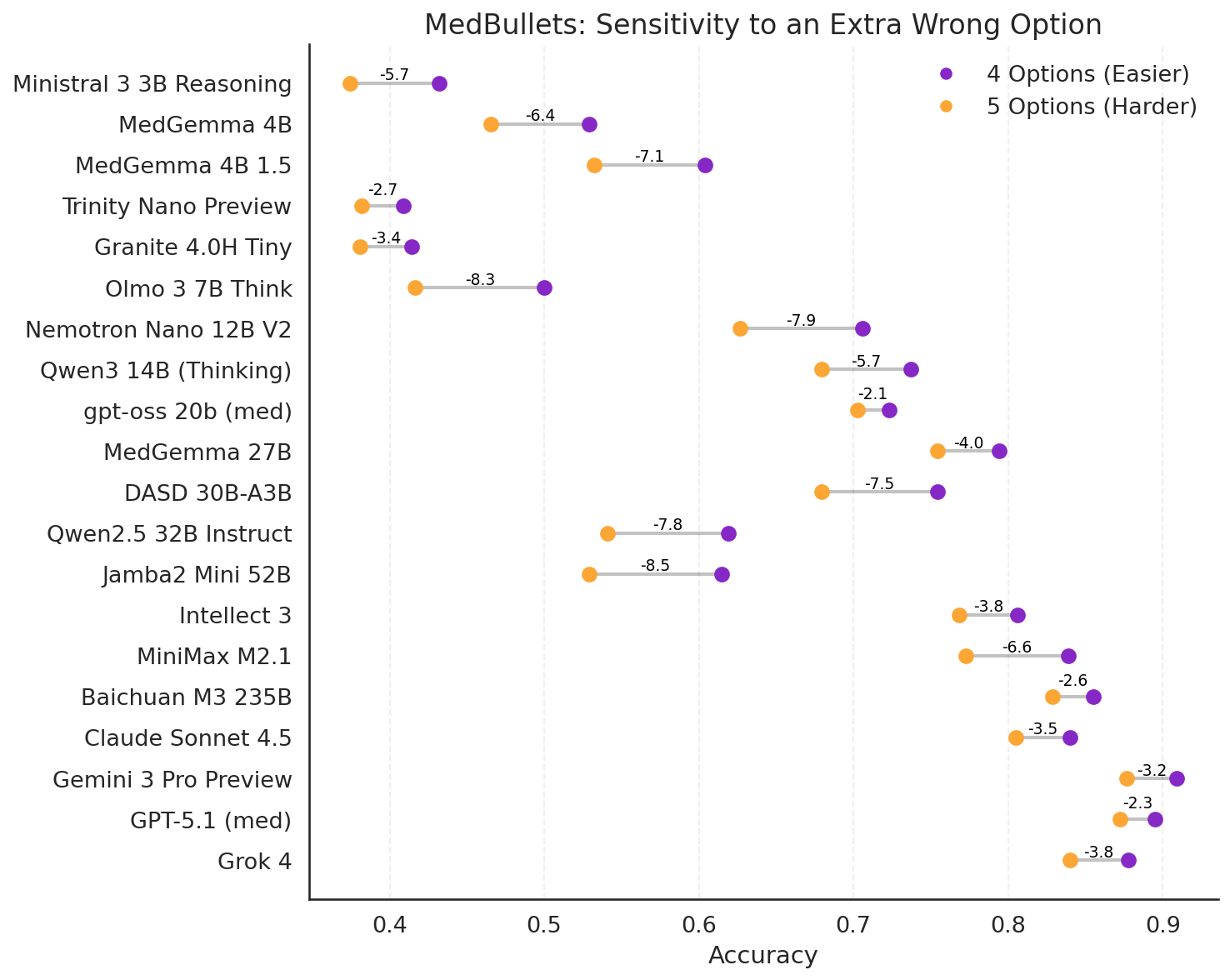}}
    \caption{Comparing model performance with and without an extra option on the Medbullets \cite{chen2025benchmarking} benchmark.}
    \label{distractor_stress_test}
  \end{center}
\end{figure}

Plotting the subset of average accuracy across all rollouts in \cref{distractor_stress_test} (all models in \cref{distractor_stress_test_all_models}), we can see that the combination of adding an additional answer along with random shuffling of the answer orders can result in a large negative reduction in performance. This ``distractability'' is greater for smaller models and/or models which are not as confident in their answers. As one would expect, larger models appear to be less distractible than smaller models. As expected from the variation analysis, Grok 4 is more easily distractible than its frontier model peers.

\subsection{Medical-specific Post-Training with \textsc{Medmarks-T}}
\label{sec:medmarks-t}

Since we implemented the datasets in the verifiers framework \citep{brown_verifiers_2025}, the datasets that come with train/test splits can be easily used as RL environments to post-train LLMs. The datasets with training splits are: MedQA, MedMCQA, PubMedQA, MedCalc-Bench, MeQSum, MEDEC, MedDialog, and MedCaseReasoning. These datasets comprise our \textsc{Medmarks-T} subset.

\begin{figure}[ht]
  \begin{center}
    \centerline{\includegraphics[width=\columnwidth]{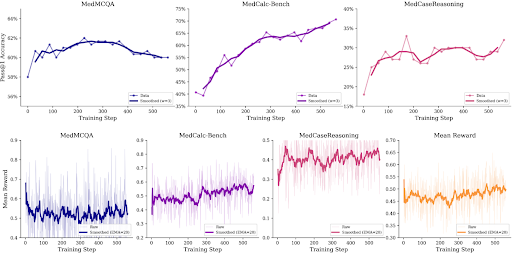}}
    \caption{Test accuracy and training reward for Qwen-3-4B-Instruct-0725 trained on MedCalc-Bench, MedMCQA, and MedCaseReasoning over the course of training for 330 steps.}
    \label{fig:post-training}
  \end{center}
\end{figure}

In \cref{fig:post-training}, we demonstrate preliminary post-training of Qwen-3-4B-Instruct-0725 on the combination of three different datasets (MedCalc-Bench, MedMCQA, and MedCaseReasoning). These datasets have different reward function types, such as a calculation verifier (MedCalc-Bench), multiple choice accuracy (MedMCQA), and LLM-as-a-Judge (MedCaseReasoning). Further details on RL training on these environments can be found in \cref{app:rl_training}. We leave it to future work to fully explore the potential of post-training with \textsc{Medmarks-T}.

\section{Conclusion}
\label{sec:future}

We present \textsc{Medmarks}, a comprehensive automated open-source LLM benchmarking suite for medical tasks. \textsc{Medmarks} includes a total of 30 benchmarks (divided into verifiable and open-ended subsets), evaluated over 61 models (proprietary and open-source, generalist and medically fine-tuned) across 71 reasoning and quantization configurations. We hope our benchmark suite brings us closer to real-world assessment of LLM medical capabilities in a more reproducible and accessible manner.

Our approach is not without limitations. Although \textsc{Medmarks} does include an open-ended subset and some agentic benchmarks, our benchmark is still heavily biased towards single-turn question answering tasks. Additionally, there is limited evaluation of fairness/bias and safety. \textsc{Medmarks} also only focuses on text-only tasks and does not evaluate multimodal medical capabilities. We hope to expand \textsc{Medmarks} and address these limitations in collaboration with the broader clinical AI community. 

\section*{Impact Statement}

This work introduces \textsc{Medmarks}, a benchmark suite intended to improve the reproducibility and transparency of measuring medical capabilities in LLMs. The use of AI in healthcare has many ethical and societal considerations, especially surrounding algorithmic fairness and biases. \textsc{Medmarks} is intended to make medical LLM evaluation more open, comparable, and analysis-driven, while explicitly acknowledging that benchmark performance is not equivalent to clinical competence and must be interpreted with care.

\section{Acknowledgements}

Thanks to FAL AI for providing compute that supported this research. Thanks to Prime Intellect for providing API inference credits. Thanks to the MedARC Discord community for being
the public forum from which this research was developed.

\bibliography{bibliography}

@techreport{brodeur2026state,
  title        = {State of Clinical AI 2026},
  author       = {Brodeur, Peter G. and Goh, Ethan and Tat, Emily and McCoy, Liam and Wu, David and Jain, Priyank and Handler, Rebecca and Hom, Jason and Zwaan, Laura and Ravi, Vishnu and Han, Brian and Schulman, Kevin and Lacar, Kathleen and Black, Kameron and Haimovich, Adrian and Horvitz, Eric and Rodman, Adam and Chen, Jonathan H.},
  institution  = {ARISE Network},
  year         = {2026},
  month        = jan,
  type         = {Technical Report}
}

@techreport{openai2026healthcare,
  title        = {AI as a Healthcare Ally: How Americans are Navigating the System with ChatGPT},
  author       = {{OpenAI}},
  institution  = {OpenAI},
  year         = {2026},
  month        = jan,
  url          = {https://cdn.openai.com/pdf/2cb29276-68cd-4ec6-a5f4-c01c5e7a36e9/OpenAI-AI-as-a-Healthcare-Ally-Jan-2026.pdf}
}

@inproceedings{kwon2023efficient,
  title={Efficient Memory Management for Large Language Model Serving with PagedAttention},
  author={Woosuk Kwon and Zhuohan Li and Siyuan Zhuang and Ying Sheng and Lianmin Zheng and Cody Hao Yu and Joseph E. Gonzalez and Hao Zhang and Ion Stoica},
  booktitle={Proceedings of the ACM SIGOPS 29th Symposium on Operating Systems Principles},
  year={2023}
}

@article{singhal2023large,
  title={Large language models encode clinical knowledge},
  author={Singhal, Karan and Azizi, Shekoofeh and Tu, Tao and Mahdavi, S Sara and Wei, Jason and Chung, Hyung Won and Scales, Nathan and Tanwani, Ajay and Cole-Lewis, Heather and Pfohl, Stephen and others},
  journal={Nature},
  volume={620},
  number={7972},
  pages={172--180},
  year={2023},
  publisher={Nature Publishing Group}
}

@article{bedi_holistic_2026,
  title = {Holistic Evaluation of Large Language Models for Medical Tasks with {{MedHELM}}},
  author = {Bedi, Suhana and Cui, Hejie and Fuentes, Miguel and Unell, Alyssa and Wornow, Michael and Banda, Juan M. and Kotecha, Nikesh and Keyes, Timothy and Mai, Yifan and Oez, Mert and Qiu, Hao and Jain, Shrey and Schettini, Leonardo and Kashyap, Mehr and Fries, Jason Alan and Swaminathan, Akshay and Chung, Philip and Haredasht, Fateme Nateghi and Lopez, Ivan and Aali, Asad and Tse, Gabriel and Nayak, Ashwin and Vedak, Shivam and Jain, Sneha S. and Patel, Birju and Fayanju, Oluseyi and Shah, Shreya and Goh, Ethan and Yao, Dong-han and Soetikno, Brian and Reis, Eduardo and Gatidis, Sergios and Divi, Vasu and Capasso, Robson and Saralkar, Rachna and Chiang, Chia-Chun and Jindal, Jenelle and Pham, Tho and Ghoddusi, Faraz and Lin, Steven and Chiou, Albert S. and Hong, Hyo Jung and Roy, Mohana and Gensheimer, Michael F. and Patel, Hinesh and Schulman, Kevin and Dash, Dev and Char, Danton and Downing, Lance and Grolleau, Francois and Black, Kameron and Mieso, Bethel and Zahedivash, Aydin and Yim, Wen-wai and Sharma, Harshita and Lee, Tony and Kirsch, Hannah and Lee, Jennifer and Ambers, Nerissa and Lugtu, Carlene and Sharma, Aditya and Mawji, Bilal and Alekseyev, Alex and Zhou, Vicky and Kakkar, Vikas and Helzer, Jarrod and Revri, Anurang and Bannett, Yair and Daneshjou, Roxana and Chen, Jonathan and Alsentzer, Emily and Morse, Keith and Ravi, Nirmal and Aghaeepour, Nima and Kennedy, Vanessa and Chaudhari, Akshay and Wang, Thomas and Koyejo, Sanmi and Lungren, Matthew P. and Horvitz, Eric and Liang, Percy and Pfeffer, Michael A. and Shah, Nigam H.},
  year = 2026,
  month = jan,
  journal = {Nature Medicine},
  issn = {1546-170X},
  doi = {10.1038/s41591-025-04151-2}
}

@article{arora2025healthbench,
  title={Healthbench: Evaluating large language models towards improved human health},
  author={Arora, Rahul K and Wei, Jason and Hicks, Rebecca Soskin and Bowman, Preston and Qui{\~n}onero-Candela, Joaquin and Tsimpourlas, Foivos and Sharman, Michael and Shah, Meghan and Vallone, Andrea and Beutel, Alex and others},
  journal={arXiv preprint arXiv:2505.08775},
  year={2025}
}

@article{kim2025limitations,
  title={Limitations of large language models in clinical problem-solving arising from inflexible reasoning},
  author={Kim, Jonathan and Podlasek, Anna and Shidara, Kie and Liu, Feng and Alaa, Ahmed and Bernardo, Danilo},
  journal={Scientific reports},
  volume={15},
  number={1},
  pages={39426},
  year={2025},
  publisher={Nature Publishing Group UK London}
}

@article{khandekar2024medcalc,
  title={Medcalc-bench: Evaluating large language models for medical calculations},
  author={Khandekar, Nikhil and Jin, Qiao and Xiong, Guangzhi and Dunn, Soren and Applebaum, Serina and Anwar, Zain and Sarfo-Gyamfi, Maame and Safranek, Conrad and Anwar, Abid and Zhang, Andrew and others},
  journal={Advances in Neural Information Processing Systems},
  volume={37},
  pages={84730--84745},
  year={2024}
}

@article{shoham2024medconceptsqa,
  title={MedConceptsQA: Open source medical concepts QA benchmark},
  author={Shoham, Ofir Ben and Rappoport, Nadav},
  journal={Computers in Biology and Medicine},
  volume={182},
  pages={109089},
  year={2024},
  publisher={Elsevier}
}

@inproceedings{pal2022medmcqa,
  title={Medmcqa: A large-scale multi-subject multi-choice dataset for medical domain question answering},
  author={Pal, Ankit and Umapathi, Logesh Kumar and Sankarasubbu, Malaikannan},
  booktitle={Conference on health, inference, and learning},
  pages={248--260},
  year={2022},
  organization={PMLR}
}

@article{jin2021disease,
  title={What disease does this patient have? a large-scale open domain question answering dataset from medical exams},
  author={Jin, Di and Pan, Eileen and Oufattole, Nassim and Weng, Wei-Hung and Fang, Hanyi and Szolovits, Peter},
  journal={Applied Sciences},
  volume={11},
  number={14},
  pages={6421},
  year={2021},
  publisher={MDPI}
}

@article{zuo2025medxpertqa,
  title={Medxpertqa: Benchmarking expert-level medical reasoning and understanding},
  author={Zuo, Yuxin and Qu, Shang and Li, Yifei and Chen, Zhangren and Zhu, Xuekai and Hua, Ermo and Zhang, Kaiyan and Ding, Ning and Zhou, Bowen},
  journal={arXiv preprint arXiv:2501.18362},
  year={2025}
}

@article{dou2025baichuan,
  title={Baichuan-m2: Scaling medical capability with large verifier system},
  author={Dou, Chengfeng and Liu, Chong and Yang, Fan and Li, Fei and Jia, Jiyuan and Chen, Mingyang and Ju, Qiang and Wang, Shuai and Dang, Shunya and Li, Tianpeng and others},
  journal={arXiv preprint arXiv:2509.02208},
  year={2025}
}

@article{sellergren2025medgemma,
  title={Medgemma technical report},
  author={Sellergren, Andrew and Kazemzadeh, Sahar and Jaroensri, Tiam and Kiraly, Atilla and Traverse, Madeleine and Kohlberger, Timo and Xu, Shawn and Jamil, Fayaz and Hughes, C{\'\i}an and Lau, Charles and others},
  journal={arXiv preprint arXiv:2507.05201},
  year={2025}
}

@article{yang2025qwen3,
  title={Qwen3 technical report},
  author={Yang, An and Li, Anfeng and Yang, Baosong and Zhang, Beichen and Hui, Binyuan and Zheng, Bo and Yu, Bowen and Gao, Chang and Huang, Chengen and Lv, Chenxu and others},
  journal={arXiv preprint arXiv:2505.09388},
  year={2025}
}

@misc{olmo2025olmo3,
      title={Olmo 3}, 
      author={Team Olmo and : and Allyson Ettinger and Amanda Bertsch and Bailey Kuehl and David Graham and David Heineman and Dirk Groeneveld and Faeze Brahman and Finbarr Timbers and Hamish Ivison and Jacob Morrison and Jake Poznanski and Kyle Lo and Luca Soldaini and Matt Jordan and Mayee Chen and Michael Noukhovitch and Nathan Lambert and Pete Walsh and Pradeep Dasigi and Robert Berry and Saumya Malik and Saurabh Shah and Scott Geng and Shane Arora and Shashank Gupta and Taira Anderson and Teng Xiao and Tyler Murray and Tyler Romero and Victoria Graf and Akari Asai and Akshita Bhagia and Alexander Wettig and Alisa Liu and Aman Rangapur and Chloe Anastasiades and Costa Huang and Dustin Schwenk and Harsh Trivedi and Ian Magnusson and Jaron Lochner and Jiacheng Liu and Lester James V. Miranda and Maarten Sap and Malia Morgan and Michael Schmitz and Michal Guerquin and Michael Wilson and Regan Huff and Ronan Le Bras and Rui Xin and Rulin Shao and Sam Skjonsberg and Shannon Zejiang Shen and Shuyue Stella Li and Tucker Wilde and Valentina Pyatkin and Will Merrill and Yapei Chang and Yuling Gu and Zhiyuan Zeng and Ashish Sabharwal and Luke Zettlemoyer and Pang Wei Koh and Ali Farhadi and Noah A. Smith and Hannaneh Hajishirzi},
      year={2025},
      eprint={2512.13961},
      archivePrefix={arXiv},
      primaryClass={cs.CL},
      url={https://arxiv.org/abs/2512.13961}, 
}

@article{zeng2025glm,
  title={Glm-4.5: Agentic, reasoning, and coding (arc) foundation models},
  author={Zeng, Aohan and Lv, Xin and Zheng, Qinkai and Hou, Zhenyu and Chen, Bin and Xie, Chengxing and Wang, Cunxiang and Yin, Da and Zeng, Hao and Zhang, Jiajie and others},
  journal={arXiv preprint arXiv:2508.06471},
  year={2025}
}

@inproceedings{jin2019pubmedqa,
  title={Pubmedqa: A dataset for biomedical research question answering},
  author={Jin, Qiao and Dhingra, Bhuwan and Liu, Zhengping and Cohen, William and Lu, Xinghua},
  booktitle={Proceedings of the 2019 conference on empirical methods in natural language processing and the 9th international joint conference on natural language processing (EMNLP-IJCNLP)},
  pages={2567--2577},
  year={2019}
}

@article{zheng2025empirical,
  title={An empirical study of qwen3 quantization},
  author={Zheng, Xingyu and Li, Yuye and Chu, Haoran and Feng, Yue and Ma, Xudong and Luo, Jie and Guo, Jinyang and Qin, Haotong and Magno, Michele and Liu, Xianglong},
  journal={arXiv preprint arXiv:2505.02214},
  year={2025}
}

@article{gu2025illusion,
  title={The illusion of readiness: Stress testing large frontier models on multimodal medical benchmarks},
  author={Gu, Yu and Fu, Jingjing and Liu, Xiaodong and Valanarasu, Jeya Maria Jose and Codella, Noel CF and Tan, Reuben and Liu, Qianchu and Jin, Ying and Zhang, Sheng and Wang, Jinyu and others},
  journal={arXiv preprint arXiv:2509.18234},
  year={2025}
}

@inproceedings{chen2025benchmarking,
  title={Benchmarking large language models on answering and explaining challenging medical questions},
  author={Chen, Hanjie and Fang, Zhouxiang and Singla, Yash and Dredze, Mark},
  booktitle={Proceedings of the 2025 Conference of the Nations of the Americas Chapter of the Association for Computational Linguistics: Human Language Technologies (Volume 1: Long Papers)},
  pages={3563--3599},
  year={2025}
}

@misc{brown_verifiers_2025,
  author       = {William Brown},
  title        = {{Verifiers}: Environments for LLM Reinforcement Learning},
  howpublished = {\url{https://github.com/PrimeIntellect-ai/verifiers}},
  year         = {2025}
}

@article{khattab2023dspy,
  title={Dspy: Compiling declarative language model calls into self-improving pipelines},
  author={Khattab, Omar and Singhvi, Arnav and Maheshwari, Paridhi and Zhang, Zhiyuan and Santhanam, Keshav and Vardhamanan, Sri and Haq, Saiful and Sharma, Ashutosh and Joshi, Thomas T and Moazam, Hanna and others},
  journal={arXiv preprint arXiv:2310.03714},
  year={2023}
}

@article{liang2022holistic,
  title={Holistic evaluation of language models},
  author={Liang, Percy and Bommasani, Rishi and Lee, Tony and Tsipras, Dimitris and Soylu, Dilara and Yasunaga, Michihiro and Zhang, Yian and Narayanan, Deepak and Wu, Yuhuai and Kumar, Ananya and others},
  journal={arXiv preprint arXiv:2211.09110},
  year={2022}
}

@article{gao2021framework,
  title={A framework for few-shot language model evaluation},
  author={Gao, Leo and Tow, Jonathan and Biderman, Stella and Black, Sid and DiPofi, Anthony and Foster, Charles and Golding, Laurence and Hsu, Jeffrey and McDonell, Kyle and Muennighoff, Niklas and others},
  journal={Zenodo},
  year={2021}
}

@article{pal2023med,
  title={Med-halt: Medical domain hallucination test for large language models},
  author={Pal, Ankit and Umapathi, Logesh Kumar and Sankarasubbu, Malaikannan},
  journal={arXiv preprint arXiv:2307.15343},
  year={2023}
}

@article{he2020meddialog,
  title={Meddialog: Two large-scale medical dialogue datasets},
  author={He, Xuehai and Chen, Shu and Ju, Zeqian and Dong, Xiangyu and Fang, Hongchao and Wang, Sicheng and Yang, Yue and Zeng, Jiaqi and Zhang, Ruisi and Zhang, Ruoyu and others},
  journal={arXiv preprint arXiv:2004.03329},
  year={2020}
}

@article{correa2025head,
  title={HEAD-QA v2: Expanding a Healthcare Benchmark for Reasoning},
  author={Correa-Guill{\'e}n, Alexis and G{\'o}mez-Rodr{\'\i}guez, Carlos and Vilares, David},
  journal={arXiv preprint arXiv:2511.15355},
  year={2025}
}

@article{krumdick2025no,
  title={No free labels: Limitations of llm-as-a-judge without human grounding},
  author={Krumdick, Michael and Lovering, Charles and Reddy, Varshini and Ebner, Seth and Tanner, Chris},
  journal={arXiv preprint arXiv:2503.05061},
  year={2025}
}

@article{yim2023aci,
  title={Aci-bench: a novel ambient clinical intelligence dataset for benchmarking automatic visit note generation},
  author={Yim, Wen-wai and Fu, Yujuan and Ben Abacha, Asma and Snider, Neal and Lin, Thomas and Yetisgen, Meliha},
  journal={Scientific data},
  volume={10},
  number={1},
  pages={586},
  year={2023},
  publisher={Nature Publishing Group UK London}
}

@article{schmidgall2024agentclinic,
  title={AgentClinic: a multimodal agent benchmark to evaluate AI in simulated clinical environments},
  author={Schmidgall, Samuel and Ziaei, Rojin and Harris, Carl and Reis, Eduardo and Jopling, Jeffrey and Moor, Michael},
  journal={arXiv preprint arXiv:2405.07960},
  year={2024}
}

@article{adams2025longhealth,
  title={Longhealth: A question answering benchmark with long clinical documents},
  author={Adams, Lisa and Busch, Felix and Han, Tianyu and Excoffier, Jean-Baptiste and Ortala, Matthieu and L{\"o}ser, Alexander and Aerts, Hugo JWL and Kather, Jakob Nikolas and Truhn, Daniel and Bressem, Keno},
  journal={Journal of Healthcare Informatics Research},
  pages={1--17},
  year={2025},
  publisher={Springer}
}

@article{wu2025medcasereasoning,
  title={MedCaseReasoning: Evaluating and learning diagnostic reasoning from clinical case reports},
  author={Wu, Kevin and Wu, Eric and Thapa, Rahul and Wei, Kevin and Zhang, Angela and Suresh, Arvind and Tao, Jacqueline J and Sun, Min Woo and Lozano, Alejandro and Zou, James},
  journal={arXiv preprint arXiv:2505.11733},
  year={2025}
}

@inproceedings{lin2004automatic,
  title={Automatic evaluation of machine translation quality using longest common subsequence and skip-bigram statistics},
  author={Lin, Chin-Yew and Och, Franz Josef},
  booktitle={Proceedings of the 42nd annual meeting of the association for computational linguistics (ACL-04)},
  pages={605--612},
  year={2004}
}

@article{zhang2019bertscore,
  title={Bertscore: Evaluating text generation with bert},
  author={Zhang, Tianyi and Kishore, Varsha and Wu, Felix and Weinberger, Kilian Q and Artzi, Yoav},
  journal={arXiv preprint arXiv:1904.09675},
  year={2019}
}

@misc{gu2025surveyllmasajudge,
      title={A Survey on LLM-as-a-Judge}, 
      author={Jiawei Gu and Xuhui Jiang and Zhichao Shi and Hexiang Tan and Xuehao Zhai and Chengjin Xu and Wei Li and Yinghan Shen and Shengjie Ma and Honghao Liu and Saizhuo Wang and Kun Zhang and Yuanzhuo Wang and Wen Gao and Lionel Ni and Jian Guo},
      year={2025},
      eprint={2411.15594},
      archivePrefix={arXiv},
      primaryClass={cs.CL},
      url={https://arxiv.org/abs/2411.15594}, 
}

@incollection{abacha2019bridging,
  title={Bridging the gap between consumers’ medication questions and trusted answers},
  author={Abacha, Asma Ben and Mrabet, Yassine and Sharp, Mark and Goodwin, Travis R and Shooshan, Sonya E and Demner-Fushman, Dina},
  booktitle={MEDINFO 2019: Health and Wellbeing e-Networks for All},
  pages={25--29},
  year={2019},
  publisher={IOS Press}
}

@article{Lehman2023DoWS,
  title={Do We Still Need Clinical Language Models?},
  author={Eric P. Lehman and Evan Hernandez and Diwakar Mahajan and Jonas Wulff and Micah J. Smith and Zachary M. Ziegler and Daniel Nadler and Peter Szolovits and Alistair E. W. Johnson and Emily Alsentzer},
  journal={ArXiv},
  year={2023},
  volume={abs/2302.08091}
}

@misc{nori2023generalistfoundationmodelsoutcompete,
      title={Can Generalist Foundation Models Outcompete Special-Purpose Tuning? Case Study in Medicine}, 
      author={Harsha Nori and Yin Tat Lee and Sheng Zhang and Dean Carignan and Richard Edgar and Nicolo Fusi and Nicholas King and Jonathan Larson and Yuanzhi Li and Weishung Liu and Renqian Luo and Scott Mayer McKinney and Robert Osazuwa Ness and Hoifung Poon and Tao Qin and Naoto Usuyama and Chris White and Eric Horvitz},
      year={2023},
      eprint={2311.16452},
      archivePrefix={arXiv},
      primaryClass={cs.CL},
      url={https://arxiv.org/abs/2311.16452}, 
}

@misc{lin2024awqactivationawareweightquantization,
      title={AWQ: Activation-aware Weight Quantization for LLM Compression and Acceleration}, 
      author={Ji Lin and Jiaming Tang and Haotian Tang and Shang Yang and Wei-Ming Chen and Wei-Chen Wang and Guangxuan Xiao and Xingyu Dang and Chuang Gan and Song Han},
      year={2024},
      eprint={2306.00978},
      archivePrefix={arXiv},
      primaryClass={cs.CL},
      url={https://arxiv.org/abs/2306.00978}, 
}

@article{griot_large_2025,
    title = {Large {Language} {Models} lack essential metacognition for reliable medical reasoning},
    volume = {16},
    issn = {2041-1723},
    url = {https://doi.org/10.1038/s41467-024-55628-6},
    doi = {10.1038/s41467-024-55628-6},
    number = {1},
    journal = {Nature Communications},
    author = {Griot, Maxime and Hemptinne, Coralie and Vanderdonckt, Jean and Yuksel, Demet},
    month = jan,
    year = {2025},
    pages = {642},
}

@misc{abacha2025medecbenchmarkmedicalerror,
      title={MEDEC: A Benchmark for Medical Error Detection and Correction in Clinical Notes}, 
      author={Asma Ben Abacha and Wen-wai Yim and Yujuan Fu and Zhaoyi Sun and Meliha Yetisgen and Fei Xia and Thomas Lin},
      year={2025},
      eprint={2412.19260},
      archivePrefix={arXiv},
      primaryClass={cs.CL},
      url={https://arxiv.org/abs/2412.19260}, 
}

@article{mccoy2025assessment,
  title={Assessment of large language models in clinical reasoning: a novel benchmarking study},
  author={McCoy, Liam G and Swamy, Rajiv and Sagar, Nidhish and Wang, Minjia and Bacchi, Stephen and Fong, Jie Ming Nigel and Tan, Nigel CK and Tan, Kevin and Buckley, Thomas A and Brodeur, Peter and others},
  journal={NEJM AI},
  volume={2},
  number={10},
  pages={AIdbp2500120},
  year={2025},
  publisher={Massachusetts Medical Society}
}

@article{griot_implementation_2025,
    title = {Implementation of large language models in electronic health records},
    volume = {4},
    url = {https://doi.org/10.1371/journal.pdig.0001141},
    doi = {10.1371/journal.pdig.0001141},
    number = {12},
    journal = {PLOS Digital Health},
    publisher = {Public Library of Science},
    author = {Griot, Maxime and Vanderdonckt, Jean and Yuksel, Demet},
    month = dec,
    year = {2025},
    pages = {1--18},
}

@misc{pandit2025medhallucomprehensivebenchmarkdetecting,
  title={MedHallu: A Comprehensive Benchmark for Detecting Medical Hallucinations in Large Language Models}, 
  author={Shrey Pandit and Jiawei Xu and Junyuan Hong and Zhangyang Wang and Tianlong Chen and Kaidi Xu and Ying Ding},
  year={2025},
  eprint={2502.14302},
  archivePrefix={arXiv},
  primaryClass={cs.CL},
  url={https://arxiv.org/abs/2502.14302},
}

@article{wang2024mmlu,
  title={Mmlu-pro: A more robust and challenging multi-task language understanding benchmark},
  author={Wang, Yubo and Ma, Xueguang and Zhang, Ge and Ni, Yuansheng and Chandra, Abhranil and Guo, Shiguang and Ren, Weiming and Arulraj, Aaran and He, Xuan and Jiang, Ziyan and others},
  journal={arXiv preprint arXiv:2406.01574},
  year={2024}
}

@misc{kim2025limitationslargelanguagemodels, 
      title={Limitations of Large Language Models in Clinical Problem-Solving Arising from Inflexible Reasoning}, 
      author={Jonathan Kim and Anna Podlasek and Kie Shidara and Feng Liu and Ahmed Alaa and Danilo Bernardo},
      year={2025},
      eprint={2502.04381},
      archivePrefix={arXiv},
      primaryClass={cs.CL},
      url={https://arxiv.org/abs/2502.04381}, 
}

@inproceedings{chen-etal-2025-benchmarking,
    title = "Benchmarking Large Language Models on Answering and Explaining Challenging Medical Questions",
    author = "Chen, Hanjie  and
      Fang, Zhouxiang  and
      Singla, Yash  and
      Dredze, Mark",
    editor = "Chiruzzo, Luis  and
      Ritter, Alan  and
      Wang, Lu",
    booktitle = "Proceedings of the 2025 Conference of the Nations of the Americas Chapter of the Association for Computational Linguistics: Human Language Technologies (Volume 1: Long Papers)",
    month = apr,
    year = "2025",
    address = "Albuquerque, New Mexico",
    publisher = "Association for Computational Linguistics",
    url = "https://aclanthology.org/2025.naacl-long.182/",
    doi = "10.18653/v1/2025.naacl-long.182",
    pages = "3563--3599",
    ISBN = "979-8-89176-189-6"
}

@misc{pteam2025supergpqascalingllmevaluation,
      title={SuperGPQA: Scaling LLM Evaluation across 285 Graduate Disciplines},
      author={M-A-P Team and Xinrun Du and Yifan Yao and Kaijing Ma and Bingli Wang and Tianyu Zheng and Kang Zhu and Minghao Liu and Yiming Liang and Xiaolong Jin and Zhenlin Wei and Chujie Zheng and Kaixin Deng and Shian Jia and Sichao Jiang and Yiyan Liao and Rui Li and Qinrui Li and Sirun Li and Yizhi Li and Yunwen Li and Dehua Ma and Yuansheng Ni and Haoran Que and Qiyao Wang and Zhoufutu Wen and Siwei Wu and Tianshun Xing and Ming Xu and Zhenzhu Yang and Zekun Moore Wang and Junting Zhou and Yuelin Bai and Xingyuan Bu and Chenglin Cai and Liang Chen and Yifan Chen and Chengtuo Cheng and Tianhao Cheng and Keyi Ding and Siming Huang and Yun Huang and Yaoru Li and Yizhe Li and Zhaoqun Li and Tianhao Liang and Chengdong Lin and Hongquan Lin and Yinghao Ma and Tianyang Pang and Zhongyuan Peng and Zifan Peng and Qige Qi and Shi Qiu and Xingwei Qu and Shanghaoran Quan and Yizhou Tan and Zili Wang and Chenqing Wang and Hao Wang and Yiya Wang and Yubo Wang and Jiajun Xu and Kexin Yang and Ruibin Yuan and Yuanhao Yue and Tianyang Zhan and Chun Zhang and Jinyang Zhang and Xiyue Zhang and Xingjian Zhang and Yue Zhang and Yongchi Zhao and Xiangyu Zheng and Chenghua Zhong and Yang Gao and Zhoujun Li and Dayiheng Liu and Qian Liu and Tianyu Liu and Shiwen Ni and Junran Peng and Yujia Qin and Wenbo Su and Guoyin Wang and Shi Wang and Jian Yang and Min Yang and Meng Cao and Xiang Yue and Zhaoxiang Zhang and Wangchunshu Zhou and Jiaheng Liu and Qunshu Lin and Wenhao Huang and Ge Zhang},
      year={2025},
      eprint={2502.14739},
      archivePrefix={arXiv},
      primaryClass={cs.CL},
      url={https://arxiv.org/abs/2502.14739}, 
}

@misc{harris2025pubhealthbench,
  title = {Healthy LLMs? Benchmarking LLM Knowledge of UK Government Public Health Information},
  author = {Harris, Joshua and Grayson, Fan and Feldman, Felix and Laurence, Timothy and Nonnenmacher, Toby and Higgins, Oliver and Loman, Leo and Patel, Selina and Finnie, Thomas and Collins, Samuel and Borowitz, Michael},
  year = {2025},
  eprint = {2505.06046},
  archivePrefix = {arXiv},
  primaryClass = {cs.CL},
  note = {Dataset and paper introduce the PubHealthBench benchmark},
  url = {https://arxiv.org/abs/2505.06046}
}

@misc{kim2024medexqamedicalquestionanswering,
      title={MedExQA: Medical Question Answering Benchmark with Multiple Explanations}, 
      author={Yunsoo Kim and Jinge Wu and Yusuf Abdulle and Honghan Wu},
      year={2024},
      eprint={2406.06331},
      archivePrefix={arXiv},
      primaryClass={cs.CL},
      url={https://arxiv.org/abs/2406.06331}, 
}

@inproceedings{benabacha2019medicationqa,
  author    = {Asma Ben Abacha and Yassine Mrabet and Mark Sharp and
               Travis Goodwin and Sonya E. Shooshan and Dina Demner{-}Fushman},
  title     = {Bridging the Gap between Consumers' Medication Questions and Trusted Answers},
  booktitle = {MEDINFO 2019},
  year      = {2019},
  doi       = {10.3233/SHTI190176}
}

@misc{bai2025qwen3vltechnicalreport,
      title={Qwen3-VL Technical Report}, 
      author={Shuai Bai and Yuxuan Cai and Ruizhe Chen and Keqin Chen and Xionghui Chen and Zesen Cheng and Lianghao Deng and Wei Ding and Chang Gao and Chunjiang Ge and Wenbin Ge and Zhifang Guo and Qidong Huang and Jie Huang and Fei Huang and Binyuan Hui and Shutong Jiang and Zhaohai Li and Mingsheng Li and Mei Li and Kaixin Li and Zicheng Lin and Junyang Lin and Xuejing Liu and Jiawei Liu and Chenglong Liu and Yang Liu and Dayiheng Liu and Shixuan Liu and Dunjie Lu and Ruilin Luo and Chenxu Lv and Rui Men and Lingchen Meng and Xuancheng Ren and Xingzhang Ren and Sibo Song and Yuchong Sun and Jun Tang and Jianhong Tu and Jianqiang Wan and Peng Wang and Pengfei Wang and Qiuyue Wang and Yuxuan Wang and Tianbao Xie and Yiheng Xu and Haiyang Xu and Jin Xu and Zhibo Yang and Mingkun Yang and Jianxin Yang and An Yang and Bowen Yu and Fei Zhang and Hang Zhang and Xi Zhang and Bo Zheng and Humen Zhong and Jingren Zhou and Fan Zhou and Jing Zhou and Yuanzhi Zhu and Ke Zhu},
      year={2025},
      eprint={2511.21631},
      archivePrefix={arXiv},
      primaryClass={cs.CV},
      url={https://arxiv.org/abs/2511.21631}, 
}

@misc{ariasduart2025automaticevaluationhealthcarellms,
      title={Automatic Evaluation of Healthcare LLMs Beyond Question-Answering}, 
      author={Anna Arias-Duart and Pablo Agustin Martin-Torres and Daniel Hinjos and Pablo Bernabeu-Perez and Lucia Urcelay Ganzabal and Marta Gonzalez Mallo and Ashwin Kumar Gururajan and Enrique Lopez-Cuena and Sergio Alvarez-Napagao and Dario Garcia-Gasulla},
      year={2025},
      eprint={2502.06666},
      archivePrefix={arXiv},
      primaryClass={cs.CL},
      url={https://arxiv.org/abs/2502.06666}, 
}

@article{qiu2025quantifying,
  title={Quantifying the reasoning abilities of LLMs on real-world clinical cases},
  author={Qiu, Pengcheng and Wu, Chaoyi and Liu, Shuyu and Zhao, Weike and Chen, Zhuoxia and Gu, Hongfei and Peng, Chuanjin and Zhang, Ya and Wang, Yanfeng and Xie, Weidi},
  journal={arXiv preprint arXiv:2503.04691},
  year={2025}
}

@inproceedings{arias-duart-etal-2025-automatic,
    title = "Automatic Evaluation of Healthcare {LLM}s Beyond Question-Answering",
    author = "Arias-Duart, Anna and Martin-Torres, Pablo Agustin and Hinjos, Daniel and Bernabeu-Perez, Pablo and Ganzabal, Lucia Urcelay and Mallo, Marta Gonzalez and Gururajan, Ashwin Kumar and Lopez-Cuena, Enrique and Alvarez-Napagao, Sergio and Garcia-Gasulla, Dario",
    booktitle = "Proceedings of NAACL 2025",
    year = "2025",
    pages = "108--130"
}

@article{metlay_diagnosis_2019,
  title = {Diagnosis and {{Treatment}} of {{Adults}} with {{Community-acquired Pneumonia}}. {{An Official Clinical Practice Guideline}} of the {{American Thoracic Society}} and {{Infectious}}  {{Diseases Society}} of {{America}}.},
  author = {Metlay, Joshua P. and Waterer, Grant W. and Long, Ann C. and Anzueto, Antonio and Brozek, Jan and Crothers, Kristina and Cooley, Laura A. and Dean, Nathan C. and Fine, Michael J. and Flanders, Scott A. and Griffin, Marie R. and Metersky, Mark L. and Musher, Daniel M. and Restrepo, Marcos I. and Whitney, Cynthia G.},
  year = 2019,
  month = oct,
  journal = {American journal of respiratory and critical care medicine},
  volume = {200},
  number = {7},
  pages = {e45-e67},
  address = {United States},
  issn = {1535-4970 1073-449X},
  doi = {10.1164/rccm.201908-1581ST},
  langid = {english},
  pmcid = {PMC6812437},
  pmid = {31573350}
}

@article{mocanu_antibiotic_2019,
  title = {Antibiotic Use in Prevention of Anal Fistulas Following Incision and Drainage of Anorectal Abscesses: {{A}} Systematic Review and Meta-Analysis.},
  author = {Mocanu, Valentin and Dang, Jerry T. and Ladak, Farah and Tian, Chunhong and Wang, Haili and Birch, Daniel W. and Karmali, Shahzeer},
  year = 2019,
  month = may,
  journal = {American journal of surgery},
  volume = {217},
  number = {5},
  pages = {910--917},
  address = {United States},
  issn = {1879-1883 0002-9610},
  doi = {10.1016/j.amjsurg.2019.01.015},
  copyright = {Copyright \copyright{} 2019 Elsevier Inc. All rights reserved.},
  langid = {english},
  pmid = {30773213}
}

@misc{arcee2025afm45b,
  title = {AFM-4.5B: The First Arcee Foundation Model},
  author = {McQuade, Mark and Fernandes Neto, Fernando and Singh, Varun and Goddard, Charles and Atkins, Lucas and {Arcee AI Team}},
  year = {2025},
  publisher = {Arcee AI},
  url = {https://www.arcee.ai/blog/deep-dive-afm-4-5b-the-first-arcee-foundational-model},
}

@misc{antangelmed_2025,
  title = {AntAngelMed: A High-Performance Medical Language Model with Efficient MoE-Powered Clinical Reasoning},
  author = {AntAngelMed Team},
  year = {2025},
  url = {https://huggingface.co/MedAIBase/AntAngelMed},
}

@misc{m2team2025baichuanm2scalingmedicalcapability,
  title = {Baichuan-M2: Scaling Medical Capability with Large Verifier System},
  author = {M2 Team and Chengfeng Dou and Chong Liu and Fan Yang and Fei Li and Jiyuan Jia and Mingyang Chen and Qiang Ju and Shuai Wang and Shunya Dang and Tianpeng Li and Xiangrong Zeng and Yijie Zhou and Chenzheng Zhu and Da Pan and Fei Deng and Guangwei Ai and Guosheng Dong and Hongda Zhang and Jinyang Tai and Jixiang Hong and Kai Lu and Linzhuang Sun and Peidong Guo and Qian Ma and Rihui Xin and Shihui Yang and Shusen Zhang and Yichuan Mo and Zheng Liang and Zhishou Zhang and Hengfu Cui and Zuyi Zhu and Xiaochuan Wang},
  year = {2025},
  eprint = {2509.02208},
  archivePrefix = {arXiv},
  primaryClass = {cs.LG},
  url = {https://arxiv.org/abs/2509.02208},
}

@misc{baichuan-m3,
  title = {Baichuan-M3: Modeling Clinical Inquiry for Reliable Medical Decision-Making},
  author = {Baichuan M3 Team},
  year = {2025},
  url = {https://github.com/baichuan-inc/Baichuan-M3-235B},
}

@misc{anthropic_introducing_sonnet_2025,
  title = {Introducing {Claude Sonnet} 4.5},
  author = {Anthropic},
  year = {2025},
  month = sep,
  howpublished = {https://www.anthropic.com/news/claude-sonnet-4-5},
  urldate = {2026-01-22},
}

@misc{gemmateam2025gemma3technicalreport,
  title = {Gemma 3 Technical Report},
  author = {Gemma Team and Aishwarya Kamath and Johan Ferret and Shreya Pathak and Nino Vieillard and Ramona Merhej and Sarah Perrin and Tatiana Matejovicova and Alexandre Ram{\'e} and Morgane Rivi{\`e}re and Louis Rouillard and Thomas Mesnard and Geoffrey Cideron and Jean-bastien Grill and Sabela Ramos and Edouard Yvinec and Michelle Casbon and Etienne Pot and Ivo Penchev and Ga{\"e}l Liu and Francesco Visin and Kathleen Kenealy and Lucas Beyer and Xiaohai Zhai and Anton Tsitsulin and Robert Busa-Fekete and Alex Feng and Noveen Sachdeva and Benjamin Coleman and Yi Gao and Basil Mustafa and Iain Barr and Emilio Parisotto and David Tian and Matan Eyal and Colin Cherry and others},
  year = {2025},
  eprint = {2503.19786},
  archivePrefix = {arXiv},
  primaryClass = {cs.CL},
  url = {https://arxiv.org/abs/2503.19786},
}

@misc{glmteam2025glm45agenticreasoningcoding,
  title = {GLM-4.5: Agentic, Reasoning, and Coding (ARC) Foundation Models},
  author = {GLM Team and Aohan Zeng and Xin Lv and Qinkai Zheng and Zhenyu Hou and Bin Chen and Chengxing Xie and Cunxiang Wang and Da Yin and Hao Zeng and Jiajie Zhang and Kedong Wang and Lucen Zhong and Mingdao Liu and Rui Lu and Shulin Cao and Xiaohan Zhang and Xuancheng Huang and Yao Wei and Yean Cheng and Yifan An and Yilin Niu and Yuanhao Wen and Yushi Bai and Zhengxiao Du and Zihan Wang and Zilin Zhu and others},
  year = {2025},
  url = {https://arxiv.org/abs/2503.xxxxx},
}

@misc{openai_gpt5.2_systemcard_2025,
  author = {{OpenAI}},
  title = {Update to GPT-5 System Card: GPT-5.2},
  year = {2025},
  howpublished = {Technical report (PDF)},
  url = {https://cdn.openai.com/pdf/3a4153c8-c748-4b71-8e31-aecbde944f8d/oai_5_2_system-card.pdf},
  note = {Accessed: 2026-01-22},
}

@misc{openai_gpt5.1_systemcard_2025,
  author = {{OpenAI}},
  title = {GPT-5.1 Instant and GPT-5.1 Thinking System Card Addendum},
  year = {2025},
  howpublished = {Technical report (PDF)},
  url = {https://cdn.openai.com/pdf/4173ec8d-1229-47db-96de-06d87147e07e/5_1_system_card.pdf},
  note = {Accessed: 2026-01-22},
}

@misc{openai_gpt5_systemcard_2025,
  author = {{OpenAI}},
  title = {GPT-5 System Card},
  year = {2025},
  howpublished = {Technical report (PDF)},
  url = {https://cdn.openai.com/gpt-5-system-card.pdf},
  note = {Includes reference to GPT-5 mini variants; accessed: 2026-01-22},
}

@misc{openai_gpt-oss-120b_2025,
  title = {GPT-OSS-120B and GPT-OSS-20B Model Card},
  author = {OpenAI and Agarwal, Sandhini and Ahmad, Lama and Ai, Jason and Altman, Sam and Applebaum, Andy and Arbus, Edwin and Arora, Rahul K. and Bai, Yu and Baker, Bowen and Bao, Haiming and Barak, Boaz and Bennett, Ally and Bertao, Tyler and Brett, Nivedita and Brevdo, Eugene and Brockman, Greg and Bubeck, Sebastien and Chang, Che and Chen, Kai and Chen, Mark and Cheung, Enoch and Clark, Aidan and Cook, Dan and Dukhan, Marat and Dvorak, Casey and Fives, Kevin and Fomenko, Vlad and Garipov, Timur and Georgiev, Kristian and others},
  year = {2025},
  month = aug,
  eprint = {2508.10925},
  archivePrefix = {arXiv},
  primaryClass = {cs},
  doi = {10.48550/arXiv.2508.10925},
}

@misc{ibm_granite4_overview_2025,
  author = {{IBM}},
  title = {Granite 4.0: Hyper-efficient, High-performance Hybrid Language Models},
  year = {2025},
  url = {https://www.ibm.com/new/announcements/ibm-granite-4-0-hyper-efficient-high-performance-hybrid-models},
  note = {Accessed: 2026-01-22},
}

@misc{xai_grok4_model_card_2025,
  author = {{xAI}},
  title = {Grok 4 Model Card},
  year = {2025},
  howpublished = {Technical model card (PDF)},
  url = {https://data.x.ai/2025-08-20-grok-4-model-card.pdf},
  note = {Accessed: 2026-01-22},
}

@misc{xai_grok4.1_model_card_2025,
  author = {{xAI}},
  title = {Grok 4.1 Model Card},
  year = {2025},
  month = nov,
  howpublished = {Technical model card (PDF)},
  url = {https://data.x.ai/2025-11-17-grok-4-1-model-card.pdf},
  note = {Accessed: 2026-01-22},
}

@misc{teknium_hermes_2025,
  title = {Hermes 4 Technical Report},
  author = {Teknium, Ryan and Jin, Roger and Suphavadeeprasit, Jai and Mahan, Dakota and Quesnelle, Jeffrey and Li, Joe and Guang, Chen and Sands, Shannon and Malhotra, Karan},
  year = {2025},
  month = sep,
  eprint = {2508.18255},
  archivePrefix = {arXiv},
  primaryClass = {cs},
  doi = {10.48550/arXiv.2508.18255},
}

@misc{primeintellectteam2025intellect3technicalreport,
  title = {INTELLECT-3: Technical Report},
  author = {Prime Intellect Team and Mika Senghaas and Fares Obeid and Sami Jaghouar and William Brown and Jack Min Ong and Daniel Auras and Matej Sirovatka and Jannik Straube and Andrew Baker and Sebastian M{\"u}ller and Justus Mattern and Manveer Basra and Aiman Ismail and Dominik Scherm and Cooper Miller and Ameen Patel and Simon Kirsten and Mario Sieg and Christian Reetz and Kemal Erdem and Vincent Weisser and Johannes Hagemann},
  year = {2025},
  eprint = {2512.16144},
  archivePrefix = {arXiv},
  primaryClass = {cs.LG},
  url = {https://arxiv.org/abs/2512.16144},
}

@misc{ai21labs_jamba2_blog_2026,
  author = {{AI21 Labs}},
  title = {Introducing Jamba2: The open source model family for enterprise reliability and efficiency},
  year = {2026},
  month = jan,
  url = {https://www.ai21.com/blog/introducing-jamba2/},
  note = {Accessed: 2026-01-22},
}

@article{grattafiori2024llama,
  title = {The Llama 3 Herd of Models},
  author = {Grattafiori, Aaron and Dubey, Abhimanyu and Jauhri, Abhinav and Pandey, Abhinav and Kadian, Abhishek and Al-Dahle, Ahmad and Letman, Aiesha and Mathur, Akhil and Schelten, Alan and Vaughan, Alex and others},
  journal = {arXiv preprint arXiv:2407.21783},
  year = {2024},
}

@misc{lingteam2025activationboostedscalinggeneral,
  title = {Every Activation Boosted: Scaling General Reasoner to 1 Trillion Open Language Foundation},
  author = {Ling Team and Ang Li and Ben Liu and Binbin Hu and Bing Li and Bingwei Zeng and Borui Ye and Caizhi Tang and Changxin Tian and Chao Huang and Chao Zhang and Chen Qian and Chenchen Ju and Chenchen Li and Chengfu Tang and Chilin Fu and Chunshao Ren and Chunwei Wu and Cong Zhang and Cunyin Peng and others},
  year = {2025},
  eprint = {2510.22115},
  archivePrefix = {arXiv},
  primaryClass = {cs.CL},
  url = {https://arxiv.org/abs/2510.22115},
}

@misc{mistralai2025magistral,
  title = {Magistral},
  author = {Mistral-AI and Abhinav Rastogi and Albert Q. Jiang and Andy Lo and Gabrielle Berrada and Guillaume Lample and Jason Rute and Joep Barmentlo and Karmesh Yadav and Kartik Khandelwal and Khyathi Raghavi Chandu and L{\'e}onard Blier and Lucile Saulnier and Matthieu Dinot and Maxime Darrin and Neha Gupta and Roman Soletskyi and Sagar Vaze and Teven Le Scao and Yihan Wang and others},
  year = {2025},
  eprint = {2506.10910},
  archivePrefix = {arXiv},
  primaryClass = {cs.CL},
  url = {https://arxiv.org/abs/2506.10910},
}

@misc{sellergren_medgemma_2025,
  title = {MedGemma Technical Report},
  author = {Sellergren, Andrew and Kazemzadeh, Sahar and Jaroensri, Tiam and Kiraly, Atilla and Traverse, Madeleine and Kohlberger, Timo and Xu, Shawn and Jamil, Fayaz and Hughes, C{\'i}an and Lau, Charles and Chen, Justin and Mahvar, Fereshteh and Yatziv, Liron and Chen, Tiffany and Sterling, Bram and Baby, Stefanie Anna and Baby, Susanna Maria and Lai, Jeremy and Schmidgall, Samuel and Yang, Lu and others},
  year = {2025},
  month = jul,
  eprint = {2507.05201},
  archivePrefix = {arXiv},
  doi = {10.48550/arXiv.2507.05201},
}

@misc{golden_next_2026,
  title = {Next Generation Medical Image Interpretation with {MedGemma} 1.5 and Medical Speech to Text with {MedASR}},
  author = {Golden, Daniel and Mahvar, Fereshteh},
  year = {2026},
  month = jan,
  howpublished = {https://research.google/blog/next-generation-medical-image-interpretation-with-medgemma-15-and-medical-speech-to-text-with-medasr/},
}

@misc{minimax_m2.1_minimaxio_2025,
  author = {{MiniMax AI}},
  title = {MiniMax M2.1: Significantly Enhanced Multi-Language Programming, Built for Real-World Complex Tasks},
  year = {2025},
  month = dec,
  howpublished = {Official announcement (MiniMax.io)},
  url = {https://www.minimax.io/news/minimax-m21},
  note = {Accessed: 2026-01-22},
}

@misc{liu2026ministral3,
  title = {Ministral 3},
  author = {Alexander H. Liu and Kartik Khandelwal and Sandeep Subramanian and Victor Jouault and Abhinav Rastogi and Adrien Sad{\'e} and Alan Jeffares and Albert Jiang and Alexandre Cahill and Alexandre Gavaudan and Alexandre Sablayrolles and Am{\'e}lie H{\'e}liou and Amos You and Andy Ehrenberg and Andy Lo and Anton Eliseev and Antonia Calvi and Avinash Sooriyarachchi and Baptiste Bout and Baptiste Rozi{\`e}re and Baudouin De Monicault and others},
  year = {2026},
  eprint = {2601.08584},
  archivePrefix = {arXiv},
  primaryClass = {cs.CL},
  url = {https://arxiv.org/abs/2601.08584},
}

@article{miromind2025mirothinker,
  title = {MiroThinker: Pushing the Performance Boundaries of Open-Source Research Agents via Model, Context, and Interactive Scaling},
  author = {MiroMind Team and Bai, Song and Bing, Lidong and Chen, Carson and Chen, Guanzheng and Chen, Yuntao and Chen, Zhe and Chen, Ziyi and Dong, Xuan and others},
  journal = {arXiv preprint arXiv:2511.11793},
  year = {2025},
}

@misc{nvidia2025nvidianemotronnano2,
  title = {NVIDIA Nemotron Nano 2: An Accurate and Efficient Hybrid Mamba-Transformer Reasoning Model},
  author = {NVIDIA and Aarti Basant and Abhijit Khairnar and Abhijit Paithankar and Abhinav Khattar and Adithya Renduchintala and Aditya Malte and Akhiad Bercovich and Akshay Hazare and Alejandra Rico and Aleksander Ficek and Alex Kondratenko and Alex Shaposhnikov and Alexander Bukharin and Ali Taghibakhshi and others},
  year = {2025},
  eprint = {2508.14444},
  archivePrefix = {arXiv},
  primaryClass = {cs.CL},
  url = {https://arxiv.org/abs/2508.14444},
}

@misc{nvidia2025nemotron3nanoopen,
  title = {Nemotron 3 Nano: Open, Efficient Mixture-of-Experts Hybrid Mamba-Transformer Model for Agentic Reasoning},
  author = {NVIDIA and Aaron Blakeman and Aaron Grattafiori and Aarti Basant and Abhibha Gupta and Abhinav Khattar and Adi Renduchintala and Aditya Vavre and Akanksha Shukla and Akhiad Bercovich and Aleksander Ficek and Aleksandr Shaposhnikov and Alex Kondratenko and Alexander Bukharin and Alexandre Milesi and others},
  year = {2025},
  eprint = {2512.20848},
  archivePrefix = {arXiv},
  primaryClass = {cs.CL},
  url = {https://arxiv.org/abs/2512.20848},
}

@misc{abdin_phi-4-reasoning_2025,
  title = {Phi-4-Reasoning Technical Report},
  author = {Abdin, Marah and Agarwal, Sahaj and Awadallah, Ahmed and Balachandran, Vidhisha and Behl, Harkirat and Chen, Lingjiao and de Rosa, Gustavo and Gunasekar, Suriya and Javaheripi, Mojan and Joshi, Neel and Kauffmann, Piero and Lara, Yash and Mendes, Caio C{\'e}sar Teodoro and Mitra, Arindam and Nushi, Besmira and Papailiopoulos, Dimitris and Saarikivi, Olli and Shah, Shital and Shrivastava, Vaishnavi and Vineet, Vibhav and Wu, Yue and Yousefi, Safoora and Zheng, Guoqing},
  year = {2025},
  month = apr,
  eprint = {2504.21318},
  archivePrefix = {arXiv},
  primaryClass = {cs},
  doi = {10.48550/arXiv.2504.21318},
}

@misc{qwenteam_qwen3-next_2025,
  title = {Qwen3-Next: Towards Ultimate Training and Inference Efficiency},
  author = {QwenTeam},
  year = {2025},
  month = sep,
  howpublished = {https://qwen.ai/blog?id=4074cca80393150c248e508aa62983f9cb7d27cd},
}

@misc{bakouch2025smollm3,
  title = {SmolLM3: smol, multilingual, long-context reasoner},
  author = {Bakouch, Elie and Ben Allal, Loubna and Lozhkov, Anton and Tazi, Nouamane and Tunstall, Lewis and Pati{\~n}o, Carlos Miguel and Beeching, Edward and Roucher, Aymeric and Reedi, Aksel Joonas and Gallou{\'e}dec, Quentin and Rasul, Kashif and Habib, Nathan and Fourrier, Cl{\'e}mentine and Kydlicek, Hynek and Penedo, Guilherme and Larcher, Hugo and Morlon, Mathieu and Srivastav, Vaibhav and Lochner, Joshua and Nguyen, Xuan-Son and Raffel, Colin and von Werra, Leandro and Wolf, Thomas},
  year = {2025},
  howpublished = {\url{https://huggingface.co/blog/smollm3}},
}

@misc{arcee2025trinitymanifesto,
  title = {The Trinity Manifesto: Arcee Introduces Trinity Mini and Trinity Nano Preview},
  author = {Atkins, Lucas and {Arcee AI Team}},
  year = {2025},
  publisher = {Arcee AI},
  url = {https://www.arcee.ai/blog/the-trinity-manifesto},
}

@article{yan2026dasd,
  title={Distribution-Aligned Sequence Distillation for Superior Long-CoT Reasoning},
  author={Yan, Shaotian and Liu, Kaiyuan and Shen, Chen and Wang, Bing and Fan, Sinan and Zhang, Jun and Wu, Yue and Wang, Zheng and Ye, Jieping},
  year={2026},
  journal={arXiv preprint arXiv:2601.09088},
  url={https://arxiv.org/abs/2601.09088}
}

@article{hendryckstest2021,
  title={Measuring Massive Multitask Language Understanding},
  author={Dan Hendrycks and Collin Burns and Steven Basart and Andy Zou and Mantas Mazeika and Dawn Song and Jacob Steinhardt},
  journal={Proceedings of the International Conference on Learning Representations (ICLR)},
  year={2021}
}

@misc{grundmann2026clinibenchclinicaloutcomeprediction,
      title={CliniBench: A Clinical Outcome Prediction Benchmark for Generative and Encoder-Based Language Models}, 
      author={Paul Grundmann and Dennis Fast and Jan Frick and Thomas Steffek and Felix Gers and Wolfgang Nejdl and Alexander Löser},
      year={2026},
      eprint={2509.26136},
      archivePrefix={arXiv},
      primaryClass={cs.CL},
      url={https://arxiv.org/abs/2509.26136}, 
}

@article{Gao_2023,
   title={DR.BENCH: Diagnostic Reasoning Benchmark for Clinical Natural Language Processing},
   volume={138},
   ISSN={1532-0464},
   url={http://dx.doi.org/10.1016/j.jbi.2023.104286},
   DOI={10.1016/j.jbi.2023.104286},
   journal={Journal of Biomedical Informatics},
   publisher={Elsevier BV},
   author={Gao, Yanjun and Dligach, Dmitriy and Miller, Timothy and Caskey, John and Sharma, Brihat and Churpek, Matthew M. and Afshar, Majid},
   year={2023},
   month=feb, pages={104286} }

@article{hager_evaluation_2024,
  title = {Evaluation and Mitigation of the Limitations of Large Language Models in Clinical Decision-Making},
  author = {Hager, Paul and Jungmann, Friederike and Holland, Robbie and Bhagat, Kunal and Hubrecht, Inga and Knauer, Manuel and Vielhauer, Jakob and Makowski, Marcus and Braren, Rickmer and Kaissis, Georgios and Rueckert, Daniel},
  year = 2024,
  month = sep,
  journal = {Nature Medicine},
  volume = {30},
  number = {9},
  pages = {2613--2622},
  issn = {1546-170X},
  doi = {10.1038/s41591-024-03097-1}
}

@misc{kweon2024ehrnoteqa,
      title={EHRNoteQA: An LLM Benchmark for Real-World Clinical Practice Using Discharge Summaries}, 
      author={Sunjun Kweon and Jiyoun Kim and Heeyoung Kwak and Dongchul Cha and Hangyul Yoon and Kwanghyun Kim and Jeewon Yang and Seunghyun Won and Edward Choi},
      year={2024},
      eprint={2402.16040},
      archivePrefix={arXiv},
      primaryClass={id='cs.CL' full_name='Computation and Language' is_active=True alt_name='cmp-lg' in_archive='cs' is_general=False description='Covers natural language processing. Roughly includes material in ACM Subject Class I.2.7. Note that work on artificial languages (programming languages, logics, formal systems) that does not explicitly address natural-language issues broadly construed (natural-language processing, computational linguistics, speech, text retrieval, etc.) is not appropriate for this area.'}
}

@misc{mullenbach2021clipdatasetextractingaction,
      title={CLIP: A Dataset for Extracting Action Items for Physicians from Hospital Discharge Notes}, 
      author={James Mullenbach and Yada Pruksachatkun and Sean Adler and Jennifer Seale and Jordan Swartz and T. Greg McKelvey and Hui Dai and Yi Yang and David Sontag},
      year={2021},
      eprint={2106.02524},
      archivePrefix={arXiv},
      primaryClass={cs.CL},
      url={https://arxiv.org/abs/2106.02524}, 
}

@article{PhysioNet-mimiciii-1.4,
  author = {Johnson, Alistair and Pollard, Tom and Mark, Roger},
  title = {{MIMIC-III Clinical Database}},
  journal = {{PhysioNet}},
  year = {2016},
  month = sep,
  note = {Version 1.4},
  doi = {10.13026/C2XW26},
  url = {https://doi.org/10.13026/C2XW26}
}

@article{PhysioNet-mimiciv-3.1,
  author = {Johnson, Alistair and Bulgarelli, Lucas and Pollard, Tom and Gow, Brian and Moody, Benjamin and Horng, Steven and Celi, Leo Anthony and Mark, Roger},
  title = {{MIMIC-IV}},
  journal = {{PhysioNet}},
  year = {2024},
  month = oct,
  note = {Version 3.1},
  doi = {10.13026/kpb9-mt58},
  url = {https://doi.org/10.13026/kpb9-mt58}
}

@article{wu2025towards,
  title        = {Towards evaluating and building versatile large language models for medicine},
  author       = {Wu, Chaoyi and Qiu, Pengcheng and Liu, Jinxin and Gu, Hongfei and
                  Li, Na and Zhang, Ya and Wang, Yanfeng and Xie, Weidi},
  journal      = {npj Digital Medicine},
  volume       = {8},
  number       = {1},
  pages        = {58},
  year         = {2025},
  publisher    = {Nature Publishing Group},
  doi          = {10.1038/s41746-024-01390-4},
  url          = {https://www.nature.com/articles/s41746-024-01390-4}
}

@inproceedings{liu2024clinicbench,
  title     = {Large Language Models Are Poor Clinical Decision-Makers: A Comprehensive Benchmark},
  author    = {Liu, Fenglin and Li, Zheng and Zhou, Hongjian and Yin, Qingyu and
               Yang, Jingfeng and Tang, Xianfeng and Luo, Chen and Zeng, Ming and
               Jiang, Haoming and Gao, Yifan and Nigam, Priyanka and Nag, Sreyashi and
               Yin, Bing and Hua, Yining and Zhou, Xuan and Rohanian, Omid and
               Thakur, Anshul and Clifton, Lei and Clifton, David A.},
  booktitle = {Proceedings of the 2024 Conference on Empirical Methods in Natural Language Processing},
  editor    = {Al-Onaizan, Yaser and Bansal, Mohit and Chen, Yun-Nung},
  month     = nov,
  year      = {2024},
  address   = {Miami, Florida, USA},
  publisher = {Association for Computational Linguistics},
  pages     = {13696--13710},
  doi       = {10.18653/v1/2024.emnlp-main.759},
  url       = {https://aclanthology.org/2024.emnlp-main.759/}
}

@article{jiang2025medagentbench,
  title     = {{MedAgentBench}: A Virtual {EHR} Environment to Benchmark Medical {LLM} Agents},
  author    = {Jiang, Yixing and Black, Kameron C. and Geng, Gloria and Park, Danny and
               Zou, James and Ng, Andrew Y. and Chen, Jonathan H.},
  journal   = {NEJM AI},
  volume    = {2},
  number    = {9},
  pages     = {AIdbp2500144},
  year      = {2025},
  publisher = {Massachusetts Medical Society},
  doi       = {10.1056/AIdbp2500144},
  url       = {https://ai.nejm.org/doi/full/10.1056/AIdbp2500144}
}

@article{tang2025medagentsbench,
  title         = {{MedAgentsBench}: Benchmarking Thinking Models and Agent Frameworks for Complex Medical Reasoning},
  author        = {Tang, Xiangru and Shao, Daniel and Sohn, Jiwoong and Chen, Jiapeng and
                   Zhang, Jiayi and Xiang, Jinyu and Wu, Fang and Zhao, Yilun and
                   Wu, Chenglin and Shi, Wenqi and Cohan, Arman and Gerstein, Mark},
  journal       = {arXiv preprint arXiv:2503.07459},
  year          = {2025},
  eprint        = {2503.07459},
  archivePrefix = {arXiv},
  primaryClass  = {cs.CL},
  doi           = {10.48550/arXiv.2503.07459},
  url           = {https://arxiv.org/abs/2503.07459}
}

@article{shao2024deepseekmath,
  title={DeepSeekMath: Pushing the Limits of Mathematical Reasoning in Open Language Models},
  author={Shao, Zhihong and Wang, Peiyi and Zhu, Qihao and Xu, Runxin and Song, Junxiao and Bi, Xiao and Zhang, Haowei and Zhang, Mingchuan and Li, Y. K. and Wu, Y. and Guo, Daya},
  journal={arXiv preprint arXiv:2402.03300},
  year={2024}
}

@article{ling2025every,
  title={Every Step Evolves: Scaling Reinforcement Learning for Trillion-Scale Thinking Model},
  author={{Ling Team} and Shen, Anqi and Li, Baihui and Hu, Bin and Jing, Bin and Chen, Cai and Huang, Chao and Zhang, Chao and Yang, Chaokun and Lin, Cheng and others},
  journal={arXiv preprint arXiv:2510.18855},
  year={2025}
}

@article{zeng2026glm5,
  title={GLM-5: from Vibe Coding to Agentic Engineering},
  author={Zeng, An and others},
  journal={arXiv preprint arXiv:2602.15763},
  year={2026}
}

@misc{primeintellect2025prime-rl,
  author = {Prime Intellect},
  title = {PRIME-RL},
  url = {https://github.com/PrimeIntellect-ai/prime-rl},
  year = {2025}
}

@inproceedings{hu2022lora,
  title={LoRA: Low-Rank Adaptation of Large Language Models},
  author={Hu, Edward J. and Shen, Yelong and Wallis, Phillip and Allen-Zhu, Zeyuan and Li, Yuanzhi and Wang, Shean and Wang, Lu and Chen, Weizhu},
  booktitle={International Conference on Learning Representations},
  year={2022},
}

@inproceedings{loshchilov2019decoupled,
  title={Decoupled Weight Decay Regularization},
  author={Loshchilov, Ilya and Hutter, Frank},
  booktitle={International Conference on Learning Representations},
  year={2019},
}
\bibliographystyle{icml2026}

\newpage
\clearpage
\onecolumn
\appendix

\section{Contribution Statement}

BW: Project lead.\\
RSG: Contributed the MedMCQA, HEAD-QA v2, HEAD-QA, and SCT-Bench Public environments, RL experiments, and to manuscript.\\
MK: Contributed the MMLU-Pro-Health, M-ARC, and Medbullets environments.\\
AO: Contributed the MetaMedQA and AgentClinic environments, project code, and RL experiments.\\
SP: Contributed the MEDEC environment, exploratory analysis, writeups, and RL training.\\
GA: Contributed the MedicationQA and Med-HALT environments and dataset appendix.\\
KB: Contributed the MedRedQA environment, LLM-as-a-judge research, and RL experiments.\\
NK: Contributed the MecCalcBench environment, ran and analyzed RL experiments.\\
AH: Contributed the CareQA environment and LLM-as-a-judge research.\\
NM: Contributed the MedExQA environment, project code, and to manuscript.\\
MR: Contributed the K-QA and MedHallu environments and to manuscript.\\
SSZY: Contributed the LongHealth and CaseReportBench environments and to manuscript.\\
AE: Contributed the MedQA environment and to manuscript.\\
AJM: Contributed the MedReason dataset and to manuscript.\\
RS: Contributed the PubMedQA environment and to manuscript.\\
BH: Contributed the SuperGPQA environment and to manuscript.\\
MB: Contributed the BioHopR environment and to manuscript.\\
SG: Contributed the MedXpertQA environment.\\
AM: Contributed the MedConceptsQA environment.\\
SK: Contributed the HealthBench environment and dataset curation.\\
MG: Contributed to methodology, conceptual discussion, qualitative evaluation, and to manuscript.\\
HB: Co-Contributed the MedR-Bench environment and to manuscript.\\
JBD: Contributed the Open-i, MedQSum, and ACI-Bench summarization environments and to manuscript.\\
SB: Contributed the MT Samples-Procedure and MT Samples-Replicate environments.\\
RC: Co-contributed the MedR-Bench environment and to manuscript.\\
AV: Contributed the BioASQ environment.\\
AZ: Co-Contributed the MedSafetyBench environment and to manuscript.\\
LKM: Co-contributed the MedSafetyBench environment.\\
HD: Co-contributed the ACI-Bench environment.\\
AP: Provided project infrastructure and support.\\
WB: Provided project infrastructure and support.\\
JH: Provided project infrastructure and support.\\
CL: Project feedback and contributed to manuscript.\\
PSS: Provided guidance and support, and contributed to manuscript.\\
TMA: Provided guidance and support, and contributed to manuscript.\\

\clearpage
\section{All Model Win Rates \textsc{Medmarks-V}}

\begin{table}[ht!]
\centering
\scriptsize
\caption{All models by mean win rate on \textsc{Medmarks-V}.}
\label{tab:medmarks-v-results}
\begin{tabular}{llr}
\toprule
Model & Size & Win Rate \\
\midrule
Gemini 3 Pro Preview & API & 0.6628 \\
GPT-5.1 (med) & API & 0.6395 \\
Grok 4 & API & 0.6343 \\
Claude Sonnet 4.5 & API & 0.6258 \\
GPT-5.2 (med) & API & 0.6236 \\
GLM 4.7 FP8 & Large & 0.6199 \\
Qwen3 235B-A22B Thinking & Large & 0.6032 \\
Baichuan M3 235B & Large & 0.5983 \\
Qwen3 Next 80B-A3B Thinking & Large & 0.5888 \\
MiniMax M2.1 & Large & 0.5882 \\
gpt-oss 120b (high) & Large & 0.5865 \\
gpt-oss 120b (med) & Large & 0.5771 \\
MiniMax M2 & Large & 0.5708 \\
Qwen3 Next 80B-A3B Instruct & Large & 0.5687 \\
Intellect 3 & Large & 0.5660 \\
Qwen3 30B-A3B Thinking FP8 & Medium & 0.5601 \\
Qwen3 30B-A3B Thinking 8-bit & Medium & 0.5591 \\
Qwen3 30B-A3B Thinking & Medium & 0.5587 \\
gpt-oss 120b (low) & Large & 0.5524 \\
AntAngelMed 100B & Large & 0.5524 \\
Baichuan M2 32B & Medium & 0.5520 \\
Qwen3 VL 30B-A3B Thinking & Medium & 0.5509 \\
Qwen3 30B-A3B Thinking 4-bit & Medium & 0.5481 \\
GLM 4.5 Air & Large & 0.5410 \\
Qwen3 14B (Thinking) & Small & 0.5367 \\
Llama 3.3 70B Instruct & Large & 0.5364 \\
gpt-oss 20b (high) & Medium & 0.5361 \\
Qwen3 30B-A3B Instruct 8-bit & Medium & 0.5321 \\
Qwen3 30B-A3B Instruct & Medium & 0.5317 \\
Qwen3 30B-A3B Instruct FP8 & Medium & 0.5311 \\
Qwen3 30B-A3B Instruct 4-bit & Medium & 0.5253 \\
gpt-oss 20b (med) & Medium & 0.5198 \\
Ling Flash 2.0 & Large & 0.5174 \\
Nemotron Nano V3 30B-A3B & Medium & 0.5114 \\
Olmo 3.1 32B Think & Medium & 0.5067 \\
MedGemma 27B & Medium & 0.5024 \\
Olmo 3 32B Think & Medium & 0.5021 \\
Qwen3 8B (Thinking) & Small & 0.5019 \\
Nemotron Nano 12B V2 & Small & 0.5006 \\
Qwen2.5 32B Instruct & Medium & 0.5003 \\
Qwen3 4B Thinking & Tiny & 0.4891 \\
Trinity Mini & Medium & 0.4853 \\
Hermes 4 70B & Large & 0.4850 \\
gpt-oss 20b (low) & Medium & 0.4820 \\
Phi 4 Reasoning & Small & 0.4815 \\
Ministral 3 14B Reasoning & Small & 0.4807 \\
Hermes 4 14B & Small & 0.4793 \\
Mirothinker 1.5 30B & Medium & 0.4768 \\
Ministral 3 14B Instruct & Small & 0.4760 \\
Magistral Small & Medium & 0.4745 \\
Gemma 3 27B & Medium & 0.4610 \\
Olmo 3.1 32B Instruct & Medium & 0.4560 \\
DASD 4B Thinking & Tiny & 0.4525 \\
Jamba2 Mini 52B & Large & 0.4458 \\
Granite 4.0H Small & Medium & 0.4452 \\
Ministral 3 8B Instruct & Small & 0.4442 \\
Gemma 3 12B & Small & 0.4379 \\
Ministral 3 8B Reasoning & Small & 0.4312 \\
Olmo 3 7B Think & Small & 0.4130 \\
DASD 30B-A3B & Medium & 0.4079 \\
Llama 3.1 8B Instruct & Small & 0.3967 \\
Ministral 3 3B Instruct & Tiny & 0.3927 \\
MedGemma 4B & Tiny & 0.3779 \\
MedGemma 4B 1.5 & Tiny & 0.3759 \\
Ministral 3 3B Reasoning & Tiny & 0.3741 \\
Trinity Nano Preview & Tiny & 0.3596 \\
SmolLM3 3B & Tiny & 0.3548 \\
Granite 4.0H Tiny & Small & 0.3522 \\
Olmo 3 7B Instruct & Small & 0.3503 \\
Gemma 3 4B & Tiny & 0.3214 \\
AFM 4.5B & Tiny & 0.3193 \\
\bottomrule
\end{tabular}
\end{table}
\FloatBarrier

\clearpage
\section{Local Inference Cost}

\begin{table}[ht!]
\centering
\scriptsize
\caption{\textsc{Medmarks-V} local inference cost at \$2 per H100 hour.}
\label{tab:estimated-inference}
\begin{tabular}{llrrr}
\toprule
Model & Size & H100 hours & Total Cost (\$) & Per Example (\$) \\
\midrule
Ministral 3 3B Instruct & Tiny & 6.69 & 13.37 & 0.0001 \\
Ministral 3 3B Reasoning & Tiny & 4.62 & 9.24 & 0.0001 \\
SmolLM3 3B & Tiny & 50.12 & 100.23 & 0.0008 \\
DASD 4B Thinking & Tiny & 66.16 & 132.33 & 0.0011 \\
Gemma 3 4B & Tiny & 2.80 & 5.60 & 0.0000 \\
MedGemma 4B & Tiny & 24.71 & 49.42 & 0.0004 \\
MedGemma 4B 1.5 & Tiny & 16.58 & 33.15 & 0.0003 \\
Qwen3 4B Thinking & Tiny & 24.48 & 48.95 & 0.0004 \\
AFM 4.5B & Tiny & 13.97 & 27.93 & 0.0002 \\
Trinity Nano Preview & Tiny & 45.90 & 91.81 & 0.0008 \\
Granite 4.0H Tiny & Small & 5.50 & 11.01 & 0.0001 \\
Olmo 3 7B Instruct & Small & 10.73 & 21.46 & 0.0002 \\
Olmo 3 7B Think & Small & 70.13 & 140.25 & 0.0012 \\
Llama 3.1 8B Instruct & Small & 18.99 & 37.98 & 0.0003 \\
Ministral 3 8B Instruct & Small & 9.00 & 18.01 & 0.0001 \\
Ministral 3 8B Reasoning & Small & 4.57 & 9.14 & 0.0001 \\
Qwen3 8B (Thinking) & Small & 12.36 & 24.72 & 0.0002 \\
Gemma 3 12B & Small & 2.98 & 5.97 & 0.0000 \\
Nemotron Nano 12B V2 & Small & 17.99 & 35.99 & 0.0003 \\
Hermes 4 14B & Small & 2.43 & 4.87 & 0.0000 \\
Ministral 3 14B Instruct & Small & 12.72 & 25.44 & 0.0002 \\
Ministral 3 14B Reasoning & Small & 7.44 & 14.87 & 0.0001 \\
Phi 4 Reasoning & Small & 29.30 & 58.59 & 0.0005 \\
Qwen3 14B (Thinking) & Small & 11.30 & 22.59 & 0.0002 \\
gpt-oss 20b (low) & Medium & 1.05 & 2.11 & 0.0000 \\
gpt-oss 20b (med) & Medium & 2.83 & 5.65 & 0.0000 \\
gpt-oss 20b (high) & Medium & 22.24 & 44.49 & 0.0003 \\
Magistral Small & Medium & 5.58 & 11.16 & 0.0001 \\
Trinity Mini & Medium & 32.69 & 65.38 & 0.0005 \\
Gemma 3 27B & Medium & 6.25 & 12.50 & 0.0001 \\
MedGemma 27B & Medium & 40.32 & 80.63 & 0.0007 \\
DASD 30B-A3B & Medium & 166.91 & 333.82 & 0.0027 \\
Mirothinker 1.5 30B & Medium & 226.14 & 452.29 & 0.0037 \\
Nemotron Nano V3 30B-A3B & Medium & 31.27 & 62.54 & 0.0005 \\
Qwen3 30B-A3B Instruct 4-bit & Medium & 13.44 & 26.89 & 0.0002 \\
Qwen3 30B-A3B Instruct 8-bit & Medium & 11.16 & 22.32 & 0.0002 \\
Qwen3 30B-A3B Instruct FP8 & Medium & 15.68 & 31.35 & 0.0003 \\
Qwen3 30B-A3B Instruct & Medium & 14.26 & 28.53 & 0.0002 \\
Qwen3 30B-A3B Thinking 4-bit & Medium & 22.15 & 44.30 & 0.0004 \\
Qwen3 30B-A3B Thinking 8-bit & Medium & 25.14 & 50.29 & 0.0004 \\
Qwen3 30B-A3B Thinking FP8 & Medium & 27.02 & 54.04 & 0.0004 \\
Qwen3 30B-A3B Thinking & Medium & 29.13 & 58.26 & 0.0005 \\
Qwen3 VL 30B-A3B Thinking & Medium & 31.41 & 62.82 & 0.0005 \\
Baichuan M2 32B & Medium & 47.12 & 94.24 & 0.0008 \\
Granite 4.0H Small & Medium & 9.36 & 18.72 & 0.0002 \\
Olmo 3 32B Think & Medium & 63.51 & 127.02 & 0.0010 \\
Olmo 3.1 32B Instruct & Medium & 9.33 & 18.66 & 0.0002 \\
Olmo 3.1 32B Think & Medium & 81.86 & 163.72 & 0.0014 \\
Qwen2.5 32B Instruct & Medium & 5.74 & 11.49 & 0.0001 \\
Jamba2 Mini 52B & Large & 15.18 & 30.36 & 0.0003 \\
Hermes 4 70B & Large & 12.90 & 25.79 & 0.0002 \\
Llama 3.3 70B Instruct & Large & 19.67 & 39.34 & 0.0003 \\
Qwen3 Next 80B-A3B Instruct & Large & 18.59 & 37.18 & 0.0003 \\
Qwen3 Next 80B-A3B Thinking & Large & 77.71 & 155.42 & 0.0013 \\
AntAngelMed 100B & Large & 69.15 & 138.31 & 0.0011 \\
Ling Flash 2.0 & Large & 17.40 & 34.79 & 0.0003 \\
GLM 4.5 Air & Large & 327.78 & 655.55 & 0.0054 \\
Intellect 3 & Large & 113.74 & 227.49 & 0.0019 \\
gpt-oss 120b (low) & Large & 2.59 & 5.18 & 0.0000 \\
gpt-oss 120b (med) & Large & 5.51 & 11.02 & 0.0001 \\
gpt-oss 120b (high) & Large & 39.01 & 78.02 & 0.0006 \\
MiniMax M2 & Large & 170.72 & 341.44 & 0.0028 \\
MiniMax M2.1 & Large & 154.69 & 309.38 & 0.0025 \\
Baichuan M3 235B & Large & 241.22 & 482.44 & 0.0039 \\
Qwen3 235B-A22B Thinking & Large & 402.98 & 805.96 & 0.0066 \\
GLM 4.7 FP8 & Large & 468.54 & 937.08 & 0.0077 \\
\midrule
TOTAL & & 3570.44 & 7140.87 &   \\
\bottomrule
\end{tabular}
\end{table}

\section{Related Works}

Medical capabilities of LLMs have mostly been evaluated with multiple-choice question answering benchmarks \cite{singhal2023large} like MedQA \cite{jin2021disease}, PubMedQA \cite{jin2019pubmedqa}, MedMCQA \cite{pal2022medmcqa}, MMLU \cite{hendryckstest2021}, and MMLU Pro Health \cite{wang2024mmlu}. 

The recent HealthBench benchmark aims to evaluate models in more realistic scenarios, with questions and rubrics designed by clinicians \cite{arora2025healthbench}. However, it focuses solely on medical conversations, so non-conversational medical capabilities are not evaluated. 

A variety of LLM benchmarking suites for clinical use-cases have been developed, like CliniBench \cite{grundmann2026clinibenchclinicaloutcomeprediction} and DR.BENCH \cite{Gao_2023}. However, most of these suites have very limited scopes. Instead, the MedHELM suite \cite{bedi_holistic_2026} expands evaluation to 37 benchmarks focused on representing real-world medical use-cases. However, only 13 of the 35 underlying datasets are publicly accessible, preventing full replication by the community.

There are a variety of benchmarks (MIMIC-CDM \cite{hager_evaluation_2024} , EHRNoteQA \cite{kweon2024ehrnoteqa}, CLIP \cite{mullenbach2021clipdatasetextractingaction}) are built on top of publicly accessible but gated datasets like MIMIC-III \cite{PhysioNet-mimiciii-1.4} and MIMIC-IV \cite{PhysioNet-mimiciv-3.1}. Instead, we focus on fully open datasets to ensure accessibility and reproducibility of our evaluation suite.

\section{Comparison with Prior Medical LLM Benchmark Suites}
\label{app:suite_comparison}

Table~\ref{tab:suite_comparison} compares \textsc{Medmarks} with prior medical LLM benchmark suites along five axes: number of datasets, fraction that are fully open (usable without credentialing or a data-use agreement), task coverage, and the number of models evaluated.

\begin{table*}[ht]
\centering
\small
\caption{Comparison of \textsc{Medmarks} with prior medical LLM benchmark suites. Model counts are from each suite's reference publication.}
\label{tab:suite_comparison}
\begin{adjustbox}{max width=\textwidth}
\begin{tabular}{lccccccc}
\toprule
Suite & \# Datasets & Fully open / Total & MCQ & Open-ended & Agentic & \# Models \\
\midrule
MultiMedQA \cite{singhal2023large}                                 & 7             & 7 / 7                & \checkmark & \checkmark & $\times$   & 3  \\
MedS-Bench \cite{wu2025towards}                                    & 28            & 28 / 28              & \checkmark & \checkmark & $\times$   & 9  \\
HealthBench \cite{arora2025healthbench}                            & 1             & 1 / 1                & $\times$   & \checkmark & $\times$   & 9  \\
ClinicBench \cite{liu2024clinicbench}                              & 17            & 17 / 17              & \checkmark & \checkmark & $\times$   & 22 \\
MedAgentBench \cite{jiang2025medagentbench}                        & 1 (10 cat.)   & 1 / 1                & $\times$   & $\times$   & \checkmark & 12 \\
MedAgentsBench \cite{tang2025medagentsbench}                       & 8             & 8 / 8                & \checkmark & $\times$   & \checkmark & 10 \\
MedHELM \cite{bedi_holistic_2026}                                  & 35            & 14 / 35              & \checkmark & \checkmark & $\times$   & 9  \\
\textbf{\textsc{Medmarks} (ours)}                                  & \textbf{30}   & \textbf{30 / 30}     & \checkmark & \checkmark & \checkmark & \textbf{61 (71 cfgs)} \\
\bottomrule
\end{tabular}
\end{adjustbox}
\end{table*}

MedHELM is the closest prior effort in scope. Of its 37 benchmarks, 16 are fully public, 7 require PhysioNet credentialing, and 14 are private~\cite{bedi_holistic_2026}. MedS-Bench is the only other prior suite at comparable scale and openness, but does not include open-ended tasks evaluated with LLM-as-a-Judge, nor agentic tasks. ClinicBench covers a similar task breadth but is smaller. MultiMedQA and MedAgentsBench are MCQ-only; HealthBench is open-ended dialogue only; MedAgentBench is a single FHIR-based agentic environment.

\textsc{Medmarks} is the only suite that is simultaneously fully open, covers verifiable, open-ended, and agentic tasks at 30-benchmark breadth, evaluates models at scale (61 models on 71 configurations), and ships every benchmark as a \texttt{verifiers} environment \cite{brown_verifiers_2025} with a reward function, so the eight datasets with train/test splits (\textsc{Medmarks-T}) can be used directly for RL post-training.

\section{Dataset Details}
\label{app:dataset_details}

All the benchmarks included in \textsc{Medmarks} have been listed in Table \ref{tab:datasets}, along with their descriptions.

\begin{table*}[ht!]
\centering
\caption{Medical benchmark datasets in \textsc{MedMarks} for LLM evaluation. ``--'' indicates no dedicated training split.}
\label{tab:datasets}
\begin{adjustbox}{max width=\textwidth}
\scriptsize
\setlength{\tabcolsep}{4pt}
\renewcommand{\arraystretch}{1.15}
\begin{tabular}{@{}lp{11cm}rr@{}}
\toprule
\textbf{Dataset} & \textbf{Description} & \textbf{\#Evaluated} & \textbf{\#Training} \\
\midrule
\multicolumn{4}{c}{\textit{\textbf{MedMarks-V (Verifiable)}}} \\
\midrule
MedQA~\cite{jin2021disease} & Multiple-choice questions from USMLE medical licensing exams. & 1,270 & 10,178 \\
MedMCQA~\cite{pal2022medmcqa} & Multiple-choice questions from Indian medical entrance exams across 21 medical subjects. & 4,180 & 182,822 \\
PubMedQA~\cite{jin2019pubmedqa} & Yes/no/maybe question answering requiring reasoning over biomedical research abstracts, labeled subset. & 500 & 500 \\
MedConceptsQA~\cite{shoham2024medconceptsqa} & Multiple-choice questions on medical coding systems, e.g., ICD-9, ICD-10, etc., only ICD-10CM subsamples evaluated. & 6,000 & -- \\
HEAD-QA v2~\cite{correa2025head} & Extended healthcare questions spanning 10 years of Spanish professional exams, English subset. & 12,800 & -- \\
MedXpertQA~\cite{zuo2025medxpertqa} & High-difficulty MCQ questions with $\sim$10 options across 17 specialties to evaluate expert-level medical knowledge, text subset. & 2,460 & -- \\
MedCalc-Bench~\cite{khandekar2024medcalc} & Clinical calculator questions evaluating medical computation and formula application skills. & 1,100 & 10,543 \\
LongHealth~\cite{adams2025longhealth} & Long-context synthetic patient cases with information extraction and sorting tasks. & 400 & -- \\
Med-HALT~\cite{pal2023med} & Clinical Reasoning Hallucination detection via false confidence tests and ``none of the above'' recognition. & 11,076 & -- \\
MedHallu~\cite{pandit2025medhallucomprehensivebenchmarkdetecting} & Medical hallucination detection benchmark with four domain-specific error categories derived from the PubMedQA dataset. & 10,000 & -- \\
MMLU-Pro-Health~\cite{wang2024mmlu} & Health subset of MMLU-Pro benchmark featuring general health-related questions with up to 10 answer options per question. & 823 & -- \\
M-ARC~\cite{kim2025limitationslargelanguagemodels} &  Long-tail medical questions designed to test model resistance to inflexible clinical reasoning patterns. & 100 & -- \\
Medbullets~\cite{chen-etal-2025-benchmarking} & USMLE Step 2 and Step 3 style clinical reasoning questions sourced from social media. & 308 & -- \\
MetaMedQA~\cite{griot_large_2025} & Questions testing model's awareness and recognition of unanswerable medical queries using uncertainty options. & 1,373 & -- \\
SuperGPQA-Med~\cite{pteam2025supergpqascalingllmevaluation} & Graduate-level questions spanning 6 medical fields at easy, medium, and hard difficulty levels. & 2,755 & -- \\
SCTPublic~\cite{mccoy2025assessment} & Script Concordance Tests evaluating clinical reasoning under diagnostic uncertainty. & 750 & -- \\
PubHealthBench~\cite{harris2025pubhealthbench} & Multiple-choice questions derived from UK government public health guidance documents. & 7,929 & -- \\
\midrule
\multicolumn{4}{c}{\textit{\textbf{MedMarks-OE (Open-Ended)}}} \\
\midrule
HealthBench~\cite{arora2025healthbench} & Multi-turn healthcare conversations evaluated using physician-written scoring rubrics. & 5,000 & -- \\
MedExQA~\cite{kim2024medexqamedicalquestionanswering} & Questions with dual expert explanations across 5 underrepresented medical specialties. & 940 & -- \\
MedicationQA~\cite{benabacha2019medicationqa} & Consumer-style medication questions with expert-validated answers from MedlinePlus. & 690 & -- \\
MedR-Bench~\cite{qiu2025quantifying} & Clinical reasoning benchmark with step-by-step diagnostic and treatment planning traces on rare disease cases. & 1,453 & -- \\
CareQA~\cite{arias-duart-etal-2025-automatic} & Healthcare QA exam questions with both MCQ and open-ended reasoning questions, English subset. & 1,134 & -- \\
MEDEC~\cite{abacha2025medecbenchmarkmedicalerror} & Medical dataset for clinical error detection, extraction, and correction in synthetic medical notes. & 597 & 2189 \\
ACI-Bench~\cite{yim2023aci} & Clinical dialogue transcripts paired with corresponding structured clinical notes. & 210 & 114 \\
AgentClinic~\cite{schmidgall2024agentclinic} & Multimodal Multi-agent OSCE-style clinical dialogues for interactive diagnostic reasoning evaluation. & 107 & -- \\
MedAgentBench v2~\cite{jiang2025medagentbench} & Agentic electronic health record tasks requiring FHIR API interactions. & 600 & -- \\
MedDialog~\cite{he2020meddialog} & Large-scale patient-doctor conversations for medical dialogue generation and understanding, we evaluated a small subsample. & 2,500 & 205,973 \\
MedCaseReasoning~\cite{wu2025medcasereasoning} & Diagnostic QA with clinician-authored reasoning traces from clinical case reports. & 500 & 13,592 \\
MTSamples-Procedures~\cite{bedi_holistic_2026} & Transcribed medical operative notes documenting surgical procedures evaluating models on procedural summary or treatment plans. & 90 & -- \\
MTSamples-Replicate~\cite{bedi_holistic_2026} & Transcribed medical reports from various specialties to evaluate a model's ability to generate clinically appropriate treatment plans & 2000 & -- \\
\bottomrule
\end{tabular}
\end{adjustbox}
\end{table*}

\subsection{Dataset-specific Evaluation Protocol Changes}
\label{app:eval_changes}

We made some modifications to the evaluation protocol used for some benchmarks:

\begin{enumerate}
  \item MedConceptsQA \citep{shoham2024medconceptsqa} and MedDialog \citep{he2020meddialog} have 819K and 25k examples, respectively, so we only evaluate on a subset of these datasets. MedConceptsQA tests the model's knowledge of ICD-10 codes, which have a hierarchical structure corresponding to different categories. We used this structure to select a representative sample of 2,000 questions from the easy, medium, and hard subsets for a total of 6,000 questions. Details of this selection process can be found in the data appendix. For MedDialog, we selected the first 2.5k examples out of 25K instead.
  \item We only perform one run of HeadQA-v2 \citep{correa2025head}, MedCalc-Bench \citep{khandekar2024medcalc}, and Med-HALT \citep{pal2023med} instead of three.
\end{enumerate}

\newpage
\FloatBarrier
\section{Models}
\label{app:models}

\begin{table*}[ht!]
\centering
\caption{Models evaluated in \textsc{MedMarks}. Size categories: Tiny ($<$7B), Small (7--19B), Medium (20--40B), Large ($>$40B on single node), API (proprietary or multi-node).}
\label{tab:models}
\scriptsize
\setlength{\tabcolsep}{6pt}
\renewcommand{\arraystretch}{1.05}
\begin{tabular}{@{}p{5.5cm}l@{\hspace{1.5cm}}p{6cm}@{}}
\toprule
\textbf{Model} & \textbf{Size} & \textbf{Sampling Parameters} \\
\midrule
GPT-5.1~\cite{openai_gpt5.1_systemcard_2025} & API & Temp : 1.0  \\
GPT-5.2~\cite{openai_gpt5.2_systemcard_2025} & API & Temp : 1.0  \\
Claude Sonnet 4.5~\cite{anthropic_introducing_sonnet_2025} & API & Temp : 0.7  \\
Grok 4~\cite{xai_grok4_model_card_2025} & API & Temp : 1.0, top\_p 0.95  \\
Gemini 3 Pro Preview~\cite{gemmateam2025gemma3technicalreport} & API & Temp : 1.0, top\_p 0.95, top\_k 64  \\
gpt-oss 120B (high/med/low)~\cite{openai_gpt-oss-120b_2025} & Large & Temp : 1.0, top\_p : 1.0, top\_k : 0 \\
Qwen3 235B-A22B Thinking~\cite{yang2025qwen3} & Large & Temp : 0.6, top\_p : 0.95, top\_k : 20 \\
Qwen3 Next 80B-A3B Thinking~\cite{qwenteam_qwen3-next_2025} & Large & Temp : 0.6, top\_p : 0.95, top\_k : 20 \\
Qwen3 Next 80B-A3B Instruct~\cite{qwenteam_qwen3-next_2025} & Large & Temp : 0.7, top\_p : 0.8, top\_k : 20 \\
Llama 3.3 70B Instruct~\cite{grattafiori2024llama} & Large & Temp : 0.7, top\_p : 0.95, top\_k : 0 \\
Hermes 4 70B~\cite{teknium_hermes_2025} & Large & Temp : 0.6, top\_p : 0.95, top\_k : 20 \\
MiniMax M2/M2.1~\cite{minimax_m2.1_minimaxio_2025} & Large & Temp : 1.0, top\_p : 0.95, top\_k : 40 \\
Intellect 3~\cite{primeintellectteam2025intellect3technicalreport} & Large & Temp : 0.6, top\_p : 0.95 \\
GLM 4.5 Air~\cite{glmteam2025glm45agenticreasoningcoding} & Large & Temp : 0.6, top\_p : 0.95 \\
GLM 4.7~\cite{glmteam2025glm45agenticreasoningcoding} & Large & Temp : 1.0, top\_p : 0.95 \\
Baichuan M3 235B~\cite{baichuan-m3} & Large & Temp : 0.6, top\_p : 0.95, top\_k : 20 \\
AntAngelMed~\cite{antangelmed_2025} & Large & Temp : 0.6, top\_p : 0.95, top\_k : 20 \\
gpt-oss 20B (high/med/low)~\cite{openai_gpt-oss-120b_2025} & Medium & Temp : 1.0, top\_p : 1.0, top\_k : 0 \\
Qwen3 30B-A3B Thinking~\cite{yang2025qwen3} & Medium & Temp : 0.6, top\_p : 0.95, top\_k : 20 \\
Qwen3 30B-A3B Instruct~\cite{yang2025qwen3} & Medium & Temp : 0.7, top\_p : 0.8, top\_k : 20 \\
DASD 30B-A3B~\cite{yan2026dasd} & Medium & Temp : 1.0, top\_p : 1.0 \\
Olmo 3 32B Think~\cite{olmo2025olmo3} & Medium & Temp : 0.6, top\_p : 0.95 \\
Olmo 3.1 32B Think~\cite{olmo2025olmo3} & Medium & Temp : 0.6, top\_p : 0.95 \\
Olmo 3.1 32B Instruct~\cite{olmo2025olmo3} & Medium & Temp : 0.6, top\_p : 0.95 \\
Baichuan M2 32B~\cite{m2team2025baichuanm2scalingmedicalcapability} & Medium & Temp : 0.6, top\_p : 0.95, top\_k : 20 \\
MedGemma 27B~\cite{sellergren_medgemma_2025} & Medium & Temp : 0.0, top\_p : 1.0, top\_k : 0 \\
Gemma 3 27B~\cite{gemmateam2025gemma3technicalreport} & Medium & Temp : 1.0, top\_p : 0.95, top\_k : 60 \\
Magistral Small~\cite{mistralai2025magistral} & Medium & Temp : 0.7, top\_p : 0.95 \\
MiroThinker 1.5 30B~\cite{miromind2025mirothinker} & Medium & Temp : 1.0, top\_p : 0.95 \\
Nemotron 3 Nano 30B-A3B~\cite{nvidia2025nemotron3nanoopen} & Medium & Temp : 1.0, top\_p : 1.0 \\
Ling 2 Flash~\cite{lingteam2025activationboostedscalinggeneral} & Medium & Temp : 0.7, top\_p : 0.8 \\
Trinity Mini 26B-A3B~\cite{arcee2025trinitymanifesto} & Medium & Temp : 0.15, top\_p : 0.75, top\_k : 50 \\
Qwen3 14B Thinking~\cite{yang2025qwen3} & Small & Temp : 0.6, top\_p : 0.95, top\_k : 20 \\
Qwen3 8B Thinking~\cite{yang2025qwen3} & Small & Temp : 0.6, top\_p : 0.95, top\_k : 20 \\
Ministral 3 14B Instruct~\cite{liu2026ministral3} & Small & Temp : 0.1, top\_p : 0.95 \\
Ministral 3 14B Reasoning~\cite{liu2026ministral3} & Small & Temp : 0.7, top\_p : 0.95 \\
Ministral 3 8B Instruct~\cite{liu2026ministral3} & Small & Temp : 0.1, top\_p : 0.95 \\
Ministral 3 8B Reasoning~\cite{liu2026ministral3} & Small & Temp : 0.7, top\_p : 0.95 \\
Nemotron Nano 12B V2~\cite{nvidia2025nvidianemotronnano2} & Small & Temp : 0.6, top\_p : 0.95 \\
Hermes 4 14B~\cite{teknium_hermes_2025} & Small & Temp : 0.6, top\_p : 0.95, top\_k : 20 \\
Gemma 3 12B~\cite{gemmateam2025gemma3technicalreport} & Small & Temp : 1.0, top\_p : 0.95, top\_k : 60 \\
Phi 4 Reasoning~\cite{abdin_phi-4-reasoning_2025} & Small & Temp : 0.8, top\_p : 0.95, top\_k : 50 \\
Llama 3.1 8B Instruct~\cite{grattafiori2024llama} & Small & Temp : 0.7, top\_p : 0.95, top\_k : 0 \\
Olmo 3 7B Think~\cite{olmo2025olmo3} & Small & Temp : 0.6, top\_p : 0.95 \\
Olmo 3 7B Instruct~\cite{olmo2025olmo3} & Small & Temp : 0.6, top\_p : 0.95 \\
Granite 4.0H Small~\cite{ibm_granite4_overview_2025} & Small & Temp : 0.0, top\_p : 1.0, top\_k : 0 \\
Granite 4.0H Tiny~\cite{ibm_granite4_overview_2025} & Small & Temp : 0.0, top\_p : 1.0, top\_k : 0 \\
Jamba2 Mini~\cite{ai21labs_jamba2_blog_2026} & Small & Temp : 0.6, top\_p : 1.0 \\
Trinity Nano 6B-A1B~\cite{arcee2025trinitymanifesto} & Tiny & Temp : 0.5, top\_p : 0.95, top\_k : 50 \\
DASD 4B~\cite{yan2026dasd} & Tiny & Temp : 1.0, top\_p : 1.0 \\
Qwen3 4B Thinking~\cite{yang2025qwen3} & Tiny & Temp : 0.6, top\_p : 0.95, top\_k : 20 \\
Ministral 3 3B Instruct~\cite{liu2026ministral3} & Tiny & Temp : 0.1, top\_p : 0.95 \\
Ministral 3 3B Reasoning~\cite{liu2026ministral3} & Tiny & Temp : 0.7, top\_p : 0.95 \\
MedGemma 4B~\cite{sellergren_medgemma_2025} & Tiny & Temp : 0.0, top\_p : 1.0, top\_k : 0 \\
MedGemma 1.5 4B~\cite{golden_next_2026} & Tiny & Temp : 0.0, top\_p : 0.95, top\_k : 64 \\
Gemma 3 4B~\cite{gemmateam2025gemma3technicalreport} & Tiny & Temp : 1.0, top\_p : 0.95, top\_k : 60 \\
SmolLM3 3B~\cite{bakouch2025smollm3} & Tiny & Temp : 0.6, top\_p : 0.95 \\
AFM 4.5B~\cite{arcee2025afm45b} & Tiny & Temp : 0.5, top\_p : 0.95, top\_k : 50 \\

\bottomrule
\end{tabular}
\end{table*}
\FloatBarrier

\section{Mean Win Rate}
\label{app:win_rate}

\textsc{Medmarks} compares models using a dataset-weighted mean win rate. For each dataset $d$, model $m$ is compared against every other model $m'\in\mathcal{M}$ by assigning a win score $\I_{m,m',d}=1$ if $\score_{m,d}>\score_{m',d}$, $\I_{m,m',d}=0.5$ when tied $(\score_{m,d}=\score_{m',d})$ and $\I_{m,m',d}=0$ if $\score_{m,d}<\score_{m',d}$.

The per-dataset win rate is then the average of these scores:

\begin{equation}
\mathrm{WR}_d(m)
=\frac{1}{|\mathcal{M}|-1}\sum_{m'\in\mathcal{M}}
\I_{m,m',d}.
\end{equation}

When aggregating across datasets, \textsc{Medmarks} uses log-size weighting. Let $N_d$ denote the number of evaluation instances in dataset $d$.

\begin{equation}
\begin{aligned}
\mathrm{MWR}_{\log N}(m)
&=\frac{\sum_{d\in\mathcal{D}}\log(N_d)\,\mathrm{WR}_d(m)}
        {\sum_{d\in\mathcal{D}}\log(N_d)} .
\end{aligned}
\end{equation}

\section{Multiple Choice Grading Function}
\label{app:mcq_grade_func}

In detail, \textsc{Medmarks-V} multiple choice grading function normalizes and strips known extraneous text, looks for an exact match answer (only the multiple choice character), looks for the answer character leading the answer text, attempts to match the answer near common answer prefixes (``the answer is:'', ``in conclusion'', ``best supported''), attempts to find the answer character in the tail without any negation (``C is incorrect'', ``not C''), and finally attempts to exact match the answer text if it exists in the first or last sentence without nearby negation. This strategy of handling answer outputs has a few pitfalls, like grading an answer as wrong if the correct answer is provided in the middle of a paragraph. Another known ``problem'' is when the wrong letter choice is followed by the correct text. This is currently graded as incorrect. There are likely more unaccounted edge cases that may be discovered during RL training in the form of reward hacking.

We plan on exploring additional multiple choice grading methods in future updates of the \textsc{Medmarks-V} benchmark.

\section{Judge Model Selection Process}
\label{app:judge_selection}

We considered Claude Haiku 4.5, Gemini 2.5 Flash \& 3 Flash (preview), Grok 4 Fast \& 4.1 Fast, GPT-4.1 mini \& nano, GPT-4o mini, and GPT-5 mini \& nano.\footnote{We also tested GLM 4.7 and Kimi K2 Thinking, but these models were too slow via API access to use as a Judge.} We selected our judge models in a two-step process. First, we sampled a subset of questions and answers from a representative set of datasets. We used LiteLLM Proxy \footnote{\href{https://docs.litellm.ai/docs/simple_proxy}{https://docs.litellm.ai/docs/simple\_proxy}} to cache the answers of a local model and scored identical rollouts across each judge. Second, we created a custom web app to crowdsource blind head-to-head judge comparisons on questions with the most disagreement between them to rank potential judges. We then used these crowdsourced rankings to inform our judge selection. When two judges were graded similarly, we chose the less expensive option.

GPT-5 mini appeared to be the most well-rounded judge and we use it for all benchmarks, except MedExQA where GPT-5 nano was widely preferred by our human graders, and paired it with Grok 4.1 Fast for numerical ratings and Gemini 3 Flash (preview) for all other LLM-as-a-Judge prompts.\footnote{While Gemini 3 Flash is an excellent judge, it was far too lenient when returning a numerical score which resulted in grade inflation.} For OpenAI's HealthBench we switched from the slower and expensive GPT-4.1 to GPT-5 mini.

\begin{table*}[ht!]
\centering
\caption{Models used as judge for LLM-as-a-judge evaluation in \textsc{MedMarks}.}
\label{tab:models-api}
\scriptsize
\setlength{\tabcolsep}{6pt}
\renewcommand{\arraystretch}{1.05}
\begin{tabular}{@{}p{5.5cm}l@{\hspace{1cm}}>{\raggedleft\arraybackslash}p{6cm}@{}}
\toprule
\textbf{Model} & \textbf{Size} & \textbf{Sampling Parameters} \\
\midrule
Gemini 3 Flash Preview~\cite{gemmateam2025gemma3technicalreport} & API & \{Temp : 1.0, top\_p 0.95, Reasoning: Low\} \\
Grok 4.1 Fast~\cite{xai_grok4.1_model_card_2025} & API & Temp 0.2 \\
GPT-5 mini~\cite{openai_gpt5_systemcard_2025} & API & \{Temp : 1.0, Reasoning: Low \} \\
GPT-5 nano~\cite{openai_gpt5_systemcard_2025} & API & \{Temp : 1.0, Reasoning: Low \} \\
\bottomrule
\end{tabular}
\end{table*}

\section{Reinforcement Learning Training Details}
\label{app:rl_training}

To demonstrate that our verifiers environments can support RL training, we do a preliminary training run of Qwen-3-4B-Instruct-0725 for 500 steps with MedMCQA, MedCaseReasoning, and MedCalc-Bench Verified using 8 H100 80 GB GPUs. We used the Prime-RL library ~\citep{primeintellect2025prime-rl} with LoRA adapters ~\citep{hu2022lora} (rank=32, alpha=64) on all linear layers in the attention and feed-forward blocks of the LLM. We combined the all three training environments at an even sampling ratio, and evaluate on the first 300 samples from each test set except for MedCaseReasoning which uses 100 samples. We used PrimeRL's default RL configuration, which is based on an IcePop-style \citep{ling2025every} variant of Group Relative Policy Optimization (GRPO)~\citep{shao2024deepseekmath}. This configuration independently followed GLM-5~\citep{zeng2026glm5} in omitting the IcePop KL regularization term and using double-sided importance-sampling/token-masking to filter updates with large training--inference policy mismatch. 

We trained on 16 rollouts per example sampled at a temperature of 1.0, a batch size 512, and a sequence length of 4096. We used the AdamW optimizer~\citep{loshchilov2019decoupled} with a learning rate of 1e-5, weight decay of 0.01, gradient clipping of 1.0, a max gradient norm of 1.0, and beta\_1 = 0.9 and beta\_2 = 0.999. We evaluatated on 1 rollout sampled from the test set every 10 steps.

For MedCalc-Bench Verified v1.0.4, we used the thresholds provided by the original paper. Equation-based calculators which output a decimal were accepted provided the answer was within a 5\% threshold. Rule-based calculators and equation-based calculations involving dates required exact-matching dates, durations, or clinical scores. The MedMCQA reward formulation involved assigning a point if the model selected the correct answer to a multiple choice question. Last, for MedCaseReasoning, the reward formulation awarded a point if the LLM-as-judge deems the answer as being correct or 0 points if the answer is incorrect. Following the implementation in  \textsc{Medmarks-OE} we used GPT 5 nano as the judge.

\section{MedCalcBench with Tools}
\label{app:medcalcbench_tools}

\begin{figure}[htbp!]
  \centering
  \includegraphics[width=0.9\columnwidth]{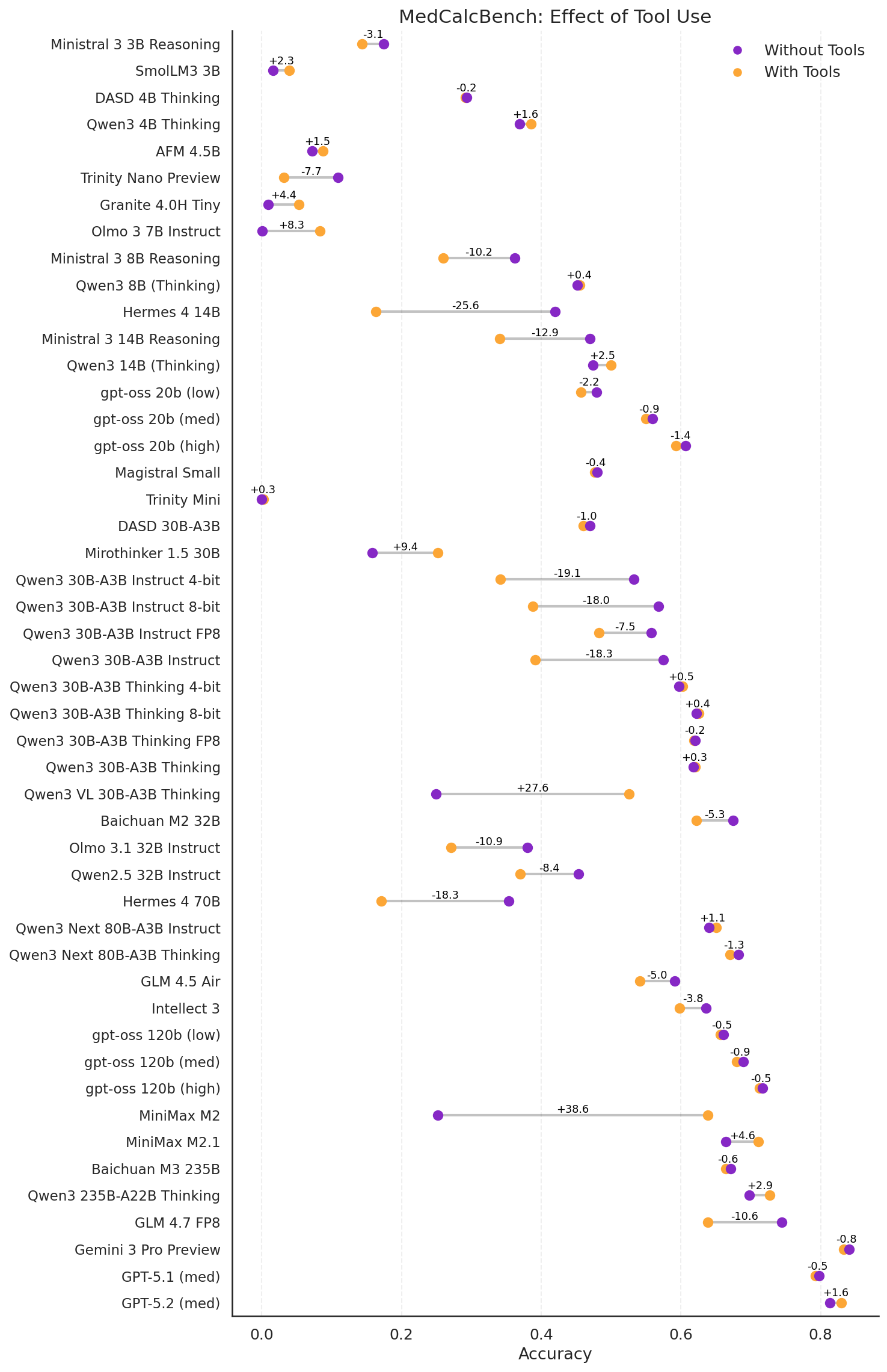}
  \caption{MedCalcBench with and without tools.}
  \label{fig:medcalcbench_tools}
\end{figure}

In addition to evaluating all models on MedCalcBench, we evaluated a subset of models with vLLM working tool templates with both a restricted code interpreter and a simple calculator tool. The results are shown below in \cref{fig:medcalcbench_tools}. Giving models access to tools results in a wide range of outcomes: significant increase in performance, slight changes in performance likely due to LLM sampling non-determinism, and some significant decreases in performance. Models with the largest improvement in calculation accuracy are MiniMax M2, Qwen3 VL, Mirothinker 1.5 30B, and Olmo 3 7B. Both sizes of gpt-oss across all reasoning levels and all Qwen Thinking models (except Qwen3 235B-A22B Thinking) had minimal changes in performance which appear to be sampling noise.

\cref{tab:medcalcbench_tools} shows the taxonomy of the regressions using the following categories:

\begin{itemize}
  \item No Tools: Model made zero tool calls
  \item Incorrect: Model called tools and produced a parseable but incorrect answer.
  \item Formatting: The final message did not have a parsable answer format.
  \item Ignored Tool: Model used a tool to generate an incorrect answer, ignored it, and submitted a different incorrect answer.
  \item Thrashing: Model made 5+ tool calls and still got the wrong answer.
  \item Malformed: Catch all for incorrectly attempting to use tools: bad tool formatting, tool calls with errors, hallucinated tool usage, etc.
  \item Changed Result: Model called a tool which returned a correct value, but ignored it and submitted different answer.
\end{itemize}

The overwhelming majority of models regressed by either never using tools and returning the incorrect answer or using tools but calculating the incorrect answer.

With the exception of Qwen Next, Qwen Instruct models appear to become confused with tools across both generations and all quantization levels, either changing the correct answer from a tool result or using tools but deriving the incorrect answer. Spot checking the Olmo 3 Instruct series of models, including Olmo 3 7B which improved its score with tools, had difficulty consistently formatting their tool calling correctly, which could be a chat template, tool parser, or training issue. Baichuan M3, which is fine-tuned from Qwen3 235B-A22B Thinking, appears to have lost tool calling capabilities during additional training. High performing models such as Gemini 3 Pro, gpt-oss 120b, and Qwen3 Next appear to be overconfident with not needing tool usage or replacing the tool's result. With GPT-5.1 and gpt-oss models, this could be due to using the old chat completions api and the models losing their reasoning traces in-between calls. Many models struggled with instruction following and incorrectly formatted their answers when given tools, including GLM-4.7 FP8 whose majority of regressions were due to an unparseable answer format. While it didn't happen often, Gemini 3 Pro Preview rejected the correct answer from a tool call thirteen times.

Further experimentation is needed on available tools, such as MCP or API tool definitions, and prompt instructions for tool usage. 

\begin{landscape}
\begin{table}[ht]
\centering
\caption{MedCalcBench with tools regressions}
\label{tab:medcalcbench_tools}
\begin{tabular}{l l r r r r r r r r}
\toprule
Model & Size & Num Regr. & No Tools & Incorrect & Formatting & Ignored Tool & Malformed & Thrashing & Changed Result \\
\midrule
Ministral 3 3B Reasoning & Tiny & 116 & \textbf{80.2\%} & 15.5\% & 0.0\% & 1.7\% & 1.7\% & 0.0\% & 0.9\% \\
DASD 4B Thinking & Tiny & 128 & \textbf{98.4\%} & 0.0\% & 0.0\% & 0.0\% & 1.6\% & 0.0\% & 0.0\% \\
Trinity Nano Preview & Tiny & 107 & 1.9\% & 18.7\% & 7.5\% & 5.6\% & 21.5\% & \textbf{44.9\%} & 0.0\% \\
Ministral 3 8B Reasoning & Small & 200 & 23.5\% & \textbf{41.0\%} & 29.0\% & 2.5\% & 2.0\% & 2.0\% & 0.0\% \\
Hermes 4 14B & Small & 339 & \textbf{86.7\%} & 1.8\% & 7.1\% & 0.3\% & 4.1\% & 0.0\% & 0.0\% \\
Ministral 3 14B Reasoning & Small & 236 & 15.3\% & 31.4\% & \textbf{43.2\%} & 3.0\% & 2.5\% & 4.2\% & 0.4\% \\
gpt-oss 20b (low) & Medium & 140 & \textbf{67.9\%} & 13.6\% & 7.9\% & 2.9\% & 3.6\% & 3.6\% & 0.7\% \\
gpt-oss 20b (med) & Medium & 114 & \textbf{64.0\%} & 15.8\% & 5.3\% & 4.4\% & 6.1\% & 0.9\% & 3.5\% \\
gpt-oss 20b (high) & Medium & 116 & \textbf{50.0\%} & 20.7\% & 10.3\% & 6.0\% & 3.4\% & 6.0\% & 3.4\% \\
Magistral Small & Medium & 145 & 41.4\% & \textbf{44.1\%} & 2.8\% & 6.2\% & 4.8\% & 0.7\% & 0.0\% \\
DASD 30B-A3B & Medium & 113 & \textbf{95.6\%} & 1.8\% & 0.0\% & 0.0\% & 2.7\% & 0.0\% & 0.0\% \\
Qwen3 30B-A3B Instruct 4-bit & Medium & 275 & 2.5\% & \textbf{62.2\%} & 1.8\% & 21.1\% & 9.5\% & 1.8\% & 1.1\% \\
Qwen3 30B-A3B Instruct 8-bit & Medium & 265 & 4.9\% & \textbf{69.1\%} & 3.8\% & 15.5\% & 3.8\% & 2.6\% & 0.4\% \\
Qwen3 30B-A3B Instruct FP8 & Medium & 185 & 24.9\% & \textbf{53.5\%} & 4.9\% & 10.8\% & 3.8\% & 1.6\% & 0.5\% \\
Qwen3 30B-A3B Instruct & Medium & 270 & 7.0\% & \textbf{67.8\%} & 3.3\% & 15.6\% & 5.2\% & 1.1\% & 0.0\% \\
Qwen3 30B-A3B Thinking FP8 & Medium & 68 & \textbf{66.2\%} & 30.9\% & 0.0\% & 0.0\% & 0.0\% & 0.0\% & 2.9\% \\
Baichuan M2 32B & Medium & 120 & 16.7\% & \textbf{28.3\%} & 20.8\% & 10.0\% & 17.5\% & 0.0\% & 6.7\% \\
Granite 4.0H Small & Medium & 295 & \textbf{98.0\%} & 0.0\% & 0.0\% & 0.0\% & 2.0\% & 0.0\% & 0.0\% \\
Olmo 3.1 32B Instruct & Medium & 223 & \textbf{89.2\%} & 6.7\% & 0.4\% & 0.0\% & 1.3\% & 2.2\% & 0.0\% \\
Qwen2.5 32B Instruct & Medium & 197 & 3.0\% & \textbf{82.2\%} & 0.0\% & 1.0\% & 9.6\% & 4.1\% & 0.0\% \\
Hermes 4 70B & Large & 278 & 0.7\% & \textbf{50.4\%} & 34.2\% & 4.7\% & 8.6\% & 1.1\% & 0.4\% \\
Qwen3 Next 80B-A3B Thinking & Large & 70 & \textbf{85.7\%} & 7.1\% & 0.0\% & 1.4\% & 5.7\% & 0.0\% & 0.0\% \\
GLM 4.5 Air & Large & 155 & 3.9\% & \textbf{45.8\%} & 28.4\% & 3.2\% & 0.0\% & 16.1\% & 2.6\% \\
Intellect 3 & Large & 169 & \textbf{29.0\%} & 27.2\% & 18.9\% & 9.5\% & 13.6\% & 0.0\% & 1.8\% \\
gpt-oss 120b (low) & Large & 96 & \textbf{67.7\%} & 19.8\% & 9.4\% & 0.0\% & 3.1\% & 0.0\% & 0.0\% \\
gpt-oss 120b (med) & Large & 89 & \textbf{43.8\%} & 29.2\% & 16.9\% & 5.6\% & 1.1\% & 2.2\% & 1.1\% \\
gpt-oss 120b (high) & Large & 82 & \textbf{46.3\%} & 37.8\% & 3.7\% & 2.4\% & 3.7\% & 3.7\% & 2.4\% \\
Baichuan M3 235B & Large & 77 & \textbf{96.1\%} & 0.0\% & 0.0\% & 0.0\% & 3.9\% & 0.0\% & 0.0\% \\
GLM 4.7 FP8 & Large & 219 & 23.7\% & 24.7\% & \textbf{46.6\%} & 2.7\% & 1.4\% & 0.9\% & 0.0\% \\
Gemini 3 Pro Preview & API & 44 & 2.3\% & \textbf{52.3\%} & 0.0\% & 13.6\% & 2.3\% & 0.0\% & 29.5\% \\
GPT-5.1 (med) & API & 54 & \textbf{44.4\%} & 35.2\% & 0.0\% & 9.3\% & 1.9\% & 1.9\% & 7.4\% \\
\bottomrule
\end{tabular}
\end{table}
\end{landscape}

\section{Sample LLM-as-a-Judge Prompt}
\label{app:sample_judge_prompt}

Where benchmarks did not have pre-existing author or community LLM-as-a-Judge prompts, we created custom prompts informed by No Free Labels \citep{krumdick2025no} and industry best practices. These prompts are lightly modified for each specific benchmark. An example of one of our new prompts is shown below. 

\begin{promptbox}[title=CareQA Open Grading Prompt]
You are grading an AI assistant's answer to a medical/science exam question using a multi-axis rubric.
Input:
- <question>: The exam question.
- <reference_answer>: The reference answer.
- <assistant_answer>: The AI's response to grade.

Task:
Evaluate the assistant's answer on four Boolean dimensions and output your assessment in the specified format.

Grading Rules:
- Assume the reference answer is a correct but often incomplete exam solution.
- Focus on factual content and meaning, not style, length, or confidence.

Rubric:

1. Semantically Correct (true/false)
- True if the assistant expresses the same core claim(s) as the reference.
- Allow synonyms, paraphrasing, acronyms, and reasonable generalizations that still unambiguously answer the question correctly.
- False if the main concept/mechanism/entity/relationship differs or if the answer is too vague to establish the reference's core claim(s).

2. Matches Details (true/false)
- True if the assistant includes all question-critical details needed to uniquely match the reference answer. Ignore extra illustrative or optional context in the reference.
- False if any required specifics or details from the reference are missing, overgeneralized where precision matters, or incorrect.
- Constraint: If Semantically Correct is false, Matches Details must be false.

3. Substantive Addition (true/false)
- True if the assistant introduces factual claim(s) that could meaningfully alter correctness assessment: tangential or off-topic content, claims or details beyond the question's scope, or alternative explanations/approaches not consistent with the reference.
- False for definitions, brief clarifying context, stylistic elaboration, standard supporting details directly tied to the reference answer, or added specificity that elaborates the same core answer rather than introducing new topics.
- False if the reference answer is incomplete relative to what the question explicitly asks and the assistant provides additional content to fully address the question's stated requirements.

4. Critical Error (true/false)
- True if the assistant states any factual claim that is clearly false relative to the reference and/or standard domain knowledge, or gives unsafe medical guidance.
- False if no clearly incorrect, contradictory, unsafe, or fabricated factual claims are present.
- Note: Missing information alone is not a critical error (it affects Matches Details).
- Note: Critical Error and Substantive Additions are independent; an incorrect added claim may make both true.

<question>{question}</question>
<reference_answer>{answer}</reference_answer>
<assistant_answer>{response}</assistant_answer>

Instructions:
- Briefly compare assistant vs reference for each rubric dimension.
- Output in this exact format:

<analysis>
[Brief dimension-by-dimension analysis]
</analysis>
<semantically_correct>[true/false]</semantically_correct>
<matches_details>[true/false]</matches_details>
<substantive_addition>[true/false]</substantive_addition>
<critical_error>[true/false]</critical_error>
\end{promptbox}

\section{Preliminary MedAgentBench V2}

\begin{figure}[htbp!]
  \centering
  \includegraphics[width=0.9\columnwidth]{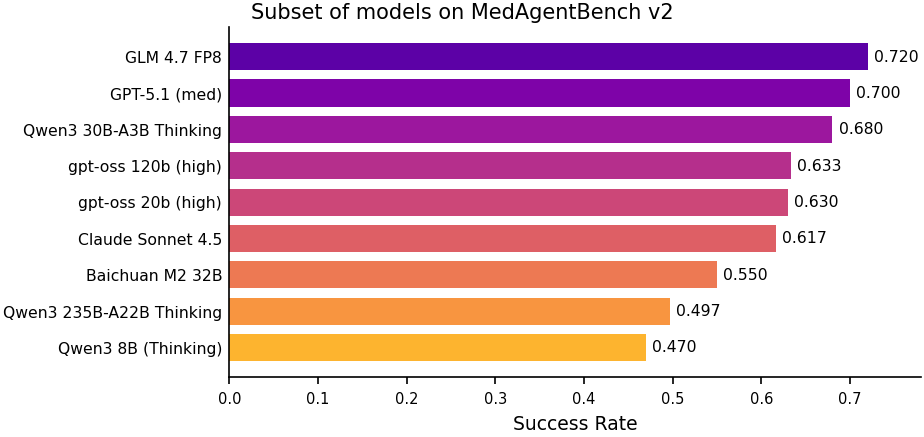}
  \caption{Subset of models on MedAgentBench V2.}
  \label{fig:medagentbenchv2}
\end{figure}

We evaluated a subset of the \textsc{Medmarks-OE} models on MedAgentBench V2 and present the results in \cref{fig:medagentbenchv2}. MedAgentBench tests the ability of models to navigate EHR records, and MedAgentBench V2 upgrades the benchmark with a modern tool-based approach, in addition to adding new questions. 

There are a couple caveats to this evaluation. First we evaluated the same set of questions from the original MedAgentBench V1 set and didn't include the additional V2 questions. Second, at the time of evaluation verifiers did not support replaying the reasoning traces of models which require it, like OpenAI's GPT models, which might explain part of their underperformance on this multi-tool use benchmark. This is a degradation from the original MedAgentBench V2 code. We plan on rectifying these issues in a future update.

This benchmark has two sets of notable results: first, GLM 4.7 narrowly outperforms both GPT-5.1 medium and significantly outperforms Claude Sonnet 4.5, and Qwen3 235B-A22B Thinking significantly underperforms Qwen3 30B-A3B Thinking. \cref{tab:medagentbenchv2} shows a taxonomy of the differences between GPT 5.1 and GLM 4.7 and the two Qwen3 models.

The small sample size of this benchmark, 300 cases, and success rate on querying the initial patient suggests that future evaluations should average multiple runs to better measure model performance.

\begin{table}[htbp!]
\centering
\scriptsize
\caption{Failure-mode classification of MedAgentBench v2 pairwise disagreements from the raw runs. Each row summarizes examples that the listed winner solved and the listed loser missed; shares are computed within each pair.}
\label{tab:medagentbenchv2}
\begin{tabular}{llrr}
\toprule
Pair & Failure Mode & Count & Share \\
\midrule
GLM 4.7 FP8 over GPT-5.1 (med) & Wrong search scope/category & 25 & 67.6\% \\
 & Winner placed order; loser abstained & 7 & 18.9\% \\
 & Missed required order action & 2 & 5.4\% \\
 & Other disagreement & 2 & 5.4\% \\
 & Tool/query formatting error & 1 & 2.7\% \\
Qwen3 30B-A3B Thinking over Qwen3 235B-A22B Thinking & Wrong patient reference & 51 & 72.9\% \\
 & Tool/query formatting error & 10 & 14.3\% \\
 & Wrong search scope/category & 9 & 12.9\% \\
\bottomrule
\end{tabular}
\end{table}

\clearpage
\section{Additional Figures}

\begin{figure}[!ht]
  \centering
  \includegraphics[width=0.825\columnwidth]{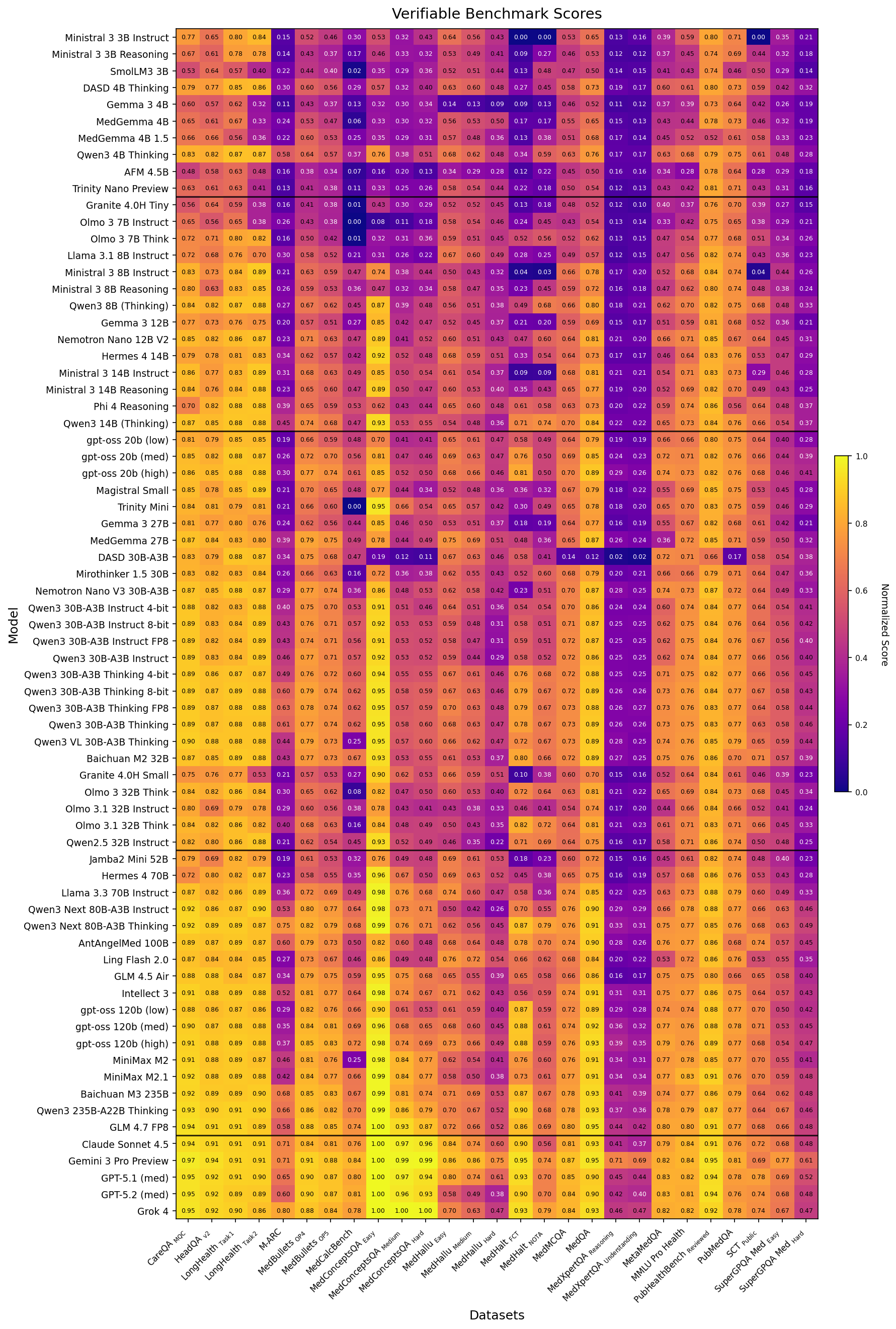}
  \caption{Heatmap table of the raw scores for each model across the 19 benchmarks of the \textsc{Medmarks-V} subset. Dark purple highlights low performance, bright yellow highlights high performance. Metrics are dependent on the benchmark.}
  \label{fig:benchmark_heatmap}
\end{figure}

\begin{figure}[htbp!]
  \centering
  \includegraphics[width=\columnwidth]{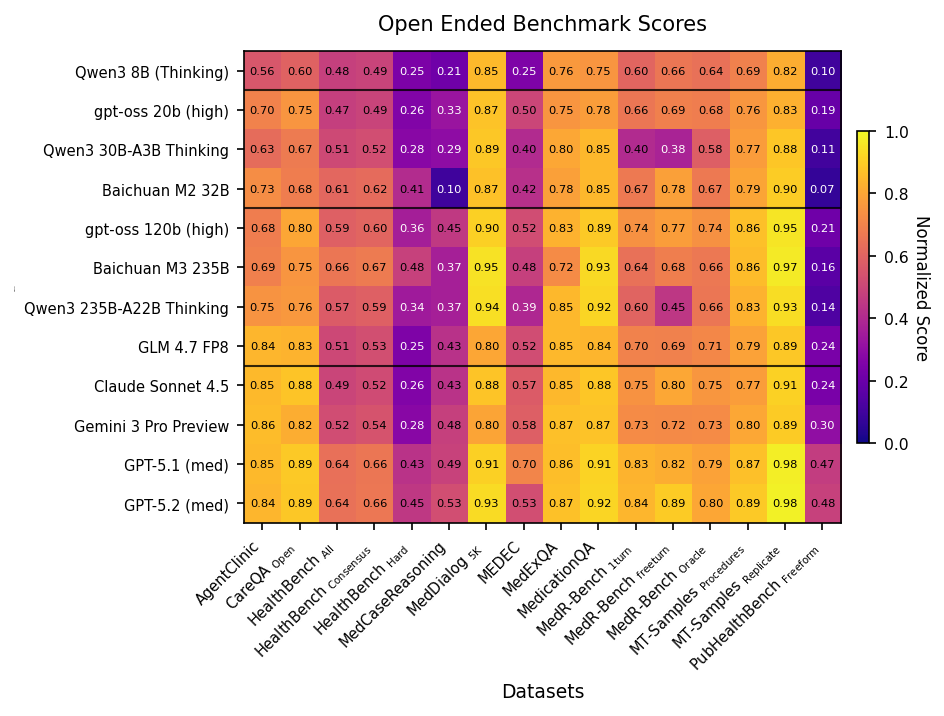}
  \caption{Heatmap table of the raw scores for each model across 11 benchmarks of the \textsc{Medmarks-OE} subset. Dark purple highlights low performance, bright yellow highlights high performance. Metrics vary by benchmark; the reported metric is a normalized LLM Judge score.}
  \label{fig:llm_judge_heatmap}
\end{figure}

\begin{figure*}[htbp!]
  \begin{center}
    \centerline{\includegraphics[width=0.85\columnwidth]{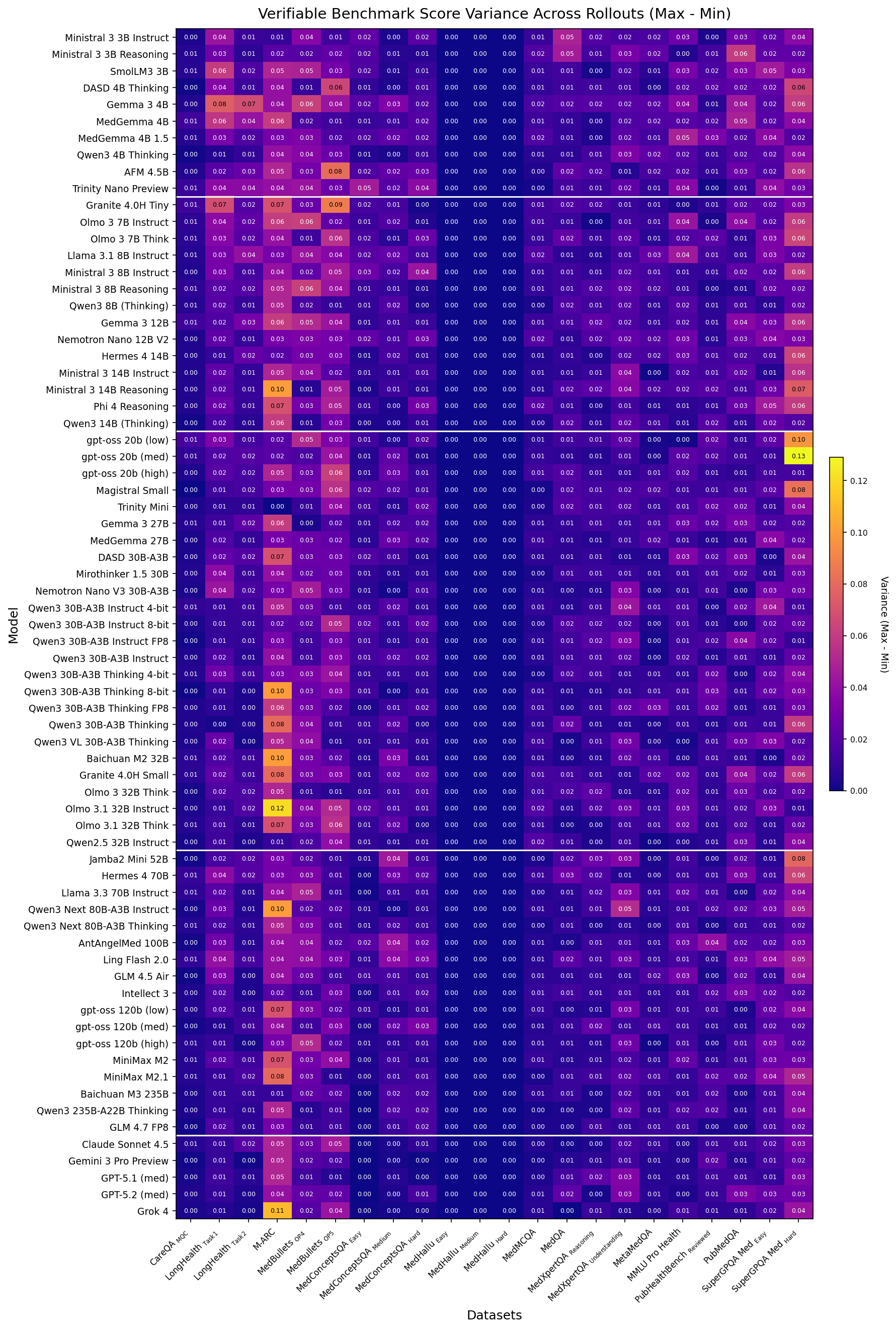}}
    \caption{Variance of model performance across all \textsc{Medmarks-V} benchmarks. Dark purple highlights low spread, bright yellow highlights high spread. Given we typically evaluate a model three times on each benchmark, we report the maximum score subtracted by the minimum score for a given benchmark. Note that since we only performed a single evaluation for HeadQA-v2 \cite{correa2025head}, MedCalc-Bench \cite{khandekar2024medcalc}, and Med-HALT \cite{pal2023med}, we omit them from the table.}
    \label{benchmark_variance_heatmap}
  \end{center}
\end{figure*}

\clearpage
\begin{landscape}

\begin{figure*}[ht]
  \begin{center}
    \centerline{\includegraphics[width=\columnwidth]{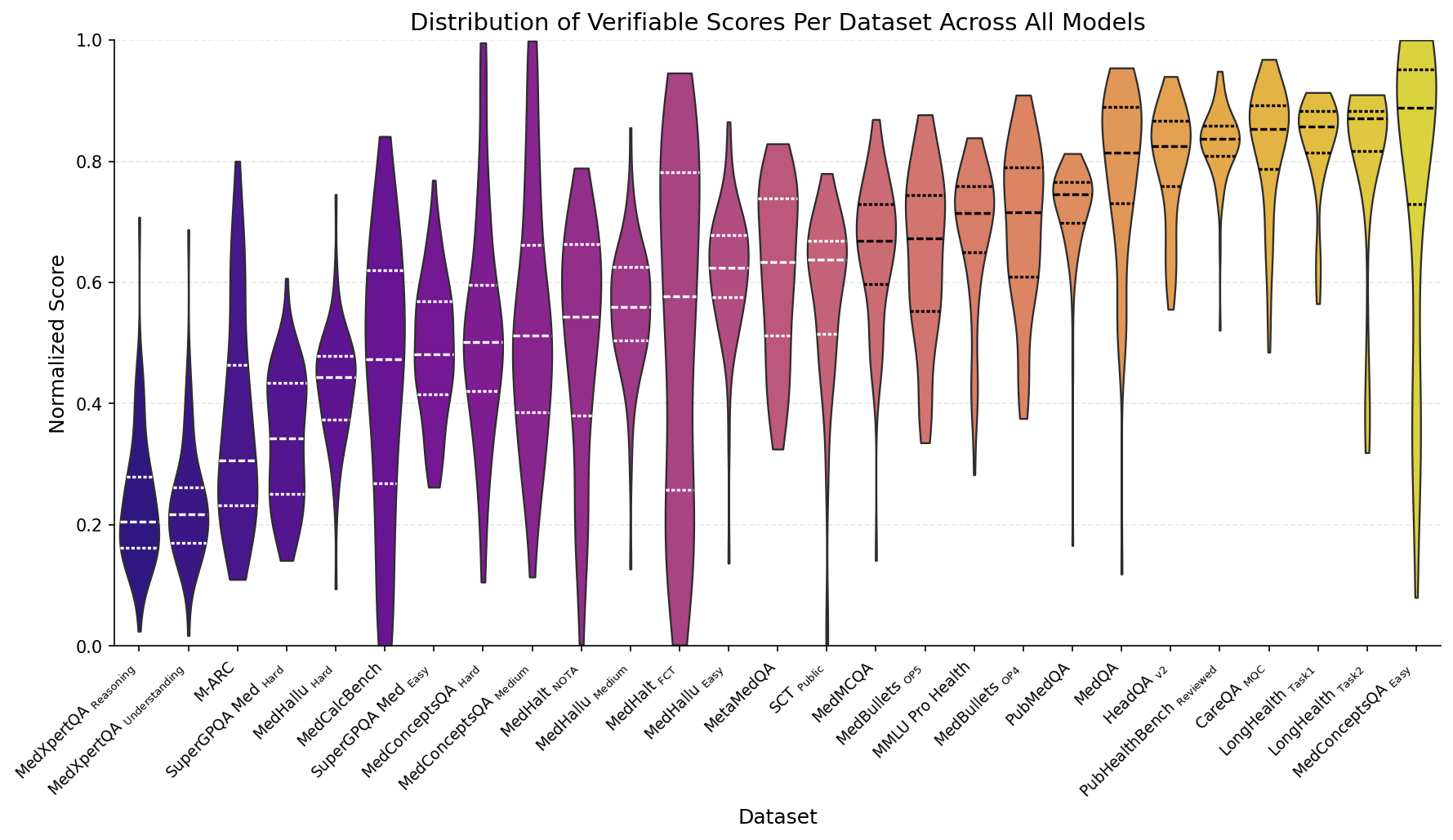}}
    \caption{Distribution of normalized model performance for each of the datasets in the \textsc{Medmarks-V} subset, across all 61 models across 71 variants tested.}
    \label{benchmark_scores_violin}
  \end{center}
\end{figure*}

\begin{figure*}[ht]
  \begin{center}
    \centerline{\includegraphics[width=\columnwidth]{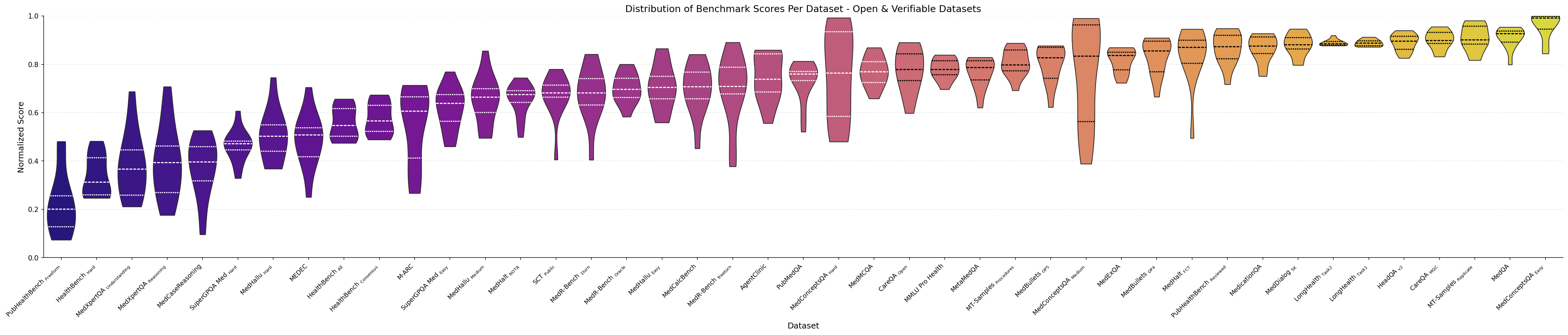}}
    \caption{Distribution of normalized model performance for each of the datasets in both the \textsc{Medmarks-V} and \textsc{Medmarks-OE} subsets across the 12 models that were evaluated on both subsets.}
    \label{benchmark_scores_violin_llm_models}
  \end{center}
\end{figure*}

\begin{figure*}[ht]
  \begin{center}
    \centerline{\includegraphics[width=\columnwidth]{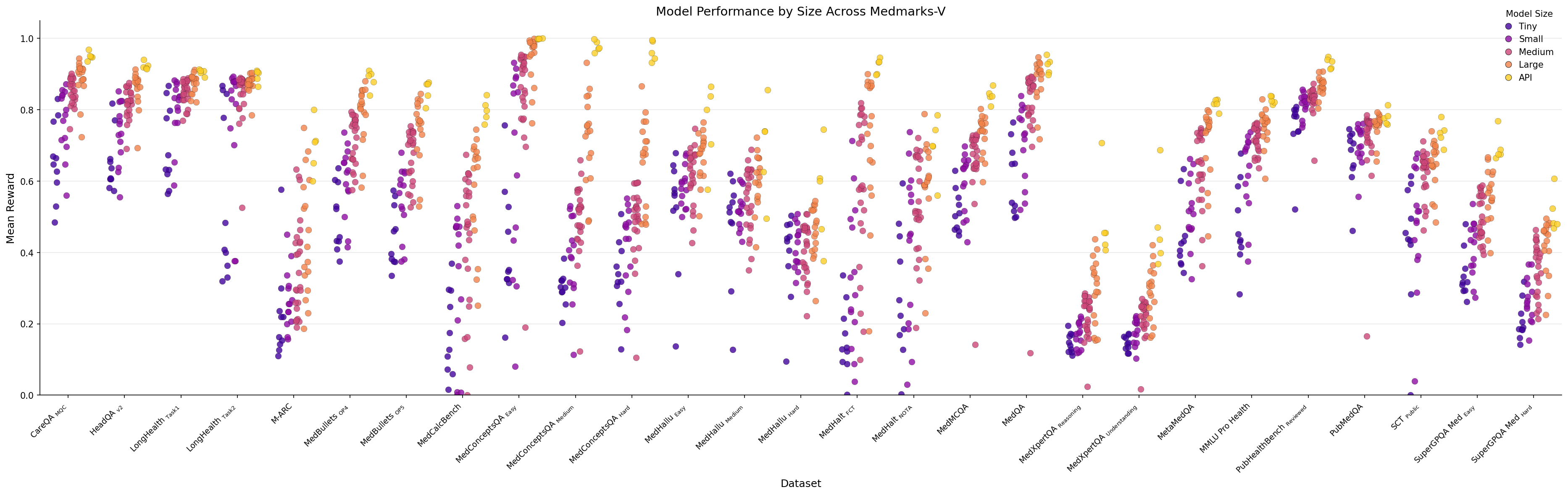}}
    \caption{Scatter plot of model scores on each of the \textsc{Medmarks-V} benchmarks, labeled by model size.}
    \label{performance_by_size_scatter}
  \end{center}
\end{figure*}

\begin{figure*}[ht]
  \begin{center}
    \centerline{\includegraphics[width=\columnwidth]{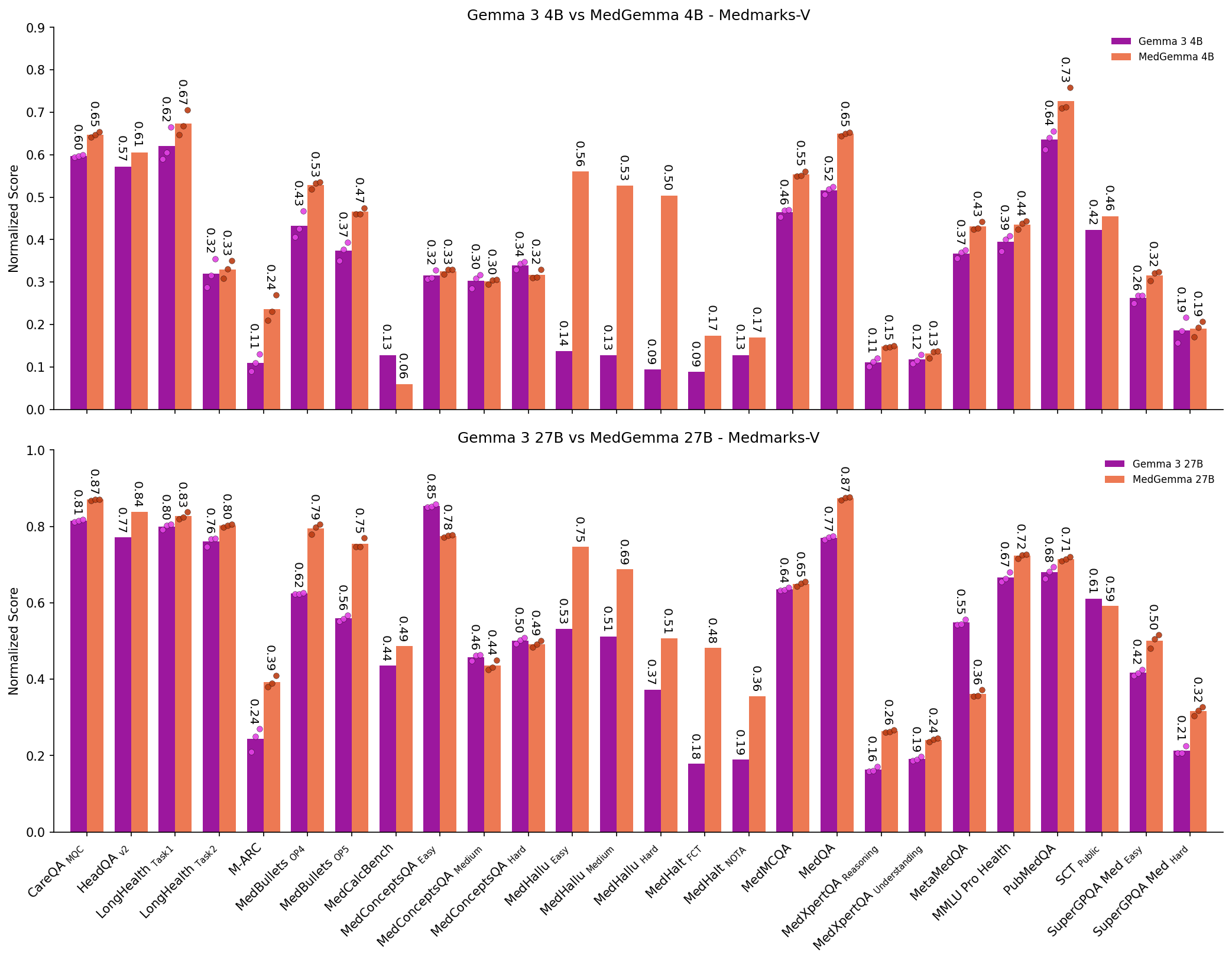}}
    \caption{Comparing the performance of Gemma 3 models to MedGemma 3 models on \textsc{Medmarks-V} tasks.}
    \label{gemma_comparison_bar}
  \end{center}
\end{figure*}

\begin{figure*}[ht]
  \begin{center}
    \centerline{\includegraphics[width=\columnwidth]{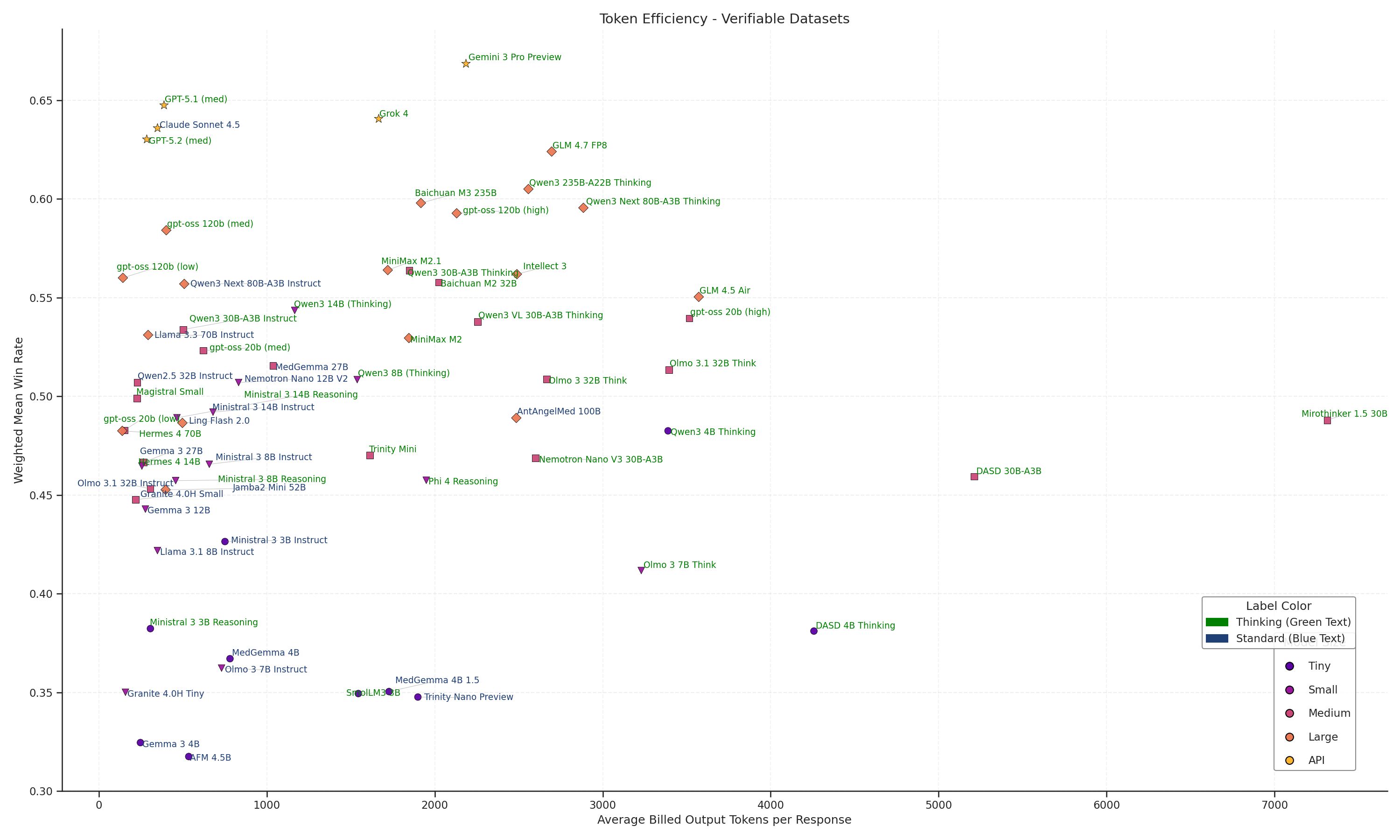}}
    \caption{Scatter plot of weighted mean win rate on \textsc{Medmarks-V} by average tokens per response for each model. Each model is labeled by model size and whether it is a thinking model or standard model.}
    \label{token_efficiency}
  \end{center}
\end{figure*}

\begin{figure*}[ht]
  \begin{center}
    \centerline{\includegraphics[width=\columnwidth]{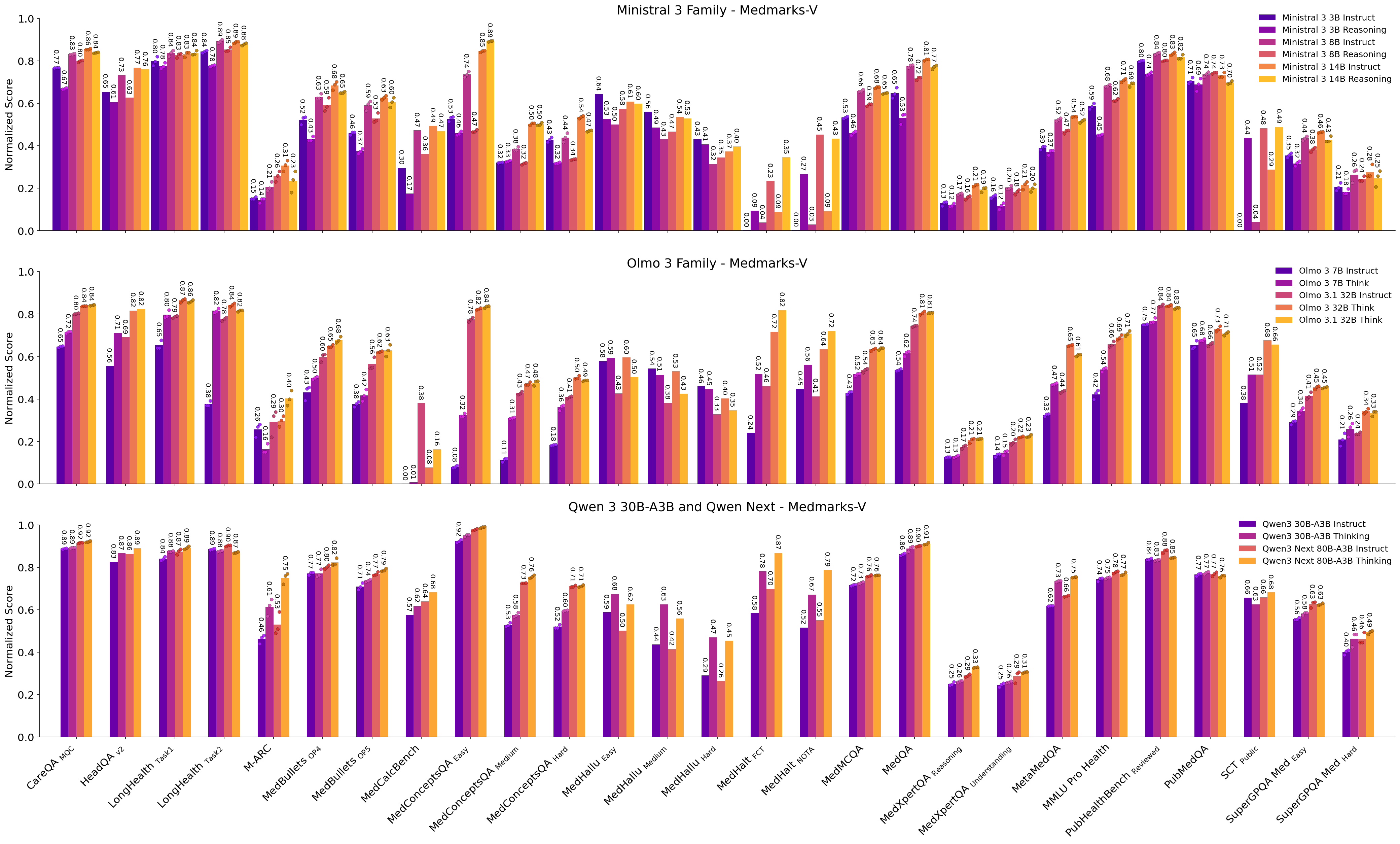}}
    \caption{Bar plots comparing performance of instruct vs. reasoning models for Ministral 3, Olmo 3, and Qwen3 models.}
    \label{model_family_comparison_bar}
  \end{center}
\end{figure*}

\begin{figure*}[ht]
  \begin{center}
    \centerline{\includegraphics[width=\columnwidth]{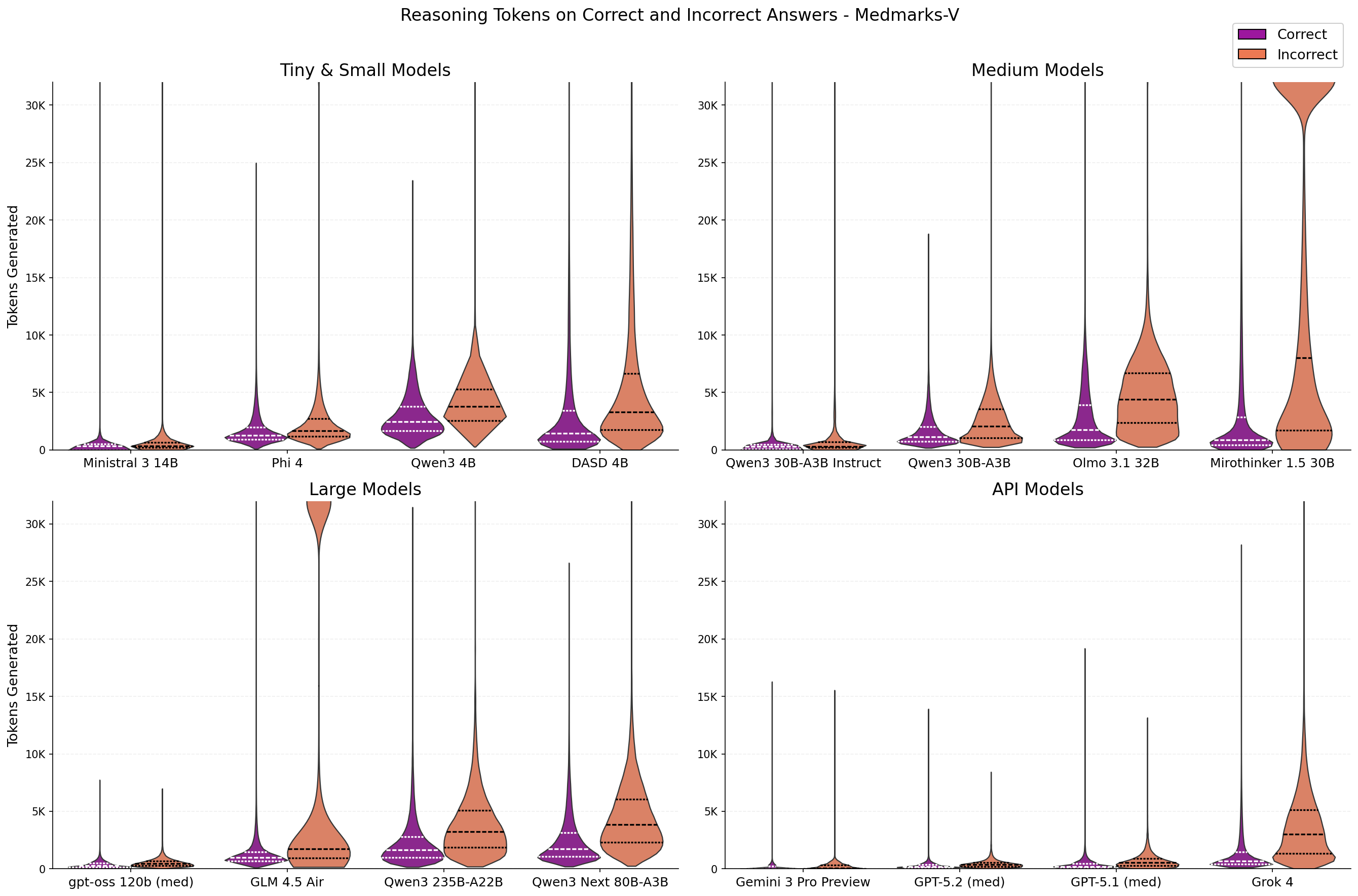}}
    \caption{Distribution of number of tokens generated for different models when the response is correct or incorrect.}
    \label{thinking_length_violin}
  \end{center}
\end{figure*}

\begin{figure*}[ht]
  \begin{center}
    \centerline{\includegraphics[width=\columnwidth]{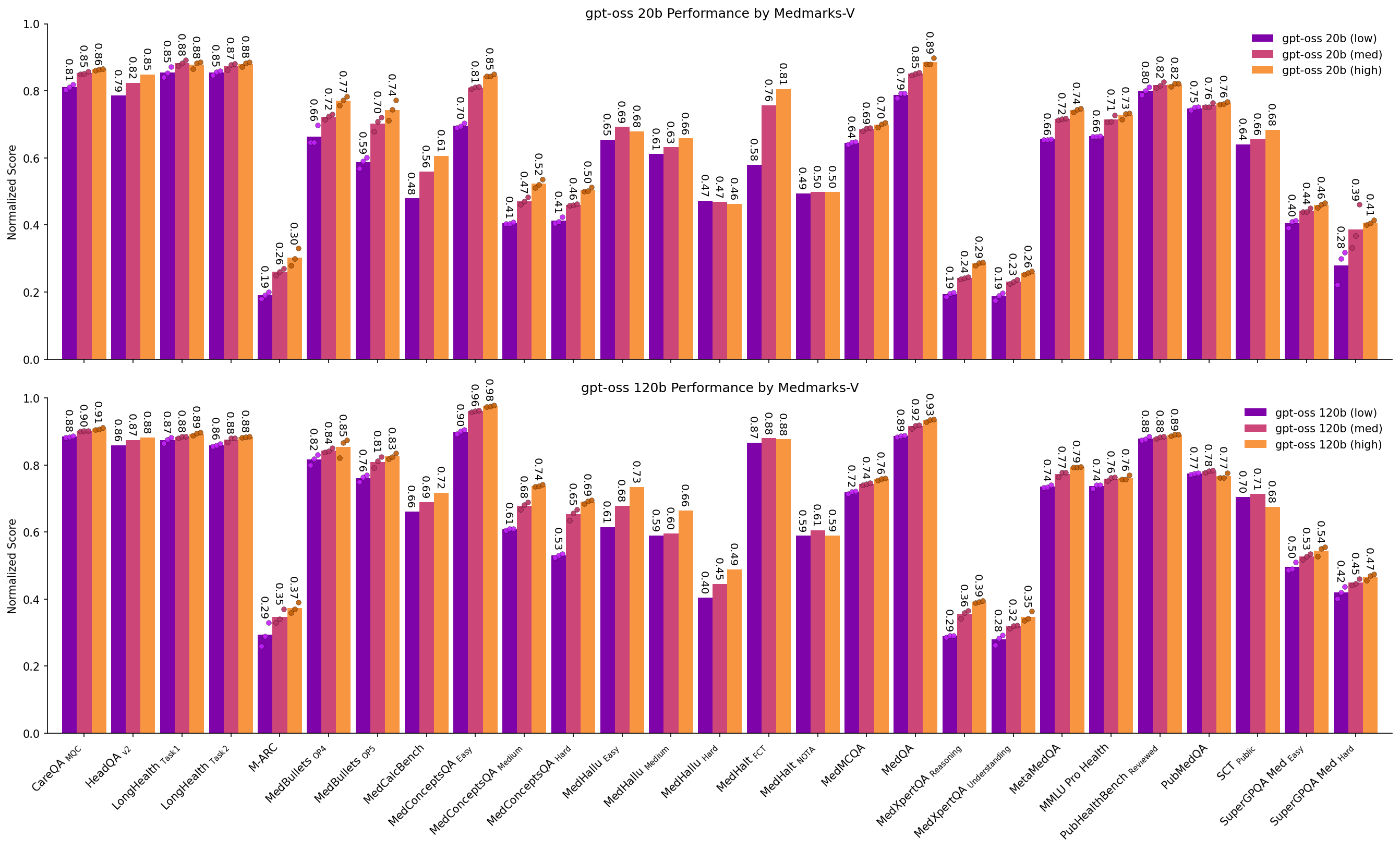}}
    \caption{Bar plots comparing performance of different reasoning levels for gpt-oss models on the \textsc{Medmarks-V} benchmarks.}
    \label{gpt_oss_bar_comparison}
  \end{center}
\end{figure*}

\begin{figure*}[ht]
  \begin{center}
    \centerline{\includegraphics[width=\columnwidth]{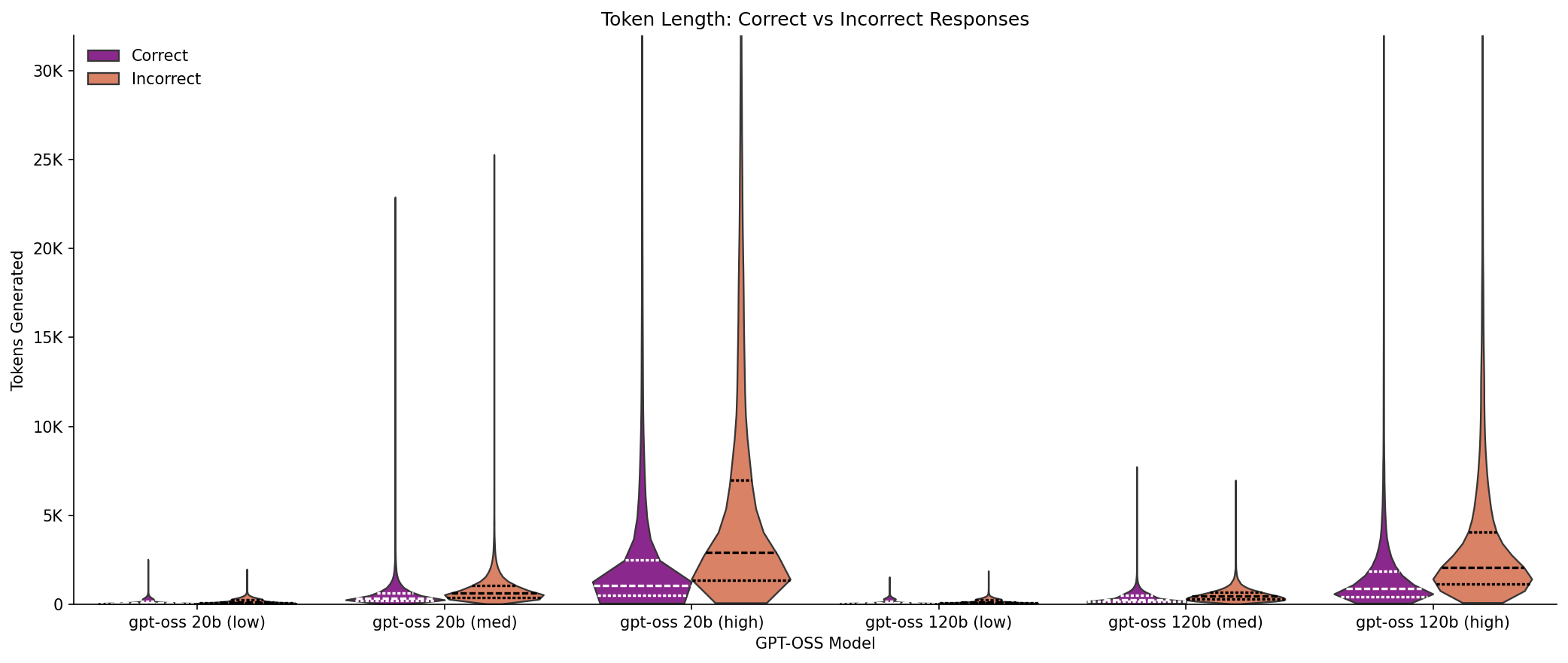}}
    \caption{Distribution of number of tokens generated for gpt-oss reasoning levels models when the response is correct or incorrect.}
    \label{gpt_oss_violin}
  \end{center}
\end{figure*}

\begin{figure*}[ht]
  \begin{center}
    \centerline{\includegraphics[width=\columnwidth]{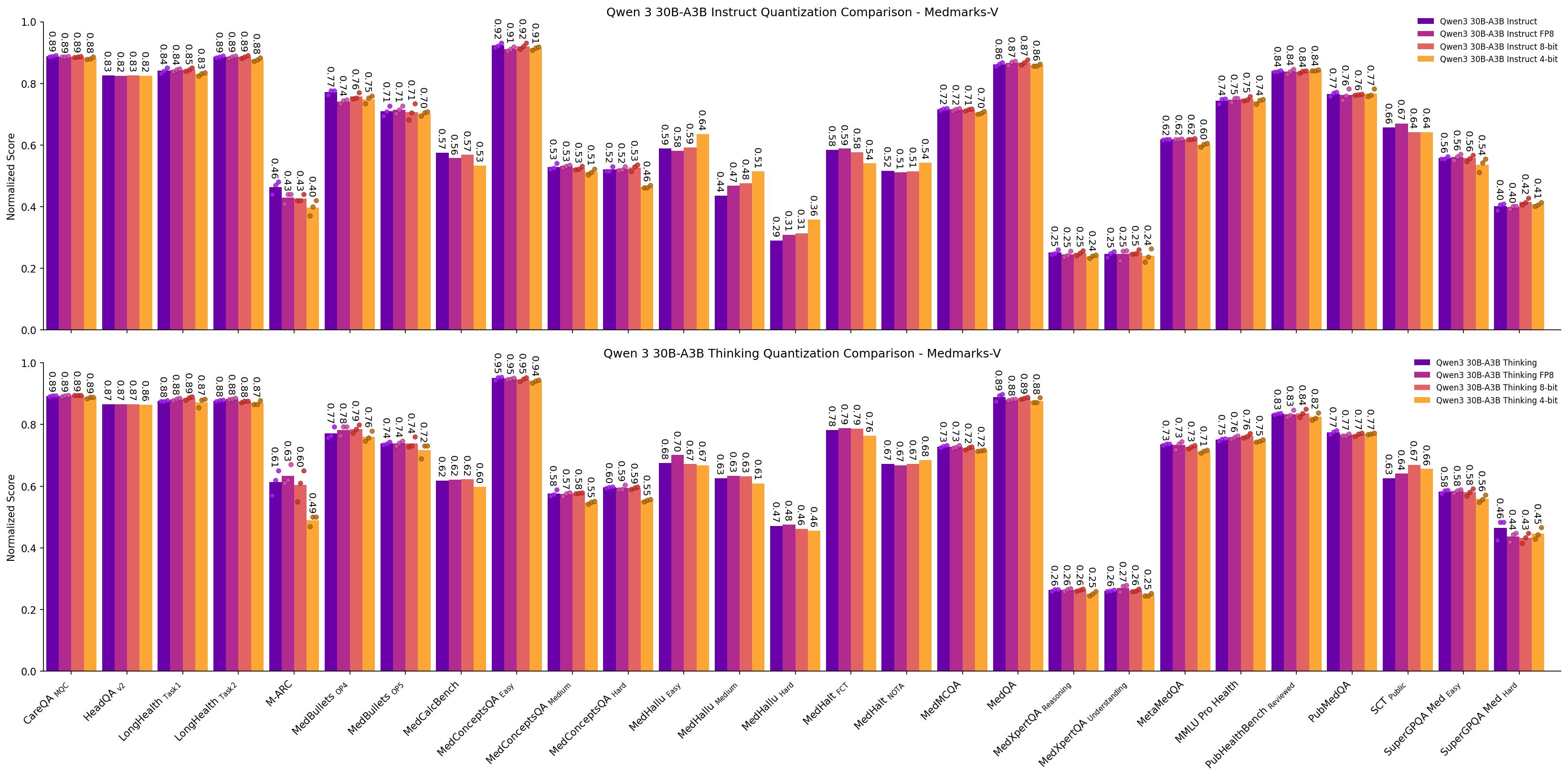}}
    \caption{Bar plots comparing performance of quantization levels for Qwen3 models on the \textsc{Medmarks-V} benchmarks.}
    \label{qwen_quantization_comparison_bar}
  \end{center}
\end{figure*}

\end{landscape}

\begin{figure}[!t]
  \centering
  \centerline{\includegraphics[width=0.73\columnwidth]{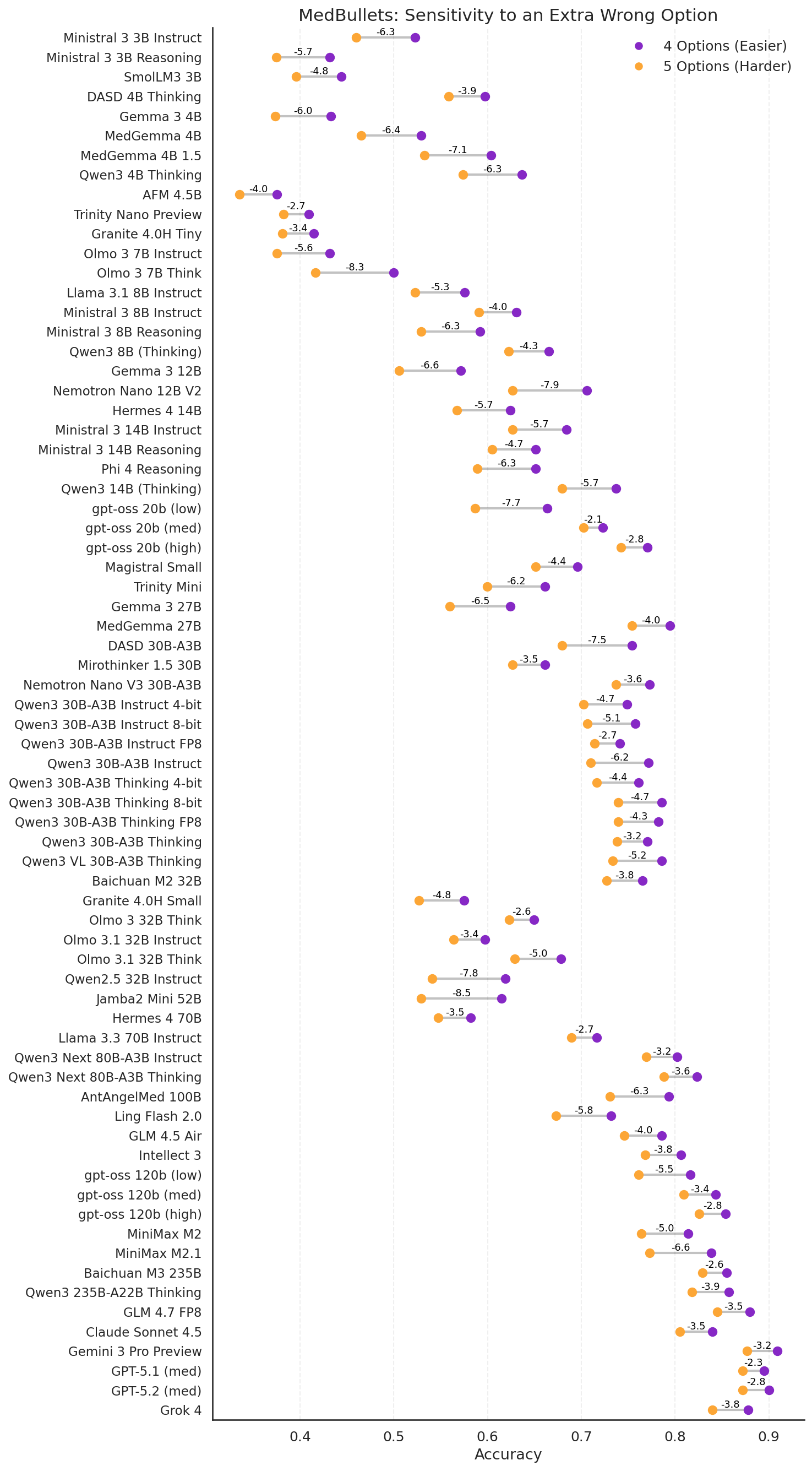}}
\caption{Comparing all model performance with and without an extra option on the Medbullets \cite{chen2025benchmarking} benchmark.}
\label{distractor_stress_test_all_models}
\end{figure}

\newpage
\onecolumn
\section{Qualitative question analysis}

In order to assess the quality and clinical validity of commonly used medical question answering benchmarks, we conducted a qualitative analysis of samples drawn from CareQA, MedQA, PubMedQA, and MedMCQA. A practicing physician manually reviewed each sample and evaluated whether the question was clearly formulated, contained sufficient clinical context to be answerable, and whether the proposed answer choices were correct and clinically appropriate. As illustrated in \cref{tab:multimedqa:medmcqa_samples,tab:multimedqa:medqa_samples}, many questions are malformed, ambiguous, or rely on missing information such as images, summaries, or units, rendering them impossible to answer as written. \cref{tab:careqa_samples,tab:multimedqa:pubmedqa_samples} further show that some question–answer pairs are incomplete, clinically misleading, or lack medical relevance.

\begin{table}[ht!]
    \centering
    \caption{Examples of samples needing clarity in MedMCQA.}
    \label{tab:multimedqa:medmcqa_samples}
    \footnotesize
    \setlength{\tabcolsep}{3pt}
    \renewcommand{\arraystretch}{1.2}
    \begin{tabular}{@{}p{0.12\columnwidth}p{0.48\columnwidth}p{0.34\columnwidth}@{}}
        \toprule
        \textbf{Question Id} & \textbf{Question and/or Answer Choices} & \textbf{Physician Analysis} \\
        \midrule
        5ce754b8... & Which of the following blade angle is appropriate for scaling and root planning (A) A (B) B (C) C (D) D & The question is unclear, punctuation is missing, and the answer options are missing. \\
        \addlinespace[2pt]
        0f810d3c... & Sho structured primi gravida has height less than (A) 140 cm (B) 145 cm (C) 150 cm (D) 135 cm & The question is unclear, punctuation is missing, we were unable to understand what the question is asking. \\
        \addlinespace[2pt]
        33bfa0d9... & HIV can - (A) Cross blood-brain barrier (B) RNA virus (C) Inhibited by 0.3\% H\textsubscript{2}O\textsubscript{2} (D) Thermostable & The choices are not compatible with the affirmation. \\
        \addlinespace[2pt]
        796a5e1c... & Blockers are indicated in (A) Phobia (B) Schizophrenia (C) Anxiety (D) Mania & The question is incomplete, ``Blockers'' by itself is not enough; we suppose they mean beta-blockers. \\
        \addlinespace[2pt]
        e33e61bf... & Screening area for trachoma is: (A) Below 5 years school child only (B) 1-9 years (C) 9-14 years (D) 5-15 years & English grammar errors and incomplete question compared to the possible answers. \\
        \addlinespace[2pt]
        17360c6c... & Concentration of tropicamide: (A) 0.01 (B) 0.02 (C) 0.03 (D) 0.04 & The concentration of tropicamide can be anything; we were unable to understand what this question is asking. Also, units are missing. \\
        \bottomrule
    \end{tabular}
\end{table}

\begin{table}[ht!]
    \centering
    \caption{Examples of samples that cannot be answered in MedQA.}
    \label{tab:multimedqa:medqa_samples}
    \footnotesize
    \setlength{\tabcolsep}{5pt}
    \renewcommand{\arraystretch}{1.25}
    \begin{tabular}{@{}p{0.68\columnwidth}p{0.27\columnwidth}@{}}
        \toprule
        \textbf{Question and/or Answer Choices} & \textbf{Physician Analysis} \\
        \midrule
        A 23-year-old woman comes to the physician because she is embarrassed about the appearance of her nails. She has no history of serious illness and takes no medications. She appears well. A photograph of the nails is shown. Which of the following additional findings is most likely in this patient? 
        & The question requires a photograph that is not present. \\
        \midrule
        A 63-year-old man comes to the emergency department because of retrosternal chest pain. He describes it as 7 out of 10 in intensity. He has coronary artery disease, hypertension, and type 2 diabetes mellitus. His current medications are aspirin, simvastatin, metformin, and enalapril. He has smoked one pack of cigarettes daily for 33 years. On arrival, his pulse is 136/min and irregular, respirations are 20/min, and blood pressure is 85/55 mm Hg. The lungs are clear to auscultation. Cardiac examination shows no abnormalities. An ECG is shown. Which of the following is the most appropriate next step in management? 
        & The ECG is missing; the question cannot be answered without it. \\
        \midrule
        Please refer to the summary above to answer this question. The authors of the study have decided to conduct a follow-up analysis on their data. They decide to stratify their results by CD4+ T-lymphocyte count at the time of diagnosis. Among patients with CD4+ cell counts below 200/mm\textsuperscript{3}, cART adherence was a significant predictor of DLBCL risk (RR = 0.52, p = 0.01). However, among patients with CD4+ cell counts above 200/mm\textsuperscript{3}, no relationship was found between DLBCL risk and cART adherence (RR = 0.96, p = 0.36). Which of the following explains for the difference observed between the two strata? 
        & The summary required to answer the question is absent. \\
        \bottomrule
    \end{tabular}
\end{table}

\begin{table}[ht!]
    \centering
    \caption{Examples of samples that cannot be answered in CareQA.}
    \label{tab:careqa_samples}
    \footnotesize
    \setlength{\tabcolsep}{3pt}
    \renewcommand{\arraystretch}{1.2}
    \begin{tabular}{@{}p{0.12\columnwidth}p{0.53\columnwidth}p{0.29\columnwidth}@{}}
        \toprule
        \textbf{Question Id} & \textbf{Question and/or Answer Choices} & \textbf{Physician Analysis} \\
        \midrule
        b0f52abd... & A 59-year-old man is admitted to the ICU due to acute hypoxemic respiratory failure secondary to severe community-acquired pneumonia. Due to respiratory failure, he requires sedation and connection to invasive mechanical ventilation. His occupational history includes working at a pig farm, so it is empirically decided to cover for methicillin-resistant Staphylococcus aureus until cultures from respiratory secretions are obtained. Which of the following antibiotics would you empirically start as monotherapy antimicrobial? (A) Piperacillin-tazobactam. (B) Ceftaroline. (C) Cefazolin. (D) Cefotaxime. & The question is forcing a monotherapy, which would be inappropriate~\cite{metlay_diagnosis_2019}. \\
        \midrule
        ffe582a6... & A 50-year-old diabetic patient presents to the emergency room with a fever of 39ºC and general discomfort, reporting anal pain for the past 5 days. Upon examination, a large, hot, and painful perianal tumor with skin necrosis is prominent. The treatment of choice is: (A) Surgical drainage-debridement. (B) Broad-spectrum antibiotic and wait for evolution. (C) Hospital admission for study. (D) Radiology-guided puncture-drainage. & While surgical drainage-debridement is correct, antibiotics should also be administered~\cite{mocanu_antibiotic_2019}. \\
        \bottomrule
    \end{tabular}
\end{table}

\begin{table}[ht!]
    \centering
    \caption{Examples of samples needing clarity or clinical relevance in PubMedQA.}
    \label{tab:multimedqa:pubmedqa_samples}
    \small
    \setlength{\tabcolsep}{4pt}
    \renewcommand{\arraystretch}{1.25}
    \begin{tabular}{@{}p{0.14\columnwidth}p{0.48\columnwidth}p{0.32\columnwidth}@{}}
        \toprule
        \textbf{Question Id} & \textbf{Question} & \textbf{Physician Analysis} \\
        \midrule
        25,429,481 & Are reasons why erupted third molars extracted in a public university in Mexico? & The sentence is malformed and we cannot understand what is asked. \\
        \addlinespace[3pt]
        25,440,451 & Do youth walking and biking rates vary by environments around 5 Louisiana schools? & No clinical relevance. \\
        \addlinespace[3pt]
        25,428,423 & Does [ Descriptive study of healthcare professionals' management of tick bite ]? & Not a sentence. \\
        \addlinespace[3pt]
        25,423,540 & Do a critical analysis of secondary overtriage to a Level I trauma center? & This is not a question. \\
        \bottomrule
    \end{tabular}
\end{table}

\section{Dataset description}
\subsection{Dataset Prompting Modifications}

For all datasets we corrected formatting mistakes, grammatical, and spelling errors. We also moved the explanation note before the score in the HELM JSON output, changed the MedicationQA \citep{abacha2019bridging} prompt to not reference the MedDialog \citep{he2020meddialog} dataset, and moved judge system prompts to the first line of the judge user prompt, as verifiers does not support judge system prompts at this time. AgentClinic received the most modifications: we included reference exams and instructions to prevent common model failures in the system and user prompts, and added Patient and System prefixes to distinguish patient responses from periodic system messages reminding the model how many turns were left.

\subsection{MedMCQA~\cite{pal2022medmcqa}}
MedMCQA is a large-scale multiple-choice medical QA benchmark spanning 21 subjects. The dataset was constructed from publicly available AIIMS and NEET-PG medical entrance exam questions (1991-2022), as well as curated mock and online test series authored by medical professionals. In total, the dataset contains approximately 194k multiple-choice questions covering a diversity of topics and reasoning patterns.

\textbf{Task}  The dataset assesses both factual recall and reasoning across medical subjects. Some example reasoning categories include:
\begin{itemize}
\item Factual  - Retrieval of facts as answers.
\item Explanation / definition - Identifying definition or explanation or a term/phenomenon
\item Diagnosis - Selecting the correct cause of a given ailment / condition.
\item Treatment - Selecting the correct treatment for a given ailment / condition.
\item Teleology / purpose - Understanding the purpose of a phenomenon.
\item Analogy - Selecting the most similar / analogous answer.
\item Comparison - Reasoning via comparing multiple options.
\item MultiHop Reasoning - Reasoning required from multiple passages.
\item Mathematical - Requiring mathematical critical thinking and logical reasoning.
\item Natural language inference - Determining whether a hypothesis is true, false, or neutral given an assumption.
\end{itemize}

\textbf{Inputs/Outputs}  Question text $\rightarrow$ Answer label; 

\textbf{Evaluation}  The dataset uses accuracy (percentage of correct answers) for evaluation. MedMCQA is a single-turn multiple-choice QA task, where the model selects one correct option (A-D) given a question and four answer choices. \textsc{Medmarks} supports both standard and reasoning-mode evaluation, depending on the parser configuration. 

\subsection{MedQA~\cite{jin2021disease}}
MedQA is a open-domain multiple-choice question answering (OpenQA) dataset for solving medical problems. It was built from medical licensing exams in three regions: the United States (USMLE, English, 12,723 questions), Mainland China (MCMLE, Simplified Chinese, 34,251 questions), and Taiwan (TWMLE, Traditional Chinese, 14,123 questions). Each question comes with four answer options and requires selecting the correct one using evidence retrieved from a large corpus of medical textbooks (18 in English and 33 in Chinese). \textsc{Medmarks} only uses the USMLE portion of the MedQA dataset

\textbf{Task} Select the most appropriate answer using the evidence found in the document collection. Given: 
\begin{itemize}
\item A question (short fact-based or long clinical case)
\item 4 answer candidates (A–D)
\item A document collection (medical textbook paragraphs)
\end{itemize}
Types of questions:
\begin{itemize}
\item Type 1 (single-knowledge) — simple factual recall ($\approx2\%$ in USMLE; $\approx70\%$ in Chinese sets)
\item Type 2 (clinical reasoning) — complex patient cases requiring multi-hop reasoning.
\end{itemize}

\textbf{Inputs/Outputs} Question text $\rightarrow$ Answer label; 

\textbf{Evaluation}  Model performance was evaluated using accuracy, defined as the percentage of questions for which the model selects the correct answer from the four multiple-choice options. This is the standard evaluation metric for the MedQA benchmark and provides a clear measure of the model's medical knowledge and clinical reasoning capabilities.

\subsection{PubMedQA~\cite{jin2019pubmedqa}}
PubMedQA is intended to improve and assess the ability of a model to answer biomedical questions requiring reasoning over research texts, in particular quantitative contents of biomedical research paper abstracts. The dataset is based on PubMed research abstracts. 

\textbf{Task}  
\begin{itemize}
\item PQA-L(abled) contains 1,000 human-annotated instances sampled from PubMed abstracts with question-mark titles and structured subsections, including a conclusion. The titles serve as questions, and the conclusion subsection is assumed to hold the answer
\item PQA-U(nlabeled) consists of unlabeled instances from the same pool, excluding titles that start with wh-words.
\item PQA-A(rtificial) contains artificially generated instances from structured abstracts whose titles follow an NP-(VBP/VBZ) part-of-speech pattern. Titles are automatically converted into questions by prepending "is/are" or "do/does," correcting for coherence, and adding a question mark, while yes/no answers are inferred from the original title’s negation. 
\end{itemize}

\textbf{Inputs/Outputs}  Abstract + Question $\rightarrow$ long answer(PQA-L and PQA-A include yes/no labels); 

\textbf{Evaluation} accuracy

\subsection{MedConceptsQA~\cite{shoham2024medconceptsqa}}
MedConceptsQA is a multiple-choice question-answering benchmark specifically targeting the medical coding domain. It evaluates model understanding across diagnoses, procedures, and drugs, using major vocabularies such as ICD9-CM, ICD10-CM, ICD9-PROC, ICD10-PROC, and ATC codes. The dataset aims to measure the ability of language models to identify correct medical code descriptions. 

The MedConceptsQA test-set span three established vocabularies—Diagnoses, Procedures, and Drugs—and target three difficulty levels (easy, medium, hard) to comprehensively evaluate both surface-level and deeper concept understanding.

MedConcepts QA is constructed programmatically by pairing medical codes from canonical ontologies (ICD9, ICD10, ATC) with their descriptions. Incorrect answer choices are sampled at matching difficulty levels from the same vocabulary, creating distractors.

\textbf{Task}  Answer questions to determine the correct medical code; 

\textbf{Inputs/Outputs}  Question $\rightarrow$ Answer; 

\textbf{Evaluation} The dataset uses standard classification accuracy (percentage of correct selections) as the main evaluation metric.

\subsection{MedCalc-Bench~\cite{khandekar2024medcalc}}
MedCalc-Bench dataset designed to evaluate language models’ ability to perform clinical calculations. This evaluation is important because clinical calculators provide a systematic, rapid way to assess a patient’s health status and support clinical decision-making. Each dataset example consists of a patient note and a question prompting the model to compute a specific medical value.

The dataset contains 1,100 questions: 20 questions for each of 55 distinct calculators. These calculators fall into two categories: rule-based and equation-based. Rule-based calculators (e.g., HAS-BLED and HEART) assign a discrete score by summing the number of criteria a patient satisfies. Equation-based calculators (e.g., Estimated Due Date and the Framingham Risk Calculator) apply a defined formula to produce a numeric value, date, or duration. The dataset also comes with step-by-step explanations for how the answer was computed. 

\textbf{Task}  Compute a numeric/structured result from a clinical math prompt; 

\textbf{Inputs/Outputs} Problem statement $\rightarrow$ numeric answer (optionally steps); 

\textbf{Evaluation} Numeric accuracy / tolerance-based scoring

\subsection{HealthBench~\cite{arora2025healthbench}}
HealthBench is an open-source benchmark designed to evaluate large language model performance and safety in realistic healthcare conversations. The dataset consists of 5,000 multi-turn conversations between a model and either an individual user or healthcare professional, spanning diverse geographies, languages, and healthcare contexts. Each conversation was created and evaluated using conversation-specific rubrics written by a cohort of 262 physicians from 60 countries across 26 medical specialties. The benchmark measures 48,562 unique rubric criteria covering various dimensions of model behavior including clinical accuracy, completeness, communication quality, context awareness, and instruction following.

Most conversations were synthetically generated using a tailored language model pipeline based on physician-enumerated situation types, with additional examples derived from physician red teaming exercises and HealthSearchQA (Google's frequently-searched health queries dataset). Conversations were filtered for relevance, realism, and self-consistency before physicians wrote conversation-specific evaluation rubrics.

HealthBench includes two important subsets: HealthBench Consensus (3,671 examples with 34 pre-defined consensus criteria validated by multiple physicians for critical behaviors like emergency referrals) and HealthBench Hard (1,000 examples selected for difficulty where current frontier models score $\leq 32\%$).

\textbf{Task}  HealthBench evaluates open-ended, multi-turn conversational responses across seven themes representing real-world health interaction challenges:
\begin{itemize}
\item Emergency referrals: Recognizing medical emergencies and providing appropriate care recommendations
\item Context-seeking: Identifying when key information is missing and seeking the most informative context
\item Global health: Adapting responses to varied healthcare contexts, resource availability, and regional disease patterns
\item Health data tasks: Completing structured tasks like clinical documentation, decision support, and research assistance
\item Expertise-tailored communication: Matching response complexity and terminology to user expertise level (clinician vs. layperson)
\item Responding under uncertainty: Recognizing and appropriately hedging when information is incomplete or medical knowledge is uncertain
\item Response depth: Adjusting detail level to match user needs without overwhelming or omitting critical information
\end{itemize}
Each rubric criterion is categorized into one of five axes representing behavioral dimensions:
\begin{itemize}
\item Accuracy: Factual correctness and alignment with medical consensus
\item Completeness: Including all safety-relevant and necessary information
\item Communication quality: Clarity, appropriate technical depth, and vocabulary matching
\item Context awareness: Responding appropriately to contextual cues and seeking clarification when needed
\item Instruction following: Adhering to specific user instructions while prioritizing safety
\end{itemize}
The benchmark tests both consumer-facing and clinician-facing interactions, requiring models to demonstrate medical knowledge, clinical reasoning, safety awareness, and adaptive communication across diverse scenarios;

\textbf{Inputs/Outputs}  Dialogue History $\rightarrow$ free-form responses; 

\textbf{Evaluation} Benchmark-defined scoring (judge-based rubric).

\subsection{MedDialog~\cite{he2020meddialog}}

MedDialog is a benchmark of real-world doctor-patient conversations focused on health-related concerns and advice. Each dialogue is paired with a one-sentence summary that reflects the core patient question or exchange. 

\textbf{Task} Condense medical dialogue into concise, informative summaries; 

\textbf{Inputs/Outputs} dialogue history $\rightarrow$ next response (or derived label); 

\textbf{Evaluation} Rubric overview: LLM-as-a-judge evaluation using prompts adapted from MedHELM (single or multi-judge)
Evaluation dimensions:
\begin{itemize}
\item Accuracy (1-5): Does the summary correctly capture the main medical issue and clinical details?
\item Completeness (1-5): Does the summary include all important medical information?
\item Clarity (1-5): Is the summary easy to understand for clinical use?
\end{itemize}

\subsection{ACI-Bench~\cite{yim2023aci}}
ACI-Bench (Automated Clinical Intelligence Benchmark) is a benchmark designed to evaluate the ability of language models to perform clinically relevant summarization. The dataset consists of transcribed patient-doctor dialogues and their corresponding ground-truth clinical notes. It measures a model's proficiency in distilling lengthy, conversational medical encounters into well-structured, concise, and accurate summaries suitable for Electronic Health Records (EHR). The dataset was generated from various sources, including real-world clinical NLP challenges and transcribed medical conversations. 

\textbf{Task}  The primary task is abstractive summarization of a clinical dialogue. The model must process a conversational transcript and generate a formal, structured clinical note. This tests a combination of skills:
\begin{itemize}
\item Information Distillation: Extracting medically salient information (symptoms, history, exam findings, plans) from a noisy, conversational format.
\item Clinical Reasoning: Understanding the context of the conversation to correctly place information within the appropriate section of the clinical note.
\item Summarization \& Formatting: Condensing the dialogue into a concise summary while adhering to the conventional structure of a medical note.
\item Orientation: The benchmark is entirely clinician-facing, as the end product is a summary intended for use by healthcare professionals.
\end{itemize}
\textbf{Inputs/Outputs} Conversational transcript $\rightarrow$ structured clinical note; 

\textbf{Evaluation} The original ACI-Bench paper used a suite of standard NLP metrics to evaluate summarization quality. \textsc{Medmarks} implements a direct replication of this metric-based approach and does not use an LLM-as-a-Judge. The calculated metrics include:
\begin{itemize}
\item N-gram based metrics: ROUGE-1, ROUGE-2, and ROUGE-L to measure textual overlap with the reference summary.
\item Semantic similarity metrics: BERTScore (using microsoft/deberta-xlarge-mnli) and BLEURT to evaluate the semantic equivalence between the generated summary and the reference.
\item The original evaluation also included a UMLS-based F1 score to measure the recall and precision of medical concepts. This is currently omitted from our implementation for simplicity but can be added in the future.
\end{itemize}

\subsection{MedAgentBench v2~\cite{jiang2025medagentbench}}
MedAgentBench is a virtual electronic health‑record (EHR) environment designed to evaluate the agentic capabilities of large language models. It comprises 300 clinically derived tasks written by two internal‑medicine physicians.  Tasks reflect common information‑retrieval and order‑entry workflows in inpatient and outpatient settings; they include retrieving patient demographics, lab results or vitals, documenting new measurements, ordering tests, referrals or medications and performing data aggregation.  The environment provides 100 de‑identified patient profiles drawn from the Stanford Research Repository (STARR); each profile includes lab tests, vital signs, procedure orders, diagnoses and medication orders collected over the previous five years.

MedAgentBench V2 refined the system prompt, added built‑in tools for mathematical calculations and formatting, and introduced a memory component that appends instructions after each failure so the agent can learn from mistakes.  The V2 paper also adds 300 new multi‑step tasks to test generalization to unseen workflows.

\textbf{Task}  MedAgentBench tasks require a combination of:
\begin{itemize}
\item Clinical knowledge and reasoning – Agents must interpret instructions about lab thresholds, medication dosing (e.g., titrating potassium replacement) and ordering criteria .
\item Planning and tool use – Tasks often involve multiple steps such as retrieving a lab value, checking its recency and ordering a test if necessary.  Agents must plan a sequence of API calls and respond within the 8‑round limit .
\item Patient‑facing communication – Some tasks require composing messages to patients (e.g., explaining what to do for a wound).  This tests the model’s ability to generate clear and professional language.
\item FHIR compliance – Agents need to issue correctly structured GET and POST requests using FHIR resource types (Observation, MedicationRequest, Procedure, etc.), which tests understanding of standard healthcare APIs.Thus MedAgentBench emphasises agentic reasoning, planning and execution in EHRs, not just factual Q\&A.
\end{itemize}
\textbf{Inputs/Outputs} Environment state + goal $\rightarrow$ GET/POST request or end conversation; 

\textbf{Evaluation}  
\begin{itemize}
\item Task success rate (SR) – The main metric is the proportion of tasks the agent completes successfully.  A task succeeds if the agent issues correct API calls (for GET or POST) and produces the requested information or orders without exceeding the 8‑round limit .  Failures occur when invalid actions are requested or the agent runs out of interaction rounds .  Success rates are reported overall and separately for query tasks (information retrieval) and action tasks (modifying the EHR). Table 3 of the paper shows that state‑of‑the‑art models achieve overall SRs between $4.0\%$ and $69.67\%$.
\item Pass@1 – Unlike code benchmarks where multiple attempts are averaged, MedAgentBench evaluates agents with a single attempt (pass@1), reflecting the high‑stakes clinical setting where even one error is unacceptable .
\item Memory‑enhanced SR – In the V2 paper, the authors report that adding a memory component raised GPT‑4.1’s success rate from $91.0\%$ without memory to $98.0\%$ with memory .  This shows that simple prompt‑engineering and memory can significantly improve agent reliability.
\end{itemize}

\subsection{AgentClinic~\cite{schmidgall2024agentclinic}}
AgentClinic is a simulated clinical environments through interactive diagnostic dialogues. It measures a model’s ability to perform clinical reasoning, gather information from a patient, order appropriate tests, and make a final diagnosis. The benchmark contains two components:
\begin{itemize}
\item AgentClinic-MedQA: dialogue-only cases derived from medical licensing-exam style problems.
\item AgentClinic-NEJM: multimodal cases adapted from New England Journal of Medicine clinical challenges that include both text and medical imagery.
\end{itemize}
In \textsc{Medmarks}, a doctor agent interacts with a simulated patient agent and measurement agent for up to 20(default) conversational turns before declaring a final diagnosis using the phrase “DIAGNOSIS READY: […]”.

\textbf{Task} Evaluate models ability to: 
\begin{itemize}
\item Sequential clinical reasoning and decision-making under uncertainty.
\item Information gathering through doctor/patient dialogue.
\item Appropriate test ordering and interpretation.
\item Synthesis of clinical findings into a diagnosis.
\item Clear clinician-facing communication.
\end{itemize}
\textbf{Inputs/Outputs} Patient case dialogue for context context $\rightarrow$ diagnosis; 

\textbf{Evaluation}  LLM-as-a-Judge binary evaluation. A moderator model determines if the doctor’s diagnosis and gold answer describe the same disease. 

\subsection{LongHealth~\cite{adams2025longhealth}}
The Longhealth benchmark consists of 20 detailed synthetic patient cases covering various diseases, with each case containing between 5,090 to 6,754 words. The LongHealth benchmark challenges LLMs with 400 multiple-choice questions categorized into information extraction, negation, and sorting, providing a robust assessment tool for LLMs in handling real-world, lengthy clinical data.

\textbf{Task}  Each task is repeated 5x
\begin{itemize}
\item Task 1: This task measures information retrieval. The model must extract the correct information from a set of long patient documents, where the answer IS definitely present. To ensure a thorough test of information extraction, the documents are repeatedly shuffled, and the question is asked five times. Standard Accuracy (percentage of correct answers) is used here.

\item Task 2: This task focuses on the model's robustness to irrelevant data. The model is presented with the target patient's documents mixed with documents from completely unrelated patients. It must accurately pull the correct information despite the distraction to answer a question related to the target patient. 

\item Task 3: This task tests the model's ability to recognize its limitations and appropriately refuse to answer. The model is queried about a patient whose documents were intentionally excluded from the context. A "Cannot be answered" option is added, and the model's accuracy is measured based on how often it correctly selects this refusal option. 
\end{itemize} 

\textbf{Inputs/Outputs} Long document + question $\rightarrow$ answer (often short-form); 

\textbf{Evaluation} Accuracy / match metrics

\subsection{MedCaseReasoning~\cite{wu2025medcasereasoning}}
MedCaseReasoning is a dataset for evaluating LLMs on diagnostic reasoning. This is needed due to both the outcome and reasoning needed to be correct for the entire answer to be accurate. The dataset includes detailed reasoning statements derived from medical case reports. 

\textbf{Task} When presented with a medical case, infer diagnoses and reasoning steps from narrative cases; 

\textbf{Inputs/Outputs} Case text $\rightarrow$ diagnosis/assessment (optionally reasoning); 

\textbf{Evaluation} LLM as a judge using accuracy metric. 

\subsection{Med-HALT~\cite{pal2023med}}
Med-HALT (Medical Domain Hallucination Test) is a clinical reasoning evaluation benchmark designed to assess hallucinations and false confidence in large language models within the medical domain. Med-HALT focuses on scenarios in which a model must judge the correctness of a proposed answer to a medical question or recognize when no provided option is correct, rather than generating a new answer from scratch.

The dataset is constructed around multiple-choice clinical questions paired with student answers of varying correctness. The core objective is to evaluate whether a model can correctly identify incorrect or unsupported answers, a common failure mode in medical reasoning systems. Med-HALT includes multiple test variants, most notably the False Confidence Test (FCT) and the None of the Above (NOTA) test, which together probe a model’s ability to avoid confidently endorsing incorrect medical statements.

In \textsc{Medmarks}, we use the publicly released Med-HALT dataset and focus on its False Confidence Test (FCT) and None of the Above (NOTA) variants, aligning the evaluation with safety-critical hallucination detection and answer-judgment scenarios.

\textbf{Task} 
\begin{itemize}
\item Task: Single-turn multiple-choice clinical reasoning and hallucination detection.
\item Skills tested (inferred from dataset design and evaluation): \begin{itemize}
\item Recognition of incorrect or unsupported medical statements
\item Calibration of confidence in clinical reasoning contexts
\item Identification of situations where no valid answer is present (NOTA)
\end{itemize}
\item Orientation: Safety- and reliability-focused clinical reasoning evaluation (not answer generation).
\end{itemize}

\textbf{Inputs/Outputs} Question $\rightarrow$ response that should remain grounded/abstain as appropriate; 

\textbf{Evaluation}
Original Metric: 
Binary accuracy based on whether the model correctly identifies:
\begin{itemize}
\item An incorrect proposed answer (False Confidence Test), or
\item The absence of any correct option (None of the Above Test)
\end{itemize}
\textsc{Medmarks} implementation:
\begin{itemize}
\item Evaluation method: Multiple-choice selection, where the model chooses the correct option (e.g., identifying an answer as incorrect or selecting “None of the Above”).
\item Test types used:
\begin{itemize}
\item reasoning\textunderscore FCT – evaluates whether the model can correctly assess a proposed answer and avoid false confidence.
\item reasoning\textunderscore nota – evaluates whether the model can correctly identify when no provided option is correct.
\end{itemize}
\item Split: The \textsc{Medmarks} environment filters to the validation (val) subset of the dataset, consistent with \textsc{Medmarks} evaluation standards.
\item Scoring: Binary accuracy
\begin{itemize}
\item 1.0 if the parsed answer choice matches the gold label
\item 0.0 otherwise
\end{itemize}
\item Parsing: Structured output parsing (XML or boxed format), with strict extraction of the selected answer option.
\end{itemize}

\subsection{MEDEC~\cite{abacha2025medecbenchmarkmedicalerror}}
MEDEC is a benchmark designed to evaluate a model's ability to detect and correct medical errors within clinical notes. The dataset consists of 3,848 clinical texts and measures five specific types of errors: Diagnosis, Management, Treatment, Pharmacotherapy, and Causal Organism.

\textbf{Task}  The benchmark's primary task is medical error detection and correction. This is divided into three specific subtasks:
\begin{itemize}
\item (Subtask A) Error Flag Prediction: Predicting whether a given clinical text contains an error (binary classification: 0 for correct, 1 for error).
\item (Subtask B) Error Sentence Detection: For texts flagged with an error, extracting the specific sentence ID that contains the error.
\item (Subtask C) Correction Generation: For texts flagged with an error, generating a corrected version of the erroneous sentence.
\end{itemize}

\textbf{Inputs/Outputs} Bote (possibly corrupted) $\rightarrow$ error spans and/or corrected note; 

\textbf{Evaluation} Detection F1 and correction quality measures

\subsection{MedHallu~\cite{pandit2025medhallucomprehensivebenchmarkdetecting}}
MedHallu is a comprehensive benchmark specifically designed to evaluate the ability of Large Language Models (LLMs) to detect hallucinations in the medical domain. It addresses the critical need for reliability in high-stakes medical question-answering where incorrect information can risk patient safety.

\textbf{Task}  Each entry includes a medical question, a ground-truth answer, and a systematically generated "hallucinated" answer. It is structured to evaluate how well models can distinguish between accurate medical information and plausibly written but factually incorrect hallucinated responses; 

\textbf{Inputs/Outputs}  Prompt + model output (and possibly references) $\rightarrow$ hallucination label/score; 

\textbf{Evaluation} The benchmark uses classification metrics to assess how effectively a model can distinguish between correct (ground truth) and hallucinated answers. This includes Overall F1 score, Overall Precision, $\Delta$ Knowledge, which is the performance change in overall F1 when knowledge is provided. 

\subsection{HEAD-QA v2~\cite{correa2025head}}
HEAD-QA v2 is a multiple-choice question answering (MCQA) benchmark designed to evaluate specialized reasoning and domain knowledge across graduate-level healthcare questions. Each sample contains a question, any referenced image, candidate options, and the correct answer. The dataset covers multiple healthcare areas, including medicine, nursing, biology, chemistry, psychology, and pharmacology, with questions ranging from technical content to some social issues. 

\textbf{Task}  Answer healthcare questions (often multiple-choice). The dataset encourages research on effective information retrieval, reasoning and cross-lingual understanding. Models are required to combine specialized knowledge with reasoning to select the correct answer from multiple options.; 

\textbf{Inputs/Outputs} Question $\rightarrow$ selected option/answer; 

\textbf{Evaluation} The dataset is evaluated using accuracy (percentage of correct answers). In \textsc{Medmarks} environments, exact-match accuracy is computed by parsing the model output and comparing the predicted answer option or answer text against the ground-truth answer. The evaluation supports questions with either four or five answer options, depending on the exam year.

\subsection{PubHealthBench~\cite{harris2025pubhealthbench}}
PubHealthBench is a public‑health knowledge benchmark released in May 2025 by the UK Health Security Agency and collaborators.  Its goal is to test large language models’ knowledge of up‑to‑date UK Government public‑health guidance.  The dataset contains 8,850 multiple‑choice questions (MCQA) with one correct answer and six distractors derived from 687 UK Government guidance documents.  The questions span 10 public‑health topic areas (e.g., infectious‑disease control, vaccination, extreme weather, chemical exposures) and were created automatically: the authors scraped text from HTML and PDF documents on the UK Government website (gov.uk) on 8 Jan 2025, chunked it into passages, then prompted a large language model to draft questions and answer options.  All questions were grounded in a passage of guidance and later reviewed by human experts; a subset of 760 questions (PubHealthBench‑Reviewed) underwent manual annotation to mark invalid questions or errors.

\textbf{Task} PubHealthBench offers two tasks:
\begin{itemize}
\item Multiple‑choice question answering (MCQA) – Models must choose the correct answer from seven options.  This tests factual knowledge of UK public‑health guidance and reading comprehension of the question.  Because the guidance is aimed at the general population, the task evaluates general public‑health knowledge rather than clinical reasoning.
\item Free‑form response – Using the same questions as MCQA, models must produce an open‑ended answer.  The LLM‑judge checks whether the answer is consistent with the source guidance .  This task examines the model’s ability to recall and summarize relevant guidance and produce coherent prose.
\end{itemize}

\textbf{Inputs/Outputs} Question $\rightarrow$ multi-choice answer or free-form response; 

\textbf{Evaluation}  
\begin{itemize}
\item MCQA accuracy – The principal metric is accuracy of the model’s answer on the multiple‑choice questions.  The paper reports overall and per‑topic accuracies for many models and uses Wilson score confidence intervals to estimate uncertainty 
\item LLM‑as‑a‑Judge (Free‑form responses) – For free‑form answers the authors use a judge model to score whether the response is consistent with the source text and the correct MCQA answer.  The judge (GPT‑4o‑mini) receives the question, ground‑truth answer, the model’s response and six retrieved context chunks, and produces a binary decision. This LLM‑judge approach allows unstructured answers to be graded automatically.
\item Reviewed vs full set – The authors created a small manually reviewed subset (PubHealthBench‑Reviewed) to validate that the full automatically generated set yields similar results.  They found high correlation between accuracies on the full test set and the reviewed subset.
\end{itemize}

\subsection{MedExQA~\cite{kim2024medexqamedicalquestionanswering}}
MedExQA is a medical QA benchmark that includes multiple explanations per question. 

\textbf{Task}  The dataset is a multiple-choice medical QA task: for each question a model must select the correct answer choice and then generate an explanation (in free text) justifying the answer. There are two pre-written “gold” explanations per item, allowing evaluation of explanation generation diversity and correctness.; 

\textbf{Inputs/Outputs} Question $\rightarrow$ answer + explanation; 

\textbf{Evaluation} 
\begin{itemize}
\item Original evaluation metrics (from paper): \begin{itemize}
\item Classification accuracy on the MCQ answer (string match heuristic for extraction)
\item Explanation generation evaluation using standard lexical/embedding metrics, including BLEU, ROUGE-L, METEOR, and BERTScore(using sciBERT) from the evaluate library
\item Human evaluation of explanation-answer pairs: annotators scored responses as 0 (wrong answer/no explanation/irrelevant), 0.5 (correct answer but incorrect/incomplete explanation) or 1.0 (correct answer + correct explanation) for a small dev set.
\end{itemize}
\item Our evaluation metric: \begin{itemize}
\item LLM-as-a-judge metric for explanation evaluation using a judge model to evaluate the generated explanation in terms of quality, and compared to reference explanations, given question, answer, and reference explanations.
\item The judge scores the explanation as equivalent or inequivalent to the two reference explanations which are defined as: \begin{itemize}
\item The assistant's reasoning is equivalent if its logic is semantically aligned with at least one reference reasoning trace. It may paraphrase or omit minor details, as long as the central reasoning and decision criteria are the same and do not conflict with that reference trace.
\item The assistant's reasoning is inequivalent if it clearly contradicts both reference reasoning traces or relies on logic that is incompatible with both traces (for example, it uses a different main reason for the answer that conflicts with the references).
\end{itemize}
\item Joint metric: e.g., $joint\_score = 0.5$ if explanation not equivalent, $1$ if explanation is equivalent, $0$ if MCQ answer is incorrect. 
\end{itemize}
\end{itemize}

\subsection{MetaMedQA~\cite{griot_large_2025}}
MetaMedQA targets metacognition in medical MCQ settings (e.g., confidence and unknown handling). It extends the MedQA-USMLE benchmark by introducing unanswerable, ambiguous, or fictional medical questions to test epistemic humility. Additionally, it incorporates questions from the Glianorex benchmark to assess model’s ability to recognize the limits of their knowledge. The dataset consists of 1,373 multiple-choice items, each with answer options labeled A–F with only one correct choice. In addition to the standard A-D answer choices, the benchmark adds “None of the above” and “I don’t know / cannot answer” to the possible choices.

\textbf{Task} 
\begin{itemize}
\item Core medical-knowledge recall on standard questions.
\item Recognition of uncertainty on malformed or impossible questions.
\item Metacognitive reasoning deciding when to answer “I don’t know.”.
\end{itemize}
\textbf{Inputs/Outputs} question $\rightarrow$ selected option + confidence; 

\textbf{Evaluation}  
\begin{itemize}
\item Original metric: Accuracy, measured by comparing the model’s selected answer to the gold option.
\item Implemented metric: Deterministic exact-match accuracy, extracts the first A–Z letter from the model’s output and compares it with the reference key.
\item Judge model: None required, evaluation is rule-based.
\end{itemize}

\subsection{MedXpertQA~\cite{zuo2025medxpertqa}}
MedXpertQA includes 4460 high-difficulty medical exam questions spanning 17 specialties (and 11 body systems. It includes two subsets, MedXpertQA Text for text medical evaluation and MedXpertQA MM for multimodal medical evaluation but as of now, we only use the Text subset for our environment.

\textbf{Task}  Answer MCQs with approximately 10 options per question; 

\textbf{Inputs/Outputs} Question $\rightarrow$ selected option; 

\textbf{Evaluation}  The model is asked to generate one and only one option as the answer, say, A’. A’ is considered correct if it exactly matches the ground truth A. The metric is hence binary as follows:
Reward $r = 1$ if $A' = A$ else $0$

\subsection{MMLU-Pro-Health~\cite{wang2024mmlu}}
MMLU-Pro-Health is built upon MMLU. Roughly half of the dataset is derived from MMLU, after filtering out erroneous and easy questions, and the other half is sourced from StemEZ, TheoremQA, and SciBench. Unlike MMLU, which only has four answer choices per question, MMLU-Pro includes up to ten options that were generated to be plausible distractors, thus reducing noise from spurious guessing. Experts reviewed the dataset to ensure that the questions are correct and that distractor answer choices are reasonable. 

\textbf{Task}  Answer MCQs with approximately 10 options per question; 

\textbf{Inputs/Outputs} Question $\rightarrow$ selected option; 

\textbf{Evaluation}  A simple correctness metric that checks the model’s answer against the ground truth letter choice. 

\subsection{M-ARC~\cite{kim2025limitationslargelanguagemodels}}
The M-ARC benchmark, also referred to by the authors as MedARC-QA (Medical Abstraction and Reasoning Corpus), tests a model’s ability to break away from rote reasoning patterns in the face of pertinent clinical information. This benchmark consists of a single test split of 100 USMLE-style multiple-choice questions, written by the authors, each with up to seven answer choices (the majority have only five). 

\textbf{Task}  Answer MCQs with approximately 5 and up to 7 answers per question; 

\textbf{Inputs/Outputs} Question $\rightarrow$ selected option; 

\textbf{Evaluation}  A simple correctness metric to check the model’s answer against the ground truth letter choice.

\subsection{Medbullets~\cite{chen-etal-2025-benchmarking}}
The Medbullets dataset is a collection of USMLE Step 2 and Step 3 style questions taken from the Medbullets X account. Each entry contains a short patient encounter note, a multiple-choice question that requires clinical reasoning, and an explanation of the correct answer. It includes two splits, each consisting of 308 rows: op4\_test, with four answer choices per question, and op5\_test, with five. Note, however, that the content between each split is identical. The only difference is that the five-option split has an additional answer choice per question, increasing the difficulty.  

\textbf{Task}  Answer MCQs; 

\textbf{Inputs/Outputs} Short patient note + question $\rightarrow$ selected option; 

\textbf{Evaluation} Simple correctness metric to check the model’s answer against the ground truth letter choice.

\subsection{SuperGPQA-Med~\cite{pteam2025supergpqascalingllmevaluation}}
SuperGPQA is a large multiple-choice QA benchmark designed to test graduate-level knowledge and reasoning across 285 subfields (grouped into 72 fields and 13 high-level disciplines). The Medicine discipline contains 2,755 questions (out of 26,529 total).
The dataset measures single-best-answer MCQ performance under standardized prompting, with analysis broken down by taxonomy (discipline/field/subfield) and difficulty. It was generated using a human–LLM collaborative pipeline with three main stages: Source Screening $\rightarrow$ Transcription $\rightarrow$ Quality Inspection, where expert annotators first collect credible sources, crowd annotators standardize them into MCQs (including generating distractors), and a multi-stage inspection process removes ambiguous/trivial items and reworks easy items to maintain discrimination.

\textbf{Task}  Answer MCQs; 

\textbf{Inputs/Outputs} Question $\rightarrow$ selected option; 

\textbf{Evaluation}  Accuracy (did the model pick the correct option), with breakdowns by discipline/field/subfield and by easy/middle/hard splits.

\subsection{SCT-Public~\cite{mccoy2025assessment}}
Sctpublic is a publicly available benchmark designed to evaluate clinical reasoning under uncertainty. It comprises 750 Script Concordance Test (SCT) questions collected from 10 international datasets spanning multiple specialities, including internal medicine, emergency medicine, neurology, pediatrics, surgery, psychiatry, and physiotherapy. Each item presents a clinical vignette followed by a new piece of information, and the model must assess how this new information changes the likelihood of a diagnosis, investigation, or treatment decision using a 5-point Likert scale. Responses are scored against a reference panel of experienced clinicians, with partial credit given when answers differ reasonably from the experts. Human performance data were available for several tests, allowing direct comparison. 

\textbf{Task}  Single-turn clinical reasoning evaluation; 

\textbf{Inputs/Outputs}  Previous reasoning and updated information $\rightarrow$ Likelihood of original reasoning being true(5 options, much less likely to much more likely); 
 
\textbf{Evaluation} SCTPublic scores model responses using a partial-credit system based on expert answers. Each response is mapped to one of five numeric ratings: -2, -1, 0, +1, +2. The score depends on how often experts chose that answer, with the most common answer earning a full score of 1.0 and other answers receiving proportionally less credit. This system captures flexible, probabilistic reasoning and measures how well models update their judgments like experts under uncertainty.

\subsection{MedicationQA~\cite{benabacha2019medicationqa}}
MedicationQA is a consumer-facing medical question answering benchmark focused on answering real-world questions about medications. The dataset originates from a gold-standard corpus introduced by Ben Abacha et al. (MedInfo 2019), consisting of 674 manually curated consumer medication question–answer pairs derived from real questions submitted to MedlinePlus. Reference answers were authored and validated by medical experts using authoritative sources such as MedlinePlus, DailyMed, and other trusted U.S. government and academic resources.

While the original MedicationQA work focused on question understanding and retrieval subtasks rather than end-to-end answer quality, more recent evaluation frameworks (e.g., MedHELM) have repurposed MedicationQA-style tasks for holistic assessment of generated answers using rubric-based LLM evaluation. We follow this modern approach by evaluating full answer derived from quality rather than isolated components.
 
\textbf{Task} Open-ended medical question answering about medications.
\begin{itemize}
\item Skills tested (inferred from dataset design and evaluation):
\item Medication-related medical knowledge (e.g., indications, dosing, interactions, contraindications)
\item Faithfulness to expert reference answers (evaluated via accuracy and completeness)
\item Clear consumer-facing medical communication (evaluated via clarity)
\end{itemize}

\textbf{Inputs/Outputs} Question $\rightarrow$ selected option; 

\textbf{Evaluation} Original metrics:
\begin{itemize}
\item Question focus recognition: Precision, Recall, and F1 (exact and partial span match)
\item Question type classification: Accuracy
\item Answer retrieval: Qualitative analysis (no automatic end-to-end QA metric)
\end{itemize}
\textsc{Medmarks} implementation
\begin{itemize}
\item Evaluation method: LLM-as-a-Judge comparison between the model-generated answer and a reference (“gold”) answer.
\item Judge rubric: The judge evaluates responses along three dimensions:
\begin{itemize}
\item Accuracy – factual correctness and absence of medical misinformation
\item Completeness – coverage of all medically relevant aspects of the question
\item Clarity – structure, readability, and appropriateness for a lay audience
\end{itemize}
\item Scoring scale: Each dimension is scored on a 1–5 ordinal scale
 (1 = very poor, 5 = excellent), with short justifications per dimension.
\item Output format: Structured JSON or XML containing per-dimension scores and rationales.
\item Judge model: Configurable at runtime via the \textsc{Medmarks} evaluation framework (default \textsc{Medmarks} LLM-as-a-Judge configuration).
\end{itemize}

\subsection{MedR-Bench~\cite{qiu2025quantifying}}
MedR‑Bench is a clinical reasoning benchmark designed to assess both the quality of medical reasoning and the correctness of final clinical decisions produced by large language models (LLMs). The dataset contains 1,453 structured patient cases drawn from real case reports in the PubMed Central Open Access Subset published after July 2024.  Cases span 13 body systems and 10 disease categories and include 656 rare disease cases.  Each case includes:
\begin{itemize}
\item Case summary – structured information such as demographics, chief complaint, history of present illness, physical exam findings and ancillary test results.
\item Reasoning process – a sequence of reasoning steps derived from the discussion sections of the source case reports, capturing how clinicians reason to arrive at a diagnosis or treatment.
\item Final outcome – the ground‑truth diagnosis or treatment extracted from the case report.
\end{itemize}
These cases are organized into two splits: MedR‑Bench‑Diagnosis (957 cases) and MedR‑Bench‑Treatment (496 cases) .  The dataset was curated using GPT‑4o to restructure narrative case reports into structured patient cases .  By combining typical and rare conditions and providing step‑by‑step reasoning chains, MedR‑Bench offers a comprehensive testbed for evaluating medical reasoning in LLMs.

\textbf{Task} Three tasks: Examination Recommendation, Diagnostic decision making, Treatment planning; 

\textbf{Inputs/Outputs} Case Summary + question $\rightarrow$ Reasoning trace + task specific recommendation; 

\textbf{Evaluation}  For the final clinical outputs (diagnosis, examination recommendations, treatment plans), MedR‑Bench adopts standard metrics augmented with LLM‑based judging:
\begin{itemize}
\item Accuracy (binary match) – used for diagnoses and other discrete outcomes; the benchmark uses GPT‑4o as a semantic equivalence judge to handle synonyms and paraphrases.
\item Precision and Recall – used for examination recommendations; models produce a list of recommended tests, which is compared with the ground‑truth list to compute list‑wise precision and recall.
\item Treatment‑plan correctness – because treatment plans can vary widely, the evaluation pipeline uses a retrieval‑assisted judge: it extracts keywords, retrieves relevant evidence and uses GPT‑4o to determine if the proposed plan aligns with the case context.
\end{itemize}

\subsection{CareQA~\cite{arias-duart-etal-2025-automatic}}
CareQA is a medical question answering dataset, available in both the English and Spanish language. The dataset consists of both multiple-choice and open-ended medical questions covering a comprehensive range of healthcare topics and specialties. The dataset contains medical questions spanning biochemistry, anatomy, physiology, pathology, pharmacology, epidemiology, immunology, genetics, pediatrics, gynecology, cardiology, neurology, psychiatry, orthopedics, and various other medical specialties. The dataset originates from official sources of the Spanish Specialized Healthcare Training (FSE) examinations, including the biology, chemistry, medicine, nursing, pharmacology, and psychology tests from 2020 to 2024. The English translation was performed using GPT-4, and the open-ended version was created via rephrasing with Qwen2.5-72B-Instruct, followed by human validation.

\textbf{Task} MCQA in English and Spanish. Open-Ended QA in English.; 

\textbf{Inputs/Outputs} Question $\rightarrow$ selected option or open-ended answer; 

\textbf{Evaluation}   
\begin{itemize} 
\item Close-ended Evaluation - For close-ended evaluations, the metric of choice is accuracy.
\item Open-ended Evaluation - For open-ended queries, eleven different metrics were used, sorted into four categories. 
\begin{enumerate} 
\item N-gram based metrics: ROUGE1, ROUGE2, ROUGEL, and BLEU - these evaluate the overlap of n-grams between generated and reference answers.
\item Semantic similarity metrics: BERTScore, BLEURT, and MoverScore - these evaluate semantic similarity between generated and reference text using embeddings or deep learning models.
\item Perplexity metrics: Word Perplexity, Bits per Byte, and Byte Perplexity - these assess the model's predictive capabilities. 
\
\end{enumerate} 
\end{itemize}

\subsection{MTSamples-Procedures~\cite{bedi_holistic_2026}}
MTSamples Procedures is a benchmark composed of transcribed operative notes, focused on documenting surgical procedures. Each example presents a brief patient case involving a surgical intervention, and the model is tasked with generating a coherent and clinically accurate procedural summary or treatment plan. Data is from MTSamples.com. 

\textbf{Task}   Given patient notes(procedure note with PLAN/SUMMARY/FINDINGS sections removed), generate procedural summary or treatment plan; 

\textbf{Inputs/Outputs}  Patient notes $\rightarrow$ treatment plan or summary; 

\textbf{Evaluation}  Rubric using LLM-as-a-Judge using three evaluation dimensions. 
\begin{itemize}
    \item \textbf{Accuracy} (1-5): Does the response provide correct clinical advice that follows established clinical guidelines? 
    \item \textbf{Completeness} (1-5): Does the response include all important aspects of patient care mentioned in the reference? 
    \item \textbf{Clarity} (1-5): Is the response written clearly and organized in a way that is easy to read for clinicians?
\end{itemize}

\subsection{MTSamples-Replicate~\cite{bedi_holistic_2026}}
MTSamples Replicate is a benchmark composed of transcribed treatment plans. Model is given Each example presents a brief patient case, with PLAN removed, but SUMMARY and FINDINGS preserved. The model is tasked with generating a treatment plan. 

\textbf{Task}   Given patient notes(with plan section removed, preserves summary and findings), generate a treatment plan; 

\textbf{Inputs/Outputs}  Patient notes $\rightarrow$ treatment plan; 

\textbf{Evaluation}  Rubric using LLM-as-a-Judge using three evaluation dimensions. `
\begin{itemize}
    \item \textbf{Accuracy} (1-5): Does the response provide correct clinical advice that follows established clinical guidelines? 
    \item \textbf{Completeness} (1-5): Does the response include all important aspects of patient care mentioned in the reference? 
    \item \textbf{Clarity} (1-5): Is the response written clearly and organized in a way that is easy to read for clinicians?
\end{itemize}
\end{document}